\documentclass[10pt,hyphens]{article} 
\usepackage[preprint]{tmlr}

\usepackage{amsmath,amsfonts,bm}

\def\eqref#1{equation~\ref{#1}}

\def\1{\bm{1}}

\DeclareMathAlphabet{\mathsfit}{\encodingdefault}{\sfdefault}{m}{sl}
\SetMathAlphabet{\mathsfit}{bold}{\encodingdefault}{\sfdefault}{bx}{n}

\usepackage{hyperref}

\usepackage{graphicx} 
\usepackage{natbib}  
\usepackage{caption} 

\usepackage{algorithm}
\usepackage{algorithmic}
\usepackage{placeins}
\usepackage{tabularx}
\usepackage{xurl}
\usepackage[most]{tcolorbox}

\tcbuselibrary{raster}
\tcbuselibrary{skins}
\usepackage{listings}
\usepackage{subcaption}

\newtcolorbox{promptbox}[1][Prompt]{%
  enhanced,
  breakable,
  colback=gray!5,
  colframe=gray!55,
  boxrule=0.4pt,
  arc=2pt,
  left=5pt,right=5pt,top=4pt,bottom=4pt,
  fonttitle=\bfseries,
  coltitle=black,
  colbacktitle=gray!18,
  attach boxed title to top left={xshift=6pt,yshift=-2pt},
  boxed title style={colframe=gray!55,boxrule=0.4pt,arc=1pt},
  title=#1,
}
\newtcolorbox{responsebox}[1][Model Response]{%
  enhanced,
  breakable,
  colback=blue!2,
  colframe=blue!35!gray,
  boxrule=0.4pt,
  arc=2pt,
  left=5pt,right=5pt,top=4pt,bottom=4pt,
  fonttitle=\bfseries,
  coltitle=black,
  colbacktitle=blue!8,
  attach boxed title to top left={xshift=6pt,yshift=-2pt},
  boxed title style={colframe=blue!35!gray,boxrule=0.4pt,arc=1pt},
  title=#1,
}

\usepackage{booktabs}
\usepackage{amsfonts}
\usepackage{nicefrac}
\usepackage{microtype}
\usepackage[table]{xcolor}
\definecolor{skyblue}{RGB}{220,240,250}
\definecolor{lightorange}{RGB}{255,235,210}

\newcommand{\bluecaption}[1]{%
    \colorbox{skyblue}{%
        \parbox{\dimexpr\linewidth-2\fboxsep\relax}{#1}%
    }%
}

\newcommand{\orangecaption}[1]{%
    \colorbox{lightorange}{%
        \parbox{\dimexpr\linewidth-2\fboxsep\relax}{#1}%
    }%
}

\newcommand{\stderr}[1]{%
  {\fontsize{6}{6}\selectfont$\pm#1$}}
\usepackage{amsmath}
\usepackage{enumitem}
\usepackage{todonotes}
\usepackage{multirow}
\usepackage{subcaption}
\definecolor{twitterblue}{HTML}{1DA1F2}
\usepackage[most]{tcolorbox}
\tcbuselibrary{breakable}
\usepackage{pifont}

\newcommand{\method}{\textsc{DiffusedDINO}}
\newcommand{\methodshort}{\textsc{DDN}}
\newcommand{\tmethod}{\textsc{TDDN}}

\newcommand{\PuzzlePerception}{\textsc{PuzzlePerception}}

\newcommand{\cmark}{\ding{51}}
\newcommand{\xmark}{\ding{55}}

\usepackage{colortbl}

\hypersetup{
  colorlinks=true,
  linkcolor=black,
  citecolor=black,
  urlcolor=blue
}
\usepackage{fontawesome5}
\usepackage{simpleicons}

\newcommand{\crgdir}{images/neurips_2026/crg_examples}

\newtcolorbox{crgbox}[1]{
  enhanced,
  colback=white,
  colframe=black!60,
  boxrule=0.6pt,
  arc=1.5pt,
  left=5pt,
  right=5pt,
  top=2pt,
  bottom=2pt,
  fonttitle=\bfseries\fontsize{8}{9.5}\selectfont,
  coltitle=black,
  colbacktitle=black!8,
  title={#1},
  equal height group=crgexamples
}
\newcommand{\imcell}[2]{%
  \begin{minipage}{2cm}
    \centering
    \includegraphics[width=2cm]{\crgdir/#1}\\[1pt]
    {\normalsize\textcolor{black!55}{#2}\par}
  \end{minipage}%
}

\newcommand{\crgexample}[7]{%
\begin{crgbox}{#2}
  {\centering
   \imcell{#1-original.pdf}{source}\hfill
   \imcell{#1-gt.pdf}{oracle (GT)}\hfill
   \imcell{#1-tddn.pdf}{TDDN}\par}
  \vspace{2pt}\hrule\vspace{2pt}
  \normalsize\raggedright
  \parbox[t]{\linewidth}
  {%
    \raggedright\strut\textcolor[HTML]{1E7B34}{\textbf{Q.}}\,#3\strut
  }\par
  \vspace{1pt}
  \textbf{InternVL3-8B:}\ raw #4
  \textbullet\ oracle #5
  \textbullet\ TDDN #6
  \textbullet\ GT #7
\end{crgbox}\vspace{3pt}%
}

\title{\tmethod{}: Text-aligned Diffused DINO Network for Puzzle Understanding}

\author{\name Harsha Patnala\thanks{These authors contributed equally.} \email pshanmukhasreeharsha@gmail.com \\
        \addr Eightfold AI
        \AND
        \name Debopriyo Banerjee\footnotemark[1] \email deb.ban89@gmail.com \\
        \addr Inception42
        \AND
        \name Ayush Sunil Munot \email munotayush6@gmail.com \\
        \addr Indian Institute of Technology Kharagpur
        \AND
        \name Somak Aditya \email saditya@cse.iitkgp.ac.in \\
        \addr Indian Institute of Technology Kharagpur}

\begin{document}

\maketitle

\setcounter{topnumber}{5}
\setcounter{totalnumber}{5}
\renewcommand{\topfraction}{0.95}
\renewcommand{\textfraction}{0.05}
\makeatletter
\setlength{\@fptop}{0pt}
\makeatother

\begin{abstract}
Structured visual reasoning, such as image puzzles, demands fine-grained visual perception, which current Vision Language Models (VLMs) often lack. VLMs built on CLIP-based ViT backbones tend to trade fine-grained visual details for high-level semantics. We show that this deficit propagates downstream, affecting various vision and language tasks. To address this, we fuse DINOv3 and CleanDIFT representations into a perception encoder (\method{}), and align it with RoBERTa-L, yielding a \textbf{text-aligned model (\tmethod{})} that preserves this perceptual advantage:
with frozen backbones and only ${\sim}$590K alignment pairs, {\tmethod{} surpasses CLIP ViT-L/14 in image--text retrieval tasks: text-to-image on Flickr30K and both directions on MS-COCO-2014.} TDDN more than triples CLIP's dense-prediction accuracy (ADE20K $5.20$ $\rightarrow$
$18.11$ mIoU, COCO-Stuff $7.35$ $\rightarrow$ $24.44$), despite CLIP's massive training corpus.
\tmethod{} leads on segmentation benchmarks among general-purpose contrastive
encoders, including SigLIP\,2. We further introduce \PuzzlePerception{}, a
segmentation and visual question answering dataset that probes fine-grained
spatial understanding, on which \tmethod{}
doubles CLIP's segmentation accuracy (11.04 $\to$ 22.51 mIoU). Our code and data are available at 
\url{https://github.com/adityaSomak/TDDN}.
\end{abstract}

\section{Introduction}

Weak visual perception is a central bottleneck for Vision Language Models~(VLMs) on structured visual reasoning tasks such as maze navigation, checker moves, Sudoku, and Raven's Progressive Matrices~\citep{chia2024puzzlevqa,zhang2019raven,ren2025vgrp}, where success requires fine-grained visual reasoning.

Part of this weakness originates in the vision encoder. Most VLMs use CLIP-based ViT backbones~\citep{clip} whose global contrastive objective trades local spatial detail for semantic alignment. \citet{tong:2024:eyeswideshut} demonstrate this limitation via ``CLIP-blind pairs'' using the MMVP benchmark--images CLIP encodes as near-identical despite \textit{visible} differences. On these pairs, VLMs such as GPT-4V~\citep{openai2023gpt4v} and Gemini~\citep{google2023gemini} hallucinate on orientation, counting, and viewpoint. Scaling ViT capacity and training data leaves \textit{seven} of \textit{nine} failure patterns unresolved~\citep{tong:2024:eyeswideshut}, indicating an architectural bottleneck. 

Solving visual puzzles therefore requires addressing the underlying perception problem rather than relying on stronger language reasoning. CLIP lacks mechanisms for encoding spatial structure, sharp boundaries, and patch-level consistency at the granularity that puzzles demand. This calls for a representation combining the semantic coherence of discriminative self-supervised learning with the boundary precision of generative models. \citet{taleoftwo} show that Stable Diffusion features complement DINOv2: self-supervised features provide globally coherent semantics while diffusion features resolve fine spatial boundaries. Their simple fusion outperforms either representation alone. LaViDa~\citep{li:2025:lavida} further shows that diffusion-derived features are strong for visual understanding, indicating that generative backbones encode fine-grained perceptual signals that discriminative encoders fail to capture. Building on these observations, we ask whether a fused representation
that is \textit{superior} at fine-grained vision can also align with language, allowing a VLM to inherit its perceptual advantage. 


We construct \method{} by fusing noise-free
CleanDIFT~\citep{cleandift} diffusion features, which capture fine spatial boundaries,
with globally coherent self-supervised tokens of 
DINOv3~\citep{dinov3}. The design targets CLIP's fine-grained failure modes
while retaining this dense-perception advantage after language
alignment under a small data and compute budget. The two feature sources play complementary roles in this construction: CleanDIFT provides the local information needed to distinguish boundaries, while DINOv3 provides the semantic organization needed to relate patches to coherent objects. {Our aim is to combine both properties in a single representation and preserve them through text alignment to improve visual puzzle understanding by helping VLMs identify individual puzzle elements and interpret their spatial relationships.}

{We align \tmethod{} with language using image--caption pairs from datasets such as LAION-5B~\citep{beaumont2022laion5b}.} However, the image captions available in LAION-5B are often noisy and of limited quality~\citep{recap}. LAION-COCO \citep{laion_relaion_coco} provides a recaptioned version of a subset of LAION-5B, while LAION-COCO-Aesthetic further filters these image--text pairs based on aesthetic and quality criteria~\citep{li_laion_coco_aesthetic}. However, the resulting captions are typically short and do not capture the complete visual content of the images. To address this limitation, we recaption a subset of LAION-COCO-Aesthetic to obtain Recaptioned-LAION-COCO-Aesthetic (ReLAION-COCOA), a set of richer and more comprehensive image--text pairs. Our recaptioning step improves the descriptive aspect of the image-text supervision: aesthetic filtering selects images, whereas richer captions describe more of the visual content within those images. It therefore complements our choice of a perception-oriented encoder by providing more comprehensive language descriptions for alignment, while keeping the paired training corpus small.
 Using frozen backbones and only
${\sim}590$K pairs, we \textit{text-align} the encoder to produce
\tmethod{}, which matches CLIP (trained on 400M pairs) on image-text
retrieval while surpassing it on dense prediction tasks, including
open-vocabulary segmentation and keypoint matching.

Our results show that complementary visual features improve spatial coherence while preserving fine-grained boundaries across diverse visual structures. Moreover, \tmethod{}-predicted regions provide actionable spatial signals that improve puzzle-based visual question answering in frozen VLMs without weight updates. These evaluations connect representation quality to its downstream use: dense prediction tests whether the encoder retains local distinctions, and region-guided question answering tests whether those distinctions can 
support visual reasoning. In this way, the same perceptual advantage is examined both within the learned visual representation and through the spatial guidance it provides to an existing VLM.

\paragraph{Key Contributions:}
\begin{itemize}

  \item \textbf{Text-aligned \method{}.}
  Our central contribution is \tmethod{}, a text-aligned vision encoder designed to preserve fine-grained perceptual capabilities while maintaining strong vision-language alignment.
  We show that \method{}'s features can be aligned with language using frozen backbones and only ${\sim}590$K alignment pairs, without sacrificing their dense-perception advantage; \tmethod{} matches CLIP on image-text retrieval while substantially surpassing it on dense prediction. We further show that regions predicted by \tmethod{}, used as contrastive region guidance~(CRG), improve a frozen VLM's puzzle-solving capability, translating this perceptual advantage into downstream reasoning gains.

  \item \textbf{Representational analysis.}
  Through a suite of representational analyses at both the patch and global levels, we show that the two feature sources of \method{} are complementary: fusion increases patch-level discriminability while preserving global semantics. This dual property helps explain both the vision encoder's gains on dense tasks and its amenability to text alignment. Examining both granularities makes explicit how local spatial information and global semantic organization contribute to the fused representation.
  \item \textbf{Puzzle Perception dataset.}
We present \PuzzlePerception{}, a visual-puzzle segmentation dataset for evaluating the fine-grained spatial understanding capabilities of existing VLMs. Through this benchmark, we show the effectiveness of TDDN to improve fine-grained recognition performance on visual puzzle images. The dataset complements natural-image evaluation with structured puzzle environments, linking the assessment of detailed visual perception to puzzle-based question answering.
\end{itemize}

\section{Related Work}
Vision-language models including GPT-4V~\citep{openai2023gpt4v}, Gemini~\citep{google2023gemini}, LLaVA~\citep{liu2024llava}, InternVL~\citep{chen2024internvl}, Gemma~\citep{gemma3}, and Qwen-VL~\citep{bai2025qwen25vl} perform strongly across multimodal tasks, yet fine-grained spatial understanding remains a persistent weakness across architectures~\citep{chia2024puzzlevqa,tong:2024:eyeswideshut,xu2025lvlmehub}. Our work builds on four strands: (i) benchmarks exposing spatial failures, (ii) representations addressing encoder limitations, (iii) low-resource vision--language alignment, and (iv) datasets for training and evaluating visual understanding.

\paragraph{Spatial Understanding Failures.}
Visual puzzle benchmarks expose spatial understanding limitations in modern VLMs, a prerequisite for downstream reasoning. GPT-4V achieves around 46\% accuracy on single-concept puzzles~\citep{chia2024puzzlevqa}, while large-scale evaluations reveal persistent \textbf{perception deficits} across model scales~\citep{xu2025lvlmehub}. These limitations appear in jigsaw reconstruction~\citep{lyu2025jigsawpuzzles}, 
grid-based logic~\citep{ren2025vgrp}, and multi-step planning settings~\citep{ghosal2025algopuzzlevqa}. Across these domains, models remain below human performance~\citep{jiang2024marvel,zou2024dynamath}, suggesting that spatial failures propagate to higher-level reasoning. These findings point to a shared upstream bottleneck---the visual encoder---motivating our focus on visual representation quality.

\paragraph{Visual Encoders.}
CLIP~\citep{clip}, the dominant vision encoder for VLMs, optimizes global image--text alignment but lacks fine-grained spatial supervision, leading to failures such as ``CLIP-blind pairs''~\citep{tong:2024:eyeswideshut}. Prior work addresses this via hard negatives~\citep{patel2024tripletclip}, local alignment~\citep{chen2025lazsl}, diffusion-guided augmentation~\citep{Luo:2025:deem_iclr}, and auxiliary segmentation encoders~\citep{jain2024vcoder}, but these largely compensate for limitations of the objective itself. Self-supervised encoders such as DINOv2/v3~\citep{dinov2,dinov3} improve patch-level semantics, while diffusion features provide complementary dense geometric correspondences~\citep{dift,cleandift}. Since DINO emphasizes global semantics and diffusion features capture local geometry~\citep{taleoftwo}, we combine CleanDIFT and DINOv3 in a unified encoder.

\paragraph{Vision-Language Alignment in low-resource condition.}
In low-data regimes, visual encoder quality largely constrains multimodal performance. Methods such as CuPL~\citep{cupl} and Tip-Adapter~\citep{tipadapter} improve transfer from frozen CLIP models without supervision, but remain limited by encoder capacity. BLIP-2~\citep{li2023blip2bootstrappinglanguageimagepretraining} showed that coupling a frozen encoder to a frozen LLM via a lightweight Q-Former can rival much larger end-to-end systems, emphasizing the importance of strong frozen representations. Further evidence comes from low-data settings: STRUCTURE~\citep{structure} preserves latent geometry to achieve over 50\% gains using less than 1\% paired data, while DINOv2-text~\citep{dinotxt} and Talk2DINO~\citep{barsellotti2025talkingdinobridgingselfsupervised} align language to frozen self-supervised backbones without backbone tuning. Motivated by this encoder-ceiling effect, we follow LiT~\citep{lit} and freeze the visual backbones (DINOv3, CleanDIFT) and RoBERTa text encoder, training only lightweight attention and projection layers to exploit richer pretrained representations while preserving their structure.

\paragraph{Datasets for Visual Understanding.}
Semantic segmentation benchmarks such as COCO-Stuff~\citep{cocostuff} and ADE20K~\citep{ade20k} provide pixel-level annotations for learning fine-grained visual representations, but lack question answering for evaluating reasoning over these representations. Conversely, reasoning datasets such as CLEVR~\citep{clevr} and GQA~\citep{gqa} emphasize compositional question answering over objects, attributes, and relations without dense visual supervision. Puzzle-oriented datasets, including PuzzleVQA~\citep{chia2024puzzlevqa} and AlgoPuzzleVQA~\citep{ghosal2025algopuzzlevqa}, extend visual reasoning to structured domains requiring spatial and algorithmic reasoning, but lack pixel-level annotations. This separation limits jointly training and evaluating fine-grained perception and downstream reasoning within a common benchmark. Our \PuzzlePerception{} addresses this gap by combining pixel-level segmentation and visual question answering, enabling perceptual and structured reasoning in puzzle environments.

\section{Methodology}
\label{sec:visual_perception}
\begin{figure}[!t]
  \centering
  \includegraphics[width=0.6\linewidth]
    {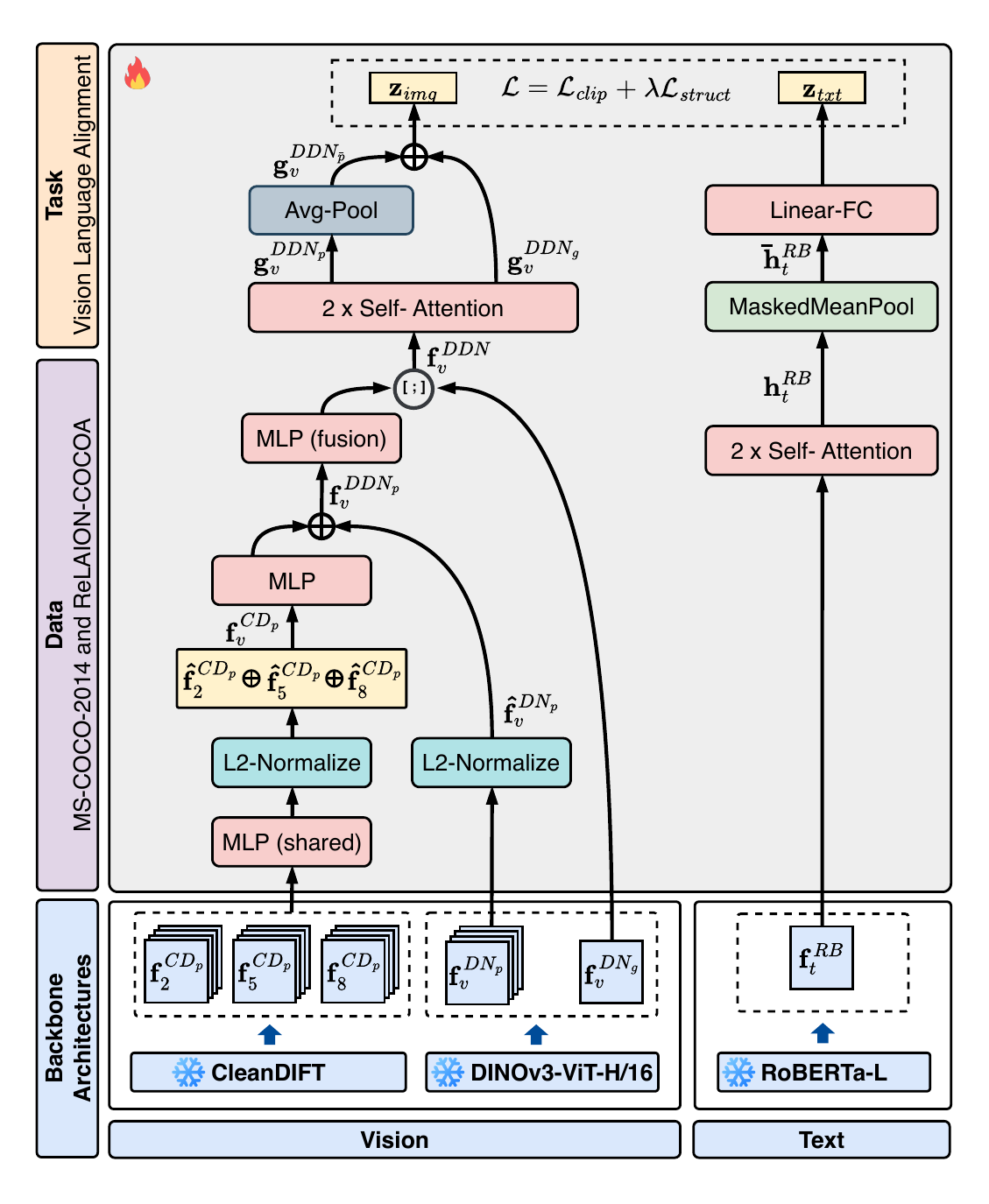}
  \caption{
  Overview of the Text-aligned \method~(\tmethod{}) architecture,
  illustrating its key components.
  }
  \label{fig:tddn_architecture}
\end{figure}

Visual puzzles demand representations that are both spatially fine-grained and semantically coherent at the object level. Building on evidence that fusing diffusion and self-supervised features beats either alone~\citep{taleoftwo}, we pair noise-free CleanDIFT (CD) features with the stronger DINOv3-H/16+ (DN) backbone to form \method{} (\methodshort{}), and ask whether we can tie it to language without losing its edge over CLIP on fine-grained perception. We present the vision-language alignment in Fig.~\ref{fig:tddn_architecture}, that yields our proposed text-aligned version of \methodshort{}, which we name \textbf{\tmethod{}}.

\subsection{Vision-Language Alignment}
\label{sec:text_align}
Aligning a perception-optimized encoder to text risks eroding the very fine-grained
sensitivity that motivates it; we design \tmethod{} to add language grounding while
protecting that sensitivity. \tmethod{} pairs \methodshort{} as the vision encoder and 
RoBERTa-L~\citep{roberta} as the text encoder. We combine the architectural recipe of
\citet{dinotxt} with the data-efficient loss of \citet{structure}, keeping all backbone
weights frozen and training only lightweight heads, multi-layer perceptron (MLP), and attention blocks on
top (Table~\ref{tab:frozen_trainable}, Appendix~\ref{app:vision-text_alignment}). 
We use
$\mathbf{z}_{\mathrm{img}}$ and $\mathbf{z}_{\mathrm{txt}}$ to denote generic image
and text embeddings, respectively. The two branches share an embedding space but operate on differently structured inputs: spatial patch tokens and a global token on the vision side, and a sequence of text tokens on the language side. The trainable modules connect these pretrained representations without requiring either backbone to relearn feature extraction.

\paragraph{Fused Vision Encoder.}
\methodshort{} fuses two frozen, pretrained backbones with complementary patch-level representations. DINOv3 final-layer patch tokens provide semantically organized, patch-consistent features, supported by Gram anchoring~\citep{dinov3}, which preserves pairwise patch-similarity structure during long training. In contrast, CleanDIFT intermediate patch features capture finer spatial detail and object boundaries without the noise or timestep tuning required by standard diffusion features. Their fusion therefore combines DINOv3's semantic structure with CleanDIFT's spatial precision, yielding a representation that preserves high-level organization while retaining the local detail needed for fine-grained visual understanding. Freezing both backbones preserves these pretrained properties while restricting learning to the fusion mechanism.

We pass each image through the frozen CleanDIFT UNet backbone and extract intermediate
activations from decoder layers $\ell \in \{2, 5, 8\}$ (best-performing combination
according to the ablation in 
Table~\ref{tab:cleandift_layers_spair}, Appendix~\ref{app:feature_extraction_fusion}), denoted as
$\mathbf{f}^{\text{CD}}_{l} \in \mathbb{R}^{N^2 \times d_\ell}$ with
$(d_2, d_5, d_8) = (1280, 640, 640)$. We spatially interpolate each layer's activation map to a shared $N \times N$ grid. We independently project each layer's features to a common dimension, $d_c = 512$, using a trainable MLP layer $f_l: \mathbb{R}^{d_l}\rightarrow \mathbb{R}^{d_c}$.
\begin{equation}
\label{eq:cd_layers}
\mathbf{f}^{\text{CD}_{p}}_{l} = f_l(\mathbf{f}^{\text{CD}_p}_{\,\ell}) \in \mathbb{R}^{N^2 \times d_c}
\end{equation}
We $\ell_2$-normalize each of $\mathbf{f}^{\text{CD}_{p}}_{l}$ and concatenate along the channel axis to form
\begin{equation}
\label{eq:cd_concat}
\mathbf{f}^{\text{CD}_{p}}_{v} =
  \hat{\mathbf{f}}^{\text{CD}_{p}}_{\,2} \oplus
  \hat{\mathbf{f}}^{\text{CD}_{p}}_{\,5} \oplus
  \hat{\mathbf{f}}^{\text{CD}_{p}}_{\,8}
  \in \mathbb{R}^{N^2 \times 3 d_c},
\end{equation}
where $\hat{\cdot}$ denotes $\ell_2$-normalization, $\oplus$ denotes concatenation, and $3 d_c = 1536$.


Let
$\mathbf{f}^{\text{CD}_p}_{v} \in \mathbb{R}^{N^2 \times {3d_c}}$ and
$\mathbf{f}^{\text{DN}_p}_{v} \in \mathbb{R}^{N^2 \times d_n}$ denote the CleanDIFT
and DINOv3 patch features respectively (where $N^2$ is the number of patches), and let
$\mathbf{f}^{\text{DN}_g}_{v} \in \mathbb{R}^{d_n}$ denote the frozen DINOv3 global CLS
token. We first project the CleanDIFT patch features with a trainable MLP $f_{\mathrm{e}}(\cdot)$ and
concatenated with the $\ell_2$-normalized DINOv3 patch features to form the fused
patch representation:
\begin{equation}
    \mathbf{f}^{\text{DDN}_p}_{v} = \alpha f_{\mathrm{e}}(\mathbf{f}^{\text{CD}_p}_{v})
    \oplus (1-\alpha) \hat{\mathbf{f}}^{\text{DN}_p}_{v}
    \in \mathbb{R}^{N^2 \times (3d_c + d_n)}
\end{equation}


We pass the fused patch features through a trainable fusion MLP
$f_{\mathrm{fuse}} : \mathbb{R}^{3d_c + d_n} \rightarrow \mathbb{R}^{d_n}$
and combined with the DINOv3 global CLS token vector $\mathbf{f}^{\mathrm{DN}_g}_v\in \mathbb{R}^{d_n}$:
\begin{equation}
    \mathbf{f}^{\text{DDN}}_v =
    \left[
    \mathbf{f}^{\text{DN}_g}_{v},
    f_{\mathrm{fuse}}\!\left(\mathbf{f}^{\text{DDN}_p}_{v}\right)
    \right]
    \in \mathbb{R}^{(N^2+1) \times d_n}.
\end{equation}

\begin{samepage}
We refine $\mathbf{f}^{\mathrm{DDN}}_v$ by $2$ stacked trainable self-attention blocks $g_{\mathrm{SA}}^{(2)}(\cdot)$.
The output is split into a CLS token and patch tokens; the patch tokens are aggregated
via average pooling to obtain $\mathbf{f}^{\text{DDN}}_{\bar{p}}$, which is
concatenated with the CLS token to produce the final image embedding:

\begin{equation}
    \left[\mathbf{g}^{\text{DDN}_g}_{v},\, \mathbf{g}^{\text{DDN}_p}_{v}\right] = g_{\mathrm{SA}}^{(2)}(\mathbf{f}^{\text{DDN}}_{v})
\end{equation}

\begin{equation}
    \mathbf{g}^{\text{DDN}_{\bar{p}}}_v = \mathrm{AvgPool}(\mathbf{g}^{\text{DDN}_p}_{{v}}) \in \mathbb{R}^{d_n}
\end{equation}

\begin{equation}
    \mathbf{z}_{\mathrm{img}} = \mathbf{g}^{\text{DDN}_g}_{v} \oplus \mathbf{g}^{\text{DDN}_{\bar{p}}}_v \in \mathbb{R}^{2d_n}
\end{equation}
We denote patch and global features by the suffixes $p$ and $g$, respectively, for CD and DDN representations.

The image embedding retains the refined CLS token and the average of the refined patch tokens as separate $d_n$-dimensional components. Concatenation yields the $2d_n$ dimensions used for global image--text comparison, while the unpooled patch tokens remain available for dense prediction at individual spatial locations.
\end{samepage}

\paragraph{Text encoder.}

We use RoBERTa-L~\citep{roberta} as the frozen text backbone. Let $x_{\mathrm{txt}}$ 
denote an input text, and let 
$\mathbf{f}^{\mathrm{RB}}_t = \mathcal{E}(x_{\mathrm{txt}}) \in \mathbb{R}^{L \times d_t}$ 
denote the frozen final-layer token representations for the input $x_{\mathrm{txt}}$, 
comprising $L$ tokens, where $d_t$ denotes the dimension of each token and 
$\mathcal{E}(\cdot)$ denotes the RoBERTa-L encoder. 
We refine the token representations with $2$ stacked trainable self-attention blocks:
\begin{equation}
    \mathbf{h}^{\mathrm{RB}}_t = h_{\mathrm{SA}}^{(2)}(\mathbf{f}^{\mathrm{RB}}_t) \in \mathbb{R}^{L \times d_t},
\end{equation}
where $h_{\mathrm{SA}}^{(2)}$ denotes two successive self-attention layers. The refined 
tokens are aggregated via masked mean pooling over the sequence dimension. Only valid text positions contribute to this average, excluding padding introduced when sequences have different lengths. The pooling step therefore produces one fixed-dimensional representation per caption:
\begin{equation}
    \bar{\mathbf{h}}^{\mathrm{RB}}_{t} = \mathrm{MaskedMeanPool}(\mathbf{h}^{\mathrm{RB}}_t) \in \mathbb{R}^{d_t}
\end{equation}
We then pass $\bar{\mathbf{h}}^{\mathrm{RB}}_{t}$ through a trainable linear projection of matching output dimension $W_t \in \mathbb{R}^{2d_n \times d_t}$, resulting in the final text embedding:
\begin{equation}
    \mathbf{z}_{\mathrm{txt}} = W_t\, \bar{\mathbf{h}}^{\mathrm{RB}}_t \in \mathbb{R}^{2d_n}
\end{equation}

\paragraph{Training Objective.}
We adopt standard alignment components to isolate the contribution of the fused representation. Following the minimal recipe of \citet{structure}, we hold the objective, text encoder, and data fixed across alignment conditions, varying only the vision encoder. This controlled setup attributes gains to visual representation quality rather than a stronger objective, text backbone, or condition-specific optimization, while retaining direct comparability with prior work. Fusion receives no additional supervision unavailable to the single-encoder baselines. We minimize the InfoNCE loss $\mathcal{L}_{\text{clip}}$~\citep{clip} over image--text pairs, combined with the STRUCTURE regulariser of \citet{structure}, which preserves the pairwise geometry of pretrained latent spaces and prevents collapse under low-data conditions:

\begin{equation}
\label{eq:total_loss}
\mathcal{L} = \mathcal{L}_{\text{clip}} + \lambda\,\mathcal{L}_{\text{struct}}
\end{equation}
Here, $\lambda$ controls the trade-off between adapting the representation toward the
language space and retaining the relational structure inherited from pretraining.
This formulation permits the visual features to move sufficiently to establish
cross-modal correspondences while constraining distortions that would erase useful
neighbourhood relationships. The resulting optimization therefore treats alignment as
a geometry-preserving adaptation problem rather than unconstrained representation
relearning. $\mathcal{L}_{\text{clip}}$ supplies language grounding, while
$\mathcal{L}_{\text{struct}}$ preserves the pretrained geometry that carries fine-grained
structure, yielding text alignment without perceptual regression. 
We retain InfoNCE rather than adopting the sigmoid objective of SigLIP~\citep{Zhai:2023:SigLIP}, keeping the alignment objective fixed across all conditions to isolate the contribution of the visual representation. We use RoBERTa~\citep{roberta} as a frozen, standard off-the-shelf text encoder across all alignment conditions, ensuring that gains do not result from adopting a more powerful text backbone.
We anchor the regulariser on frozen DINOv3 features, whose CLS
token achieves the strongest hypersphere coverage and class-level separability among
all representations (\S~\ref{sec:rep_results}); DINOv3 thus anchors the geometry while
CleanDIFT contributes through the learned fusion path.

%


\section{Datasets}
\label{sec:datasets}

{We train and evaluate TDDN on datasets spanning vision--language alignment, segmentation, classification, keypoint matching, image--text retrieval, and puzzle-based VQA. Image--caption corpora provide supervision for aligning visual and textual embeddings, while downstream benchmarks assess the resulting representations at both global and local scales. Natural-image datasets cover diverse object categories and scenes, and our \PuzzlePerception{} dataset tests fine-grained spatial understanding in structured puzzle environments. Our evaluations examine different outputs of the same representation: global embeddings support retrieval and classification, patch embeddings support dense prediction and correspondence, and predicted regions supply spatial guidance for puzzle-based question answering. Together, these settings test whether the representation retains semantic information while preserving the local visual distinctions needed for structured understanding. Table~\ref{tab:puzzle_perception_dataset} summarizes the tasks and splits of \PuzzlePerception{}.}


\subsection{Existing Datasets}
\paragraph{Training corpora.}
Following the data-efficient regime of \citet{structure}, we train the text-alignment module on $\sim$590k image--caption pairs. These comprise the \textbf{82k-image MS-COCO-2014} training set with its original captions~\citep{coco}, and \textbf{Recaptioned-LAION-COCO-Aesthetic (ReLAION-COCOA)}, a $\sim$\textbf{508k-image subset of LAION-COCO-Aesthetic}~\citep{li_laion_coco_aesthetic}. We recaption the latter using Gemma-3-27B-it~\citep{gemma3}, generating a single high-quality, visually descriptive caption per image to replace the original noisy web-derived text. Our training scale follows the low-resource setting of \citet{structure}, rather than CLIP-scale pretraining. Since large-scale recaptioning is computationally expensive, we leave scaling ReLAION-COCOA and exploring more effective data-selection strategies to future work.



%


\paragraph{Evaluation benchmarks.}
We evaluate on widely used public benchmarks spanning \textbf{image classification}: Food-101~\citep{food101}, CIFAR-100~\citep{cifar100}, Caltech-101~\citep{caltech101}, GTSRB~\citep{gtsrb}, and ImageNet-1k~\citep{imagenet}; \textbf{image--caption retrieval}: Flickr30K~\citep{flickr30k} and MS--COCO--2014~\citep{coco}; \textbf{segmentation}: ADE20K~\citep{ade20k}, Cityscapes~\citep{cityscapes}, COCO-Stuff~\citep{cocostuff}, and PASCAL-Context~\citep{pascalcontext}; and \textbf{keypoint matching}: SPair-71k~\citep{spair71k}. We use the official splits; dataset statistics are provided in Table~\ref{tab:eval_datasets} of Appendix~\ref{app:additional_dataset_details}.

\begin{table}[!t]
\centering
\caption{Statistics of the \PuzzlePerception{} dataset, comprising four structured visual reasoning tasks designed for both dense segmentation and puzzle-based visual question answering (PVQA). Under the PVQA task, the Chess and N-Queens puzzles have eight and four questions, respectively.}
\label{tab:puzzle_perception_dataset}

\setlength{\tabcolsep}{5pt}

\begin{tabular*}{\linewidth}{
  @{\extracolsep{\fill}}lcccccc@{}
}
\toprule
\textbf{Puzzle} & \textbf{Seg} & \textbf{PVQA} & \textbf{Cls}
& \textbf{Train} & \textbf{Validation} & \textbf{Test} \\
\midrule
Maze           & \cmark & \xmark & 8  & 2{,}000 & 500 & 500 \\
Chess          & \cmark & \cmark & 15 & 2{,}000 & 500 & 500 \\
Tower-of-Hanoi & \cmark & \xmark & 7  & 2{,}000 & 500 & 500 \\
N-Queens       & \xmark & \cmark & 2  & --      & --  & 100 \\
\bottomrule
\end{tabular*}
\end{table} 

\subsection{Puzzle Perception}
Benchmarks such as ADE20K~\citep{ade20k}, Cityscapes~\citep{cityscapes} and COCO-Stuff~\citep{cocostuff} emphasize coarse natural-image semantics, under-testing fine-grained structured perception. 
We introduce \PuzzlePerception{}, a dense segmentation and puzzle-based visual question-answering (PVQA) dataset covering \textit{Maze}, \textit{Chess}, \textit{Tower-of-Hanoi}, and \textit{N-Queens} across diverse structured visual domains. \textit{Maze}, \textit{Chess} and \textit{Tower-of-Hanoi} support segmentation; PVQA targets the spatially and compositionally complex \textit{Chess} and \textit{N-Queens}. Images are $512{\times}512$ with 30 pixel-level classes. Dataset splits are 2,000/500/500 train/validation/test samples (seed 42) per puzzle, except \textit{N-Queens}, which has 100 test-only samples (Table~\ref{tab:puzzle_perception_dataset}). We derive segmentation masks directly from the rendering pipeline and vary visual styles and puzzle configurations, enabling controlled evaluation of spatial representations without ambiguity from manual annotation. See Appendix~\ref{app:additional_dataset_details} for details.

\section{Experiments}
\label{sec:experiments}
We structure our investigation around the following research questions:

\textbf{RQ1:} Is our proposed visual representation more informative and discriminative compared to standard ViT/CLIP backbones?

\textbf{RQ2:} Can downstream vision--text alignment retain the gains from the new visual perception backbone?

\textbf{RQ3:} Does the proposed visual perception backbone support stronger visual puzzle understanding?

\begin{table}[!t]
\centering
\caption{\textbf{Comparison of $\mathrm{DDN}'$, TDN and TDDN with
baselines on downstream visual perception tasks.}
Seg: segmentation; KPM: keypoint matching; Cls: classification.
Higher ($\uparrow$) is better.}
\label{tab:quantitative_results}

\setlength{\tabcolsep}{2.8pt}

\begin{tabularx}{\linewidth}{
  >{\raggedright\arraybackslash}X
  >{\centering\arraybackslash}p{0.15\linewidth}
  >{\centering\arraybackslash}p{0.15\linewidth}
  >{\centering\arraybackslash}p{0.15\linewidth}
}
\toprule
\multirow{2}{*}{\textbf{Method}}
& \textbf{Seg}
& \textbf{KPM}
& \textbf{Cls} \\
& (mIoU$\uparrow$)
& (PCK@0.1$\uparrow$)
& (top-1$\uparrow$) \\
\midrule

\rowcolor{gray!15}
\multicolumn{4}{l}{\textit{Vision-only}} \\

DINOv2-B/14 \citep{dinov2}
& 75.79 & 55.44 & 78.22 \\

DINOv2-G/14 \citep{dinov2}
& 78.80 & 55.72 & 79.29 \\

$\mathrm{SD}'$ \citep{sd21}
& 74.77 & 56.46 & 12.54 \\

$\mathrm{SD}'$ + DINOv2-B/14 \citep{taleoftwo}
& 76.05 & 60.08 & 77.29 \\

$\mathrm{SD}'$ + DINOv2-G/14 \citep{taleoftwo}
& 78.70 & 61.28 & 78.39 \\

$\mathrm{DN}$ \citep{dinov3}
& 76.18 & 58.36 & 83.11 \\

$\mathrm{CD}'$ \citep{cleandift}
& 79.20 & 61.20 & 38.74 \\

\rowcolor{gray!35}
\textbf{$\mathrm{DDN}'$ (ours)}
& \textbf{81.15}
& \textbf{64.39}
& \textbf{83.80} \\

\midrule

\rowcolor{gray!15}
\multicolumn{4}{l}{\textit{Text-aligned / vision-language}} \\

CLIP ViT-L/14 \citep{clip}
& 71.33 & 24.89 & 73.10 \\

\rowcolor{gray!35}
TDN (ours)
& 68.89 & 28.29 & 75.80 \\

\rowcolor{gray!35}
\textbf{TDDN (ours)}
& \textbf{74.70}
& \textbf{32.39}
& \textbf{76.08} \\

\bottomrule
\end{tabularx}
\end{table}

\subsection{Representation Quality Analysis}
We compare \tmethod{} with CLIP-ViT-L/14~\citep{clip}, DINOv2~\citep{dinov2}, DINOv3~\citep{dinov3}, vanilla Stable Diffusion~2.1 (SD)~\citep{sd21}, and CleanDIFT~\citep{cleandift}, together with text-aligned DINOv3 (TDN), an ablation of \methodshort{} (see Appendix~\ref{app:tdn_details}), through qualitative and quantitative representation analysis across three downstream tasks: segmentation, keypoint matching, and classification.

\paragraph{Tasks and Metrics.}
We evaluate frozen visual representations on the following three tasks: 

(i) \textbf{Segmentation} (Seg) uses identical linear probes on frozen patch features over \PuzzlePerception{}, reporting mIoU. 

\noindent
(ii) \textbf{Keypoint Matching} (KPM) over SPair-71k~\citep{spair71k}, reporting keypoint-weighted PCK@$0.1$ via zero-shot nearest-neighbor matching on bounding-box-cropped pairs.

\noindent
(iii) \textbf{Classification} (Cls) uses $k$-NN ($k{=}20$, cosine similarity) on frozen global features over ImageNet-1k~\citep{nakata2022knn}. Together, they probe patch discriminability, correspondence, and global semantic separability without backbone adaptation.

\paragraph{Feature Extraction.}
For Table~\ref{tab:quantitative_results}, we use
the frozen representations defined in Appendix~\ref{app:feature_extraction_fusion}.
At the \emph{patch level}, the prime ($'$) denotes fixed PCA-based feature
construction for diffusion representations: intermediate layers $\{2,5,8\}$
are independently projected to 512 dimensions using PCA fitted on the training
set, $\ell_2$-normalized, and concatenated, yielding
$\mathrm{CD}'$ and $\mathrm{SD}'$. For dense evaluation, we form $\mathrm{DDN}'$ by normalizing and concatenating $\mathrm{CD}'$ and final-layer DINOv3 (DN) patch tokens.
For segmentation and keypoint matching, we use the patch-level
representations. At the \emph{global level}, $\mathrm{SD}'$ and $\mathrm{CD}'$
use mean-pooled patch features, while DN uses its CLS token. For $\mathrm{DDN}'$,
the fused CD--DN patch representation is mean-pooled and concatenated with the
DN CLS token to form the global representation used for classification. TDN
and TDDN use the learned patch and global representations defined in
\S~\ref{sec:text_align} and Appendix~\ref{app:tdn_details}, without PCA.
Full equations and further extraction details are provided in
Appendix~\ref{app:feature_extraction_fusion}.

\begin{figure}[!t]
  \centering
  \begin{minipage}[b]{0.39\linewidth}
    \centering
    \includegraphics[width=\linewidth]{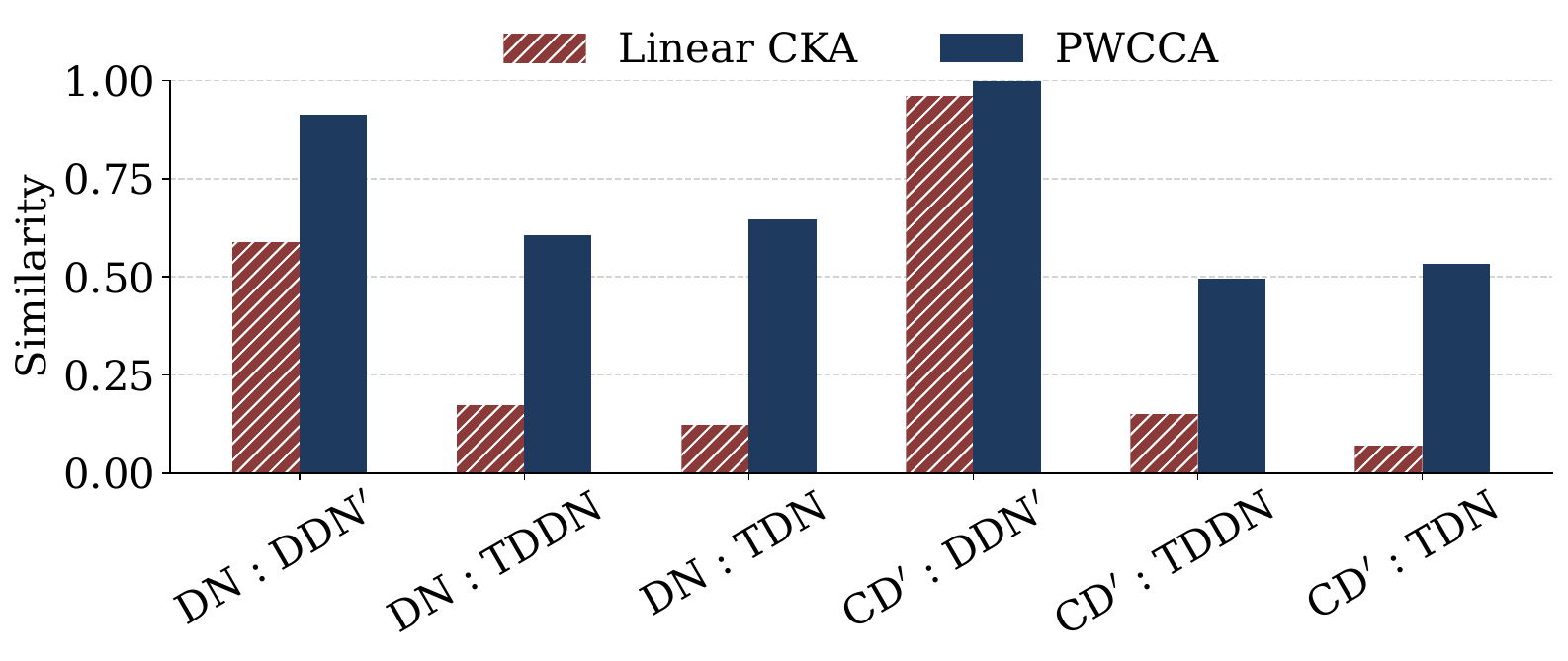}\\[2pt]
    (a) Patch similarity between encoders
  \end{minipage}
  \hfill
  \begin{minipage}[b]{0.28\linewidth}
    \centering
    \includegraphics[width=\linewidth]{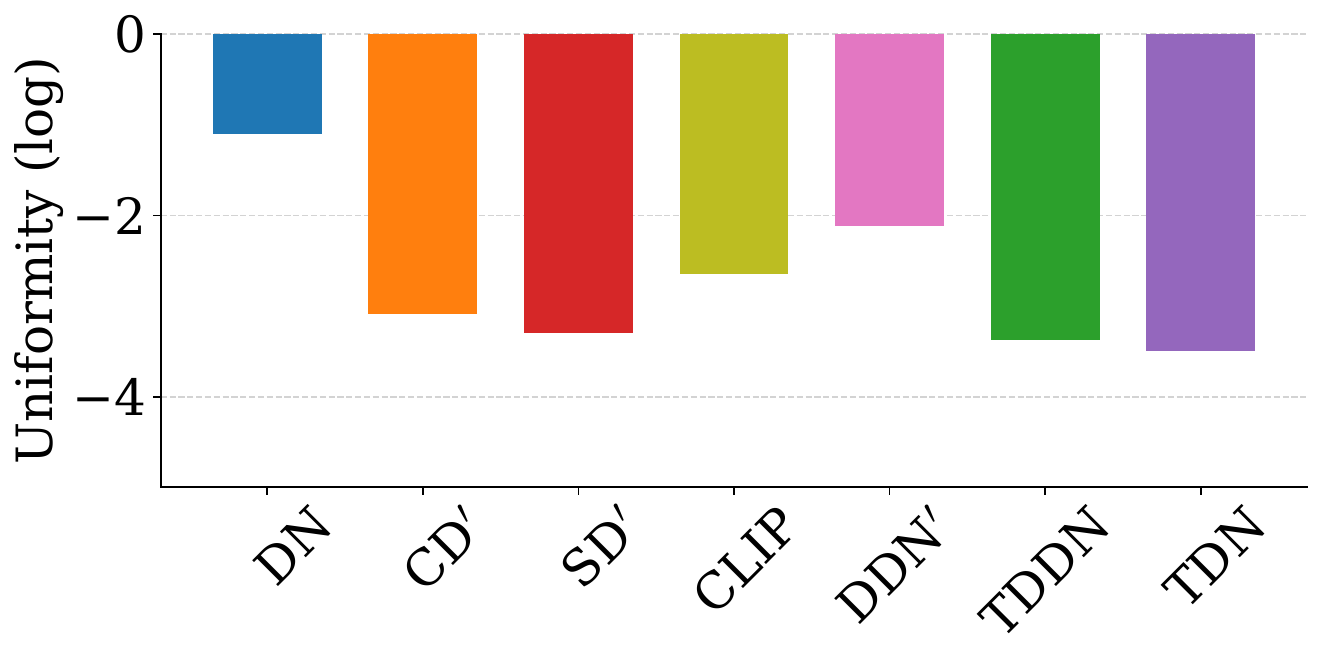}\\[2pt]
    (b) Uniformity
  \end{minipage}
  \hfill
  \begin{minipage}[b]{0.28\linewidth}
    \centering
    \includegraphics[width=\linewidth]{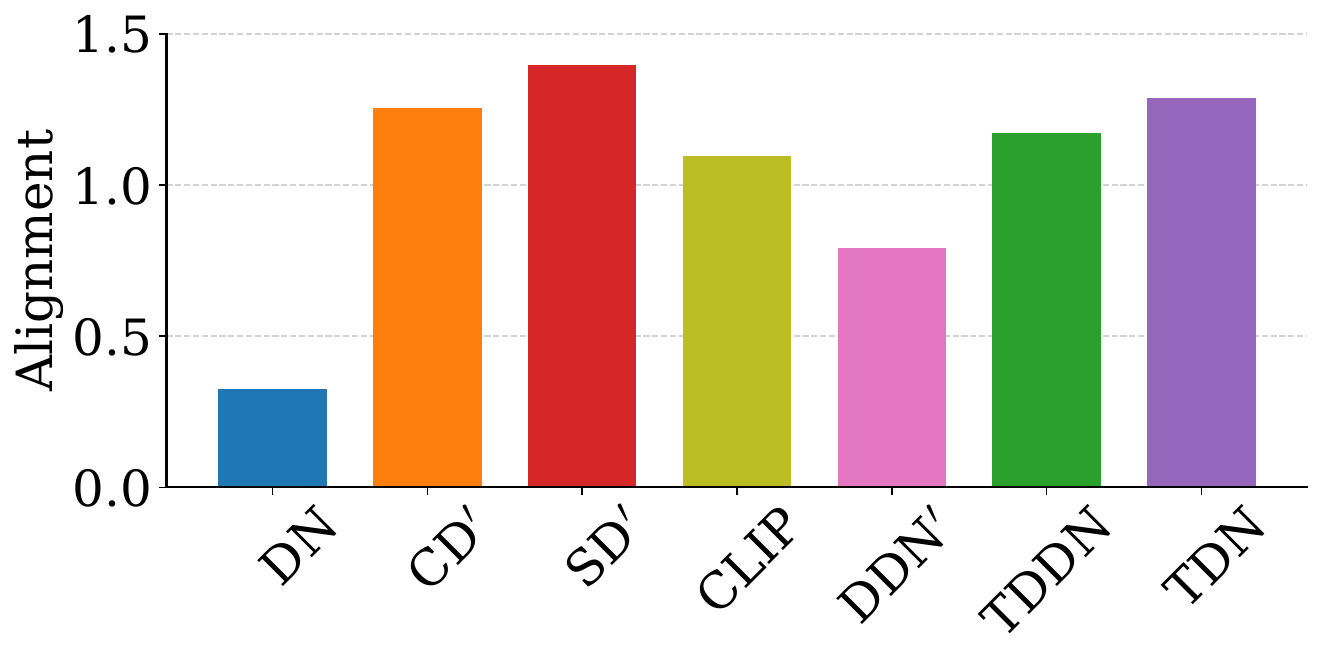}\\[2pt]
    (c) Alignment
  \end{minipage}
  \caption{\textbf{ Representational similarity and quality analysis of $\text{DDN}'$, $\text{DN}$, and $\text{CD}'$ features (Linear CKA and PWCCA on $\text{200{,}000}$ patches from $\text{2{,}000}$ images).} $X'$ denotes the PCA projection of $X \in \{\text{SD}, \text{CD}, \text{DDN}\}$. Lower $\downarrow$ is better for (b) and (c).}
  \label{fig:similarity_analysis}
\end{figure}

\paragraph{Analysis Protocol.}
For the intrinsic representation analysis in
Fig.~\ref{fig:similarity_analysis}, we sample $2{,}000$ images from the
MS-COCO-2014 validation split~\citep{coco} and use the corresponding image- and
patch-level representations defined above. For patch-level analysis, we resize
each spatial feature grid to $21\times21$ and sample the same $100$ locations
per image using a fixed seed, yielding $200{,}000$ patch tokens per representation.

\paragraph{Quantitative Results.}
\label{sec:rep_results}
Table~\ref{tab:quantitative_results} shows that $\mathrm{DDN}'$ outperforms the strongest single-encoder baseline on each of the three tasks: it improves over $\mathrm{CD}'$ by $1.95$ mIoU on segmentation and $3.19$ PCK on keypoint matching, and over $\mathrm{DN}$ by $0.69$ percentage points on classification. After vision--language alignment, \tmethod{} exceeds CLIP ViT-L/14~\citep{clip} by $3.37$ mIoU, $7.50$ PCK, and $2.98$ classification percentage points. It also improves over TDN on all three tasks, providing a direct comparison of fusion within the shared alignment pipeline. Despite competitive segmentation and keypoint matching, $\mathrm{CD}'$ and $\mathrm{SD}'$ achieve substantially lower classification accuracy ($38.74\%$ and $12.54\%$). This contrast highlights the value of evaluating both local spatial features and global semantic representations: strong dense-task performance alone does not ensure strong image-level classification. \textbf{The fused representation therefore supports spatial precision and image-level discrimination within a single feature construction.}

We use CKA and PWCCA to characterize the relationships between the source and fused representations at patch and global levels (Fig.~\ref{fig:similarity_analysis}). The source encoders exhibit low CKA ($0.22$ for $\mathrm{DN}\!:\!\mathrm{CD}'$ and $0.21$ for $\mathrm{DN}\!:\!\mathrm{SD}'$). At the patch level, $\mathrm{DDN}'$ is more similar to $\mathrm{CD}'$ (CKA~$0.96$, PWCCA~$1.00$) than to $\mathrm{DN}$ ($0.59$, $0.91$), whereas at the global level it is more similar to $\mathrm{DN}$ ($0.81$, $0.99$) than to $\mathrm{CD}'$ ($0.61$, $0.90$). Thus, \textbf{the fused representation's relative similarity to its sources changes with representational granularity.}
Because $\mathrm{DDN}'$ concatenates source features, some similarity is expected by construction; these diagnostics describe the resulting representation, while Table~2 provides evidence of the performance benefits of fusion.

Uniformity and alignment characterize within-representation geometry (lower is better for both). $\mathrm{DDN}'$ occupies an intermediate position (alignment $0.79$, uniformity $-2.12$): $\mathrm{DN}$ has lower alignment ($0.33$) but higher uniformity ($-1.10$), whereas $\mathrm{CD}'$ and $\mathrm{SD}'$ have lower uniformity ($-3.09$, $-3.30$) but higher alignment ($1.26$, $1.40$). \textbf{Fusion therefore exhibits an alignment--uniformity trade-off}, without simultaneously minimizing both measures. Following vision--language training, TDDN and TDN show improved \textbf{hypersphere uniformity}, consistent with the InfoNCE objective~\citep{wang2020uniformity}, alongside higher geometric alignment loss, a measure distinct from image--text alignment. These diagnostics use sampled MS-COCO features; performance on the distinct task datasets is established separately by downstream evaluations. The fused features thus need not optimize either geometric measure individually to improve downstream performance.

\paragraph{Key Takeaways.} Frozen-feature evaluations on segmentation, keypoint matching, and classification support RQ1.
\textbf{(i) Fusion improves vision-task performance.} CleanDIFT--DINOv3 fusion outperforms the strongest single encoder on all three tasks across patch and global representations. \textbf{(ii) The gains are consistent with useful complementarity.} Improvements over both constituents and limited cross-encoder predictability support this interpretation, while geometric diagnostics remain descriptive. \textbf{(iii) Fusion remains useful after vision--language alignment.} TDDN surpasses both TDN and CLIP on all three tasks in our low-data alignment regime. Full global-level and qualitative analyses appear in Appendices D.1 and D.5.


\subsection{Vision--Text Alignment}\label{sec:vision-text_alignment}
We examine whether a text-aligned DN+CD fusion, i.e., \tmethod{}, preserves its vision-task advantages while remaining competitive with CLIP on vision-language benchmarks.

\paragraph{Experimental Setup.}
Inputs are $336{\times}336$ with a $21{\times}21$ patch grid. Per-layer MLPs project CD layers $\{2,5,8\}$ to $512$ dimensions; their concatenation with DN patches is fused to $1280$ dimensions. Each vision/text branch uses two trainable RoPE SelfAttentionBlocks, producing $\mathbf{z}_{\text{img}},\mathbf{z}_{\text{txt}}\in\mathbb{R}^{2560}$. Baselines include CLIP ViT-L/14~\citep{clip}, OpenCLIP~\citep{cherti2023scaling}, MetaCLIP~\citep{xu2024metaclip}, DFN~\citep{fang2024dfn}, SigLIP\,2~\citep{tschannen2025siglip}, and FG-CLIP~2~\citep{xie2025fg2}. Dense prediction uses MaskCLIP except for FG-CLIP~2, whose trained dense head is an additional task-specific component.

\paragraph{Tasks.}
We evaluate \tmethod{} across four tasks: image--text retrieval, open-vocabulary segmentation, keypoint matching, and classification.

\begin{itemize}
\item \textbf{Open-vocabulary segmentation \& keypoint matching:}
We evaluate zero-shot segmentation on ADE20K~\citep{ade20k}, Cityscapes~\citep{cityscapes}, COCO-Stuff~\citep{cocostuff}, PASCAL-Context~\citep{pascalcontext}, and \PuzzlePerception{} (Maze, Chess, Tower-of-Hanoi). Each patch is assigned the class with the most similar text embedding, without segmentation-specific training. We report mIoU using the MaskCLIP~\citep{maskclip} readout for encoders without final-patch supervision (Appendix~\ref{app:ovseg}). Keypoint matching on SPair-71k~\citep{spair71k} is measured by PCK@0.1.
\item \textbf{Image--text retrieval \& classification:}
On Flickr30K~\citep{flickr30k} and MS--COCO--2014~\citep{coco}, we rank captions for each image and images for each caption by cosine similarity, reporting Recall@1 in both directions (I2T/T2I). An I2T hit requires any ground-truth caption at rank~$1$. Zero-shot classification on Food-101~\citep{food101}, CIFAR-100~\citep{cifar100}, Caltech-101~\citep{caltech101}, and GTSRB~\citep{gtsrb} matches image embeddings to class-name embeddings and reports top-$1$ accuracy.
\end{itemize}
We compare models within each task metric, since region overlap, keypoint accuracy, and retrieval rank measure different forms of correspondence.

\paragraph{Results.}
The gap between $\mathrm{TDDN}$ and CLIP widens sharply on dense prediction
(Tables~\ref{tab:matching_retrieval_results} and~\ref{tab:main_results}). \textbf{$\mathrm{TDDN}$ outperforms CLIP on every
segmentation benchmark}, notably ADE20K ($18.11$ vs.\ $5.20$ mIoU) and Puzzle
Perception ($22.51$ vs.\ $11.04$). \textbf{Gains over $\mathrm{TDN}$}
($+1.6$ ADE20K, $+8.1$ Cityscapes, $+3.8$ COCO-Stuff, $+4.1$ SPair)
\textbf{isolate CleanDIFT's contribution}. With only $\sim$$590$K pairs and frozen
backbones, $\mathrm{TDDN}$ matches or exceeds SigLIP\,2 on all five segmentation
benchmarks (e.g., Cityscapes $32.38$ vs.\ $19.98$, COCO-Stuff $24.44$ vs.\ $17.38$)
and surpasses MetaCLIP, OpenCLIP, and DFN. Only FG-CLIP\,2, with a separately trained
dense-feature head, performs better overall; $\mathrm{TDDN}$ still leads on
Cityscapes, COCO-Stuff, and SPair-71k, attributing its dense advantage to feature
fusion rather than alignment scale or task-specific heads.
Additional segmentation results on ADE20K, COCO-Stuff, and PASCAL-Context are
reported in Table~\ref{tab:main_results} (Appendix~\ref{app:ovseg_results}).

On retrieval (Table~\ref{tab:matching_retrieval_results}), \textbf{$\mathrm{TDN}$/$\mathrm{TDDN}$ outperform CLIP ViT-L/14 on three image--text retrieval tasks}: text-to-image (T2I) on Flickr30K ($72.9$/$71.2$ vs.\ $66.96$) and MS-COCO-2014 ($24.1$/$24.0$ vs.\ $18.53$), and image-to-text (I2T) on MS-COCO-2014 ($37.1$/$36.1$ vs.\ $34.60$). TDN/TDDN trail CLIP only on image-to-text (I2T) on Flickr30K ($85.8$/$85.3$ vs.\ $87.7$), despite $590$K vs.\ $400$M training pairs. SigLIP\,2~\citep{tschannen2025siglip},
DFN~\citep{fang2024dfn}, MetaCLIP~\citep{xu2024metaclip}, and
OpenCLIP~\citep{cherti2023scaling} lead retrieval but use billion-scale corpora.


On classification (Table~\ref{tab:classification_tasks},
Appendix~\ref{app:zeroshot_fewshot}), $\mathrm{TDN}$/$\mathrm{TDDN}$ trail CLIP
and modern encoders on Food-101, Caltech-101, and GTSRB, consistent with their
smaller alignment corpus. Light supervision substantially closes this gap:
TIP-Adapter~\citep{tipadapter} with $K{=}16$ examples/class
(Appendix~\ref{app:zeroshot_fewshot}) raises $\mathrm{TDDN}$ on Caltech-101 from
$79.66$ to $90.50$, surpassing CLIP's $86.49$.

\begin{table}[!t]
    \centering
    \caption{\textbf{Comparison of TDN and \tmethod{} with baselines on Cityscapes (City) and Puzzle segmentation (Seg), keypoint matching (KPM), and image--text retrieval (I2T and T2I)}. Higher ($\uparrow$) is better for all metrics. \textbf{Bold} = best in column. STRUCTURE applies a lightweight alignment to DINOv2; its results are copied directly from the source paper~\citep{structure}, which does not report segmentation or keypoint matching, so those entries are empty ($-$).}
    \label{tab:matching_retrieval_results}
    \setlength{\tabcolsep}{1pt}
    \newlength{\tableFourRetrievalWidth}
    \settowidth{\tableFourRetrievalWidth}{\textbf{MS-COCO-14}}
    \setlength{\tableFourRetrievalWidth}{\dimexpr\tableFourRetrievalWidth/2-\tabcolsep\relax}
    \begin{tabularx}{\linewidth}{>{\raggedright\arraybackslash}X ccc *{4}{>{\centering\arraybackslash}p{\tableFourRetrievalWidth}}}
        \toprule
        \multirow{3}{*}{\textbf{Model}} & \textbf{City} & \textbf{Puzzle} & \textbf{SPair-71k}
        & \multicolumn{2}{c}{\textbf{Flickr30K}} & \multicolumn{2}{c}{\textbf{MS-COCO-14}} \\
        \cmidrule(lr){2-2}\cmidrule(lr){3-3}\cmidrule(lr){4-4}\cmidrule(lr){5-6}\cmidrule(lr){7-8}
        & \textbf{Seg} & \textbf{Seg} & \textbf{KPM} & \textbf{I2T} & \textbf{T2I} & \textbf{I2T} & \textbf{T2I} \\
        & (mIoU$\uparrow$) & (mIoU$\uparrow$) & (PCK@0.1$\uparrow$) & \multicolumn{2}{c}{(R@1$\uparrow$)} & \multicolumn{2}{c}{(R@1$\uparrow$)} \\
        \midrule
        CLIP ViT-L/14 \citep{clip} & 10.05 & 11.04 & 24.89 & 87.7 & 66.96 & 34.60 & 18.53 \\

        OpenCLIP ViT-L/14 \citep{cherti2023scaling} & 3.37 & 9.10 & 23.80 & 89.5 & 75.52 & 39.25 & 25.39 \\

        MetaCLIP ViT-L/14 \citep{xu2024metaclip} & 10.62 & 20.20 & 27.74 & 90.0 & 76.44 & 40.85 & 25.73 \\

        DFN ViT-L/14 \citep{fang2024dfn} & 10.24 & 11.16 & 28.55 & 89.8 & 75.34 & 41.68 & 26.90 \\

        SigLIP\,2 ViT-L/16 \citep{tschannen2025siglip} & 19.98 & 19.65 & \textbf{34.37} & \textbf{95.0} & \textbf{82.56} & \textbf{49.77} & \textbf{33.84} \\

        FG-CLIP\,2 \citep{xie2025fg2} & 31.39 & \textbf{23.64} & 28.88 & 93.8 & 80.86 & 47.05 & 32.69 \\

        STRUCTURE \citep{structure} & -- & -- & -- & 65.9 & 53.8 & 18.3 & 13.4 \\

        \midrule

        \rowcolor{gray!25}
        TDN (ours) & 24.29 & 20.92 & 28.29 & 85.8 & 72.88 & 37.1 & 24.1 \\

        \rowcolor{gray!25}
        \textbf{TDDN (ours)} & \textbf{32.38} & 22.51 & 32.39 & 85.3 & 71.24 & 36.1 & 24.0 \\
        \bottomrule
    \end{tabularx}
\end{table}

\paragraph{Key Takeaways.} Evaluations on segmentation, keypoint matching, and image--text retrieval support RQ2.
(i) \textbf{Fusion gains survive alignment.} TDDN exceeds TDN on segmentation and keypoint matching.
(ii) \textbf{Segmentation consistently exceeds CLIP.} TDDN outperforms CLIP across all segmentation benchmarks, spanning natural images and structured puzzles.
(iii) \textbf{Retrieval remains competitive.} TDDN surpasses CLIP ViT-L/14 in three of four retrieval settings despite limited alignment data. 

\subsection{Region-guided VLM Perception with \tmethod{}}

\label{sec:crg}

Set-of-Mark prompting~\citep{yang2023setofmark} relies on VLM visual priors; we use training-free CRG~\citep{Wan2024CRG} to test whether \tmethod{} regions improve puzzle grounding without costly end-to-end VLM training.

\paragraph{Datasets.}
We evaluate on two puzzle domains from \PuzzlePerception{}: Chess,
with $8$ multiple-choice questions per board spanning localization,
relational, geometric, tactical, and existence categories, and N-Queens,
with $4$ coordinate-free questions referencing extreme positions. All
questions are multiple choice and scored by accuracy. Further dataset
details are provided in Appendix~\ref{app:crg}.

\paragraph{Experimental setup.}
We test whether \tmethod{}'s predicted regions can steer a frozen VLM through Contrastive Region Guidance (CRG)~\citep{Wan2024CRG}, a training-free method that adjusts a VLM's output by contrasting its predictions on the full image against the image with the queried region blacked out. We compare three conditions: the image only (Img), CRG with ground-truth regions (oracle, $\Delta_{\text{o}}$), and CRG with \tmethod{}-predicted regions ($\Delta_{\text{TDDN}}$). The oracle provides an approximate upper bound on the achievable CRG gain and identifies which models can benefit at all, while $\Delta_{\text{TDDN}}$ measures the gain obtained from \tmethod{}'s predicted regions. \tmethod{} produces these regions from its text-conditioned segmentation, with the exact derivation per benchmark given in Appendix~\ref{app:crg}. No VLM weights are updated. We evaluate eight instruction-tuned VLMs across three families (Qwen2.5-VL/Qwen3.6, Gemma-3/4, InternVL-3).

\begin{table}[!t]
\centering
\caption{\textbf{Contrastive Region Guidance (CRG) with TDDN-predicted
regions on \PuzzlePerception{} (Chess).}
VLM question accuracy (\%), macro-averaged over questions.
\textbf{Img} is the image-only baseline. $\Delta_{\mathrm{o}}$ and
$\Delta_{\mathrm{TDDN}}$ are the accuracy changes from CRG using
ground-truth (oracle) and TDDN-predicted regions.}
\label{tab:crg_chess}

\setlength{\tabcolsep}{4pt}

\begin{tabular*}{0.75\linewidth}{
  @{\extracolsep{\fill}}lccc@{}
}
\toprule
\textbf{Model}
& \textbf{Img}
& $\boldsymbol{\Delta}_{\mathrm{o}}$
& $\boldsymbol{\Delta}_{\mathrm{TDDN}}$ \\
\midrule
InternVL3-8B            & 48.0 & $+3.2$ & $+2.1$ \\
Qwen2.5-VL-7B-Instruct  & 48.4 & $+5.6$ & $+3.8$ \\
Gemma-4-12B-it          & 60.0 & $+6.4$ & $+2.6$ \\
Gemma-3-27B-it          & 57.8 & $+1.7$ & $-0.6$ \\
Gemma-4-26B-A4B-it      & 72.8 & $+2.5$ & $+1.1$ \\
Qwen3.6-35B-A3B         & 79.9 & $-2.1$ & $-3.8$ \\
Qwen3.6-27B             & 80.1 & $+4.6$ & $+2.2$ \\
Gemma-4-31B-it          & 81.1 & $+1.5$ & $-0.4$ \\
\bottomrule
\end{tabular*}
\end{table}
\paragraph{Results.}
Table~\ref{tab:crg_chess} shows that \textbf{\tmethod{}-predicted regions improve Chess question answering for five of eight frozen VLMs}. Qwen2.5-VL-7B gains $3.8$ percentage points over its image-only baseline, increasing accuracy from $48.4\%$ to $52.2\%$; InternVL3-8B and Gemma-4-12B gain $2.1$ and $2.6$ points, respectively. Improvements also extend to Gemma-4-26B-A4B ($+1.1$) and Qwen3.6-27B ($+2.2$), showing that the benefit spans all three evaluated model families. \textbf{Predicted regions recover part of the oracle benefit}: for Qwen2.5-VL-7B and InternVL3-8B, they retain approximately $68\%$ and $66\%$ of the respective oracle gains ($+5.6$ and $+3.2$). Recovery is smaller for Gemma-4-12B, whose predicted-region gain is $+2.6$ compared with $+6.4$ under oracle guidance, indicating scope to improve the regions supplied to CRG. However, \textbf{guidance remains model-dependent}. Gemma-3-27B and Gemma-4-31B decline by $0.6$ and $0.4$ points with predicted regions despite positive oracle gains, while Qwen3.6-35B-A3B declines under both oracle ($-2.1$) and predicted-region ($-3.8$) guidance. Thus, accurate regions alone do not guarantee that CRG improves a given VLM. Baseline accuracy also does not fully explain the differences: Qwen3.6-27B benefits at $80.1\%$ image-only accuracy, whereas Gemma-4-31B does not at $81.1\%$.
\paragraph{Key Takeaways.} Region-guided PVQA evaluations support RQ3.
(i) \textbf{Predicted regions can improve frozen VLMs.} TDDN provides actionable spatial signals without updating VLM weights.
(ii) \textbf{Benefits remain model-dependent.} Five of eight evaluated models improve with TDDN-predicted regions.
(iii) \textbf{Oracle comparisons reveal guidance limits.} Even ground-truth regions can reduce accuracy, distinguishing CRG limitations from learned-mask errors. Extensions to chain-of-thought prompting, symbolic solvers, and richer grounding remain future work.


\section{Conclusion}
Modern VLM visual backbones, which primarily use ViT/CLIP, can sacrifice fine-grained perceptual detail for semantic alignment, limiting performance on structured visual reasoning tasks. We introduce \textbf{TDDN}, a text-aligned visual representation model with two stages: (1) \method{}, which fuses the perceptual precision of CleanDIFT with the semantic consistency of DINOv3, and (2) lightweight vision-language alignment using $\sim590$k image-text pairs. We further introduce the \PuzzlePerception{} dataset for puzzle-based segmentation and question answering. Despite the low-compute and low-data setting, \textbf{TDDN matches or surpasses CLIP} across multiple vision and vision-language benchmarks, 
with particularly large gains on segmentation tasks and comparable or better performance in {image--text retrieval tasks, surpassing CLIP ViT-L/14 in text-to-image on Flickr30K and both retrieval directions on MS-COCO-2014.}
On puzzle-based visual question answering, our \textbf{Contrastive Region Guidance} experiments show that TDDN representations provide useful visual signals that improve the performance of state-of-the-art open-weight VLMs. Our results demonstrate that diffusion-based and ViT-based visual representations offer \textit{useful}, complementary strengths for downstream puzzle-based tasks, and that TDDN successfully \textit{retains} these strengths after image-text alignment, \textit{making the ``V'' in VLM matter for puzzles}. We identify three directions for future work: scaling and diversifying the alignment corpus while assessing whether TDDN retains its dense-perception capabilities; exploring task-specific adaptation; {and integrating TDDN’s vision encoder directly into a VLM for end-to-end instruction tuning and evaluation across a broader range of visual reasoning tasks.}

\subsubsection*{Broader Impact Statement}

We use public benchmarks and procedurally generated puzzles without collecting participant data. We construct ReLAION-COCOA by filtering \textit{Not Safe for Work (NSFW)} content and recaptioning LAION-COCO-Aesthetic images to improve image--caption correspondence. Unsuitable samples and captioning errors may remain, alongside unresolved licensing, provenance, and representation bias concerns. Reuse should respect source licenses and distinguish generated captions from original annotations. Beyond puzzles, region prediction errors under distribution shift or for underrepresented categories require domain-specific validation, systematic error analysis, and human oversight proportionate to potential harm. Improved localization does not guarantee correct reasoning or suitability for high-stakes decisions. Our frozen-backbone, low-data alignment approach may broaden access to perception research, although pretraining and inference still incur computational and environmental costs.

\bibliography{main}

@String(PAMI  = {IEEE Trans. Pattern Anal. Mach. Intell.})

@String(CVPR  = {IEEE Conf. Comput. Vis. Pattern Recog.})

@String(ICCV  = {Int. Conf. Comput. Vis.})

@String(ECCV  = {Eur. Conf. Comput. Vis.})

@String(NeurIPS = {Adv. Neural Inform. Process. Syst.})

@String(ICML  = {Int. Conf. Mach. Learn.})

@String(ICLR  = {Int. Conf. Learn. Represent.})

@String(CVPRW = {IEEE Conf. Comput. Vis. Pattern Recog. Worksh.})

@String(TMLR  = {Trans. Mach. Learn Res.})

@String(PAMI  = {IEEE TPAMI})

@String(CVPR  = {CVPR})

@String(ICCV  = {ICCV})

@String(ECCV  = {ECCV})

@String(NeurIPS = {NeurIPS})

@String(ICML  = {ICML})

@String(ICLR  = {ICLR})

@String(CVPRW = {CVPRW})

@String(TMLR  = {TMLR})

@inproceedings{Luo:2025:deem_iclr,
  title     = {{DEEM}: Diffusion Models Serve as the Eyes of Large Language Models for Image Perception},
  author    = {Luo, Run and Li, Yunshui and Chen, Longze and He, Wanwei and Lin, Ting-En and Liu, Ziqiang and Zhang, Lei and Song, Zikai and Xia, Xiaobo and Liu, Tongliang and Yang, Min and Hui, Binyuan},
  booktitle = ICLR,
  year      = {2025},
  note      = {Spotlight}
}

@inproceedings{li:2025:lavida,
  title     = {{LaViDa}: A Large Diffusion Language Model for Multimodal Understanding},
  author    = {Li, Shufan and Kallidromitis, Konstantinos and Bansal, Hritik and Gokul, Akash and Kato, Yusuke and Kozuka, Kazuki and Kuen, Jason and Lin, Zhe and Chang, Kai-Wei and Grover, Aditya},
  booktitle = NeurIPS,
  year      = {2025},
  note      = {Spotlight}
}

@inproceedings{tong:2024:eyeswideshut,
  title     = {Eyes Wide Shut? {E}xploring the Visual Shortcomings of Multimodal {LLMs}},
  author    = {Tong, Shengbang and Liu, Zhuang and Zhai, Yuexiang and Ma, Yi and LeCun, Yann and Xie, Saining},
  booktitle = CVPR,
  pages     = {9568--9578},
  year      = {2024}
}

@inproceedings{chia2024puzzlevqa,
  title     = {{PuzzleVQA}: Diagnosing Multimodal Reasoning Challenges of Language Models with Abstract Visual Patterns},
  author    = {Chia, Yew Ken and Toh, Vernon Yuen Han and Ghosal, Deepanway and Bing, Lidong and Poria, Soujanya},
  booktitle = {Findings of the Assoc. Comput. Linguist.: ACL 2024},
  pages     = {16259--16273},
  year      = {2024}
}

@inproceedings{patel2024tripletclip,
  title={TripletCLIP: Improving Compositional Reasoning of CLIP via Synthetic Vision-Language Negatives},
  author={Patel, Maitreyi and Kusumba, Aakash and Cheng, Shen and Kim, Changdae and Gokhale, Tejas and Baral, Chitta and Yang, Yezhou},
  booktitle={NeurIPS},
  year={2024}
}

@article{dinov3,
title={{DINO}v3},
author={Oriane Sim{\'e}oni and Huy V. Vo and Maximilian Seitzer and Federico Baldassarre and Maxime Oquab and Cijo Jose and Vasil Khalidov and Marc Szafraniec and Seung Eun Yi and Michael Ramamonjisoa and Francisco Massa and Daniel HAZIZA and Luca Wehrstedt and Jianyuan Wang and Timoth{\'e}e Darcet and Th{\'e}o Moutakanni and Leonel Sentana and Claire Roberts and Andrea Vedaldi and Jamie Tolan and John Brandt and Camille Couprie and Julien Mairal and Herve Jegou and Patrick Labatut and Piotr Bojanowski},
journal={Transactions on Machine Learning Research},
issn={2835-8856},
year={2026},
url={https://openreview.net/forum?id=2NlGyqNjns},
note={Featured Certification}
}

@article{dinov2,
  title   = {{DINOv2}: Learning Robust Visual Features without Supervision},
  author  = {Oquab, Maxime and Darcet, Timoth{\'e}e and Moutakanni, Th{\'e}o and Vo, Huy V. and Szafraniec, Marc and Khalidov, Vasil and Fernandez, Pierre and Haziza, Daniel and Massa, Francisco and El-Nouby, Alaaeldin and Assran, Mahmoud and Ballas, Nicolas and Galuba, Wojciech and Howes, Russell and Huang, Po-Yao and Li, Shang-Wen and Misra, Ishan and Rabbat, Michael and Sharma, Vasu and Synnaeve, Gabriel and Xu, Hu and J{\'e}gou, Herv{\'e} and Mairal, Julien and Labatut, Patrick and Joulin, Armand and Bojanowski, Piotr},
  journal = TMLR,
  year    = {2024},
  url     = {https://arxiv.org/abs/2304.07193}
}

@inproceedings{dift,
  title     = {Emergent Correspondence from Image Diffusion},
  author    = {Tang, Luming and Jia, Menglin and Wang, Qianqian and Phoo, Cheng Perng and Hariharan, Bharath},
  booktitle = NeurIPS,
  year      = {2023}
}

@inproceedings{cleandift,
  title     = {{CleanDIFT}: Diffusion Features without Noise},
  author    = {Stracke, Nick and Baumann, Stefan Andreas and Bauer, Kolja and Fundel, Frank and Ommer, Bj{\"o}rn},
  booktitle = CVPR,
  year      = {2025}
}

@inproceedings{taleoftwo,
  title     = {A Tale of Two Features: Stable Diffusion Complements {DINO} for Zero-Shot Semantic Correspondence},
  author    = {Zhang, Junyi and Herrmann, Charles and Hur, Junhwa and Polan{\'i}a Cabrera, Luisa and Jampani, Varun and Sun, Deqing and Yang, Ming-Hsuan},
  booktitle = NeurIPS,
  year      = {2023}
}

@inproceedings{sd21,
  title     = {High-Resolution Image Synthesis with Latent Diffusion Models},
  author    = {Rombach, Robin and Blattmann, Andreas and Lorenz, Dominik and Esser, Patrick and Ommer, Bj{\"o}rn},
  booktitle = CVPR,
  pages     = {10684--10695},
  year      = {2022}
}

@inproceedings{clip,
  title     = {Learning Transferable Visual Models From Natural Language Supervision},
  author    = {Radford, Alec and Kim, Jong Wook and Hallacy, Chris and Ramesh, Aditya and Goh, Gabriel and Agarwal, Sandhini and Sastry, Girish and Askell, Amanda and Mishkin, Pamela and Clark, Jack and Krueger, Gretchen and Sutskever, Ilya},
  booktitle = ICML,
  pages     = {8748--8763},
  year      = {2021}
}

@misc{spair71k,
  title        = {{SPair-71k}: A Large-scale Benchmark for Semantic Correspondence},
  author       = {Min, Juhong and Lee, Jongmin and Ponce, Jean and Cho, Minsu},
  year         = {2019},
  eprint       = {1908.10543},
  archivePrefix = {arXiv},
  primaryClass = {cs.CV}
}

@inproceedings{imagenet,
  title     = {{ImageNet}: A Large-Scale Hierarchical Image Database},
  author    = {Deng, Jia and Dong, Wei and Socher, Richard and Li, Li-Jia and Li, Kai and Fei-Fei, Li},
  booktitle = CVPR,
  pages     = {248--255},
  year      = {2009}
}

@inproceedings{wang2020uniformity,
  title     = {Understanding Contrastive Representation Learning through Alignment and Uniformity on the Hypersphere},
  author    = {Wang, Tongzhou and Isola, Phillip},
  booktitle = ICML,
  pages     = {9929--9939},
  year      = {2020}
}

@inproceedings{zhang2019raven,
  title={Raven: A dataset for relational and analogical visual reasoning},
  author={Zhang, Chi and Gao, Feng and Jia, Baoxiong and Zhu, Yixin and Zhu, Song-Chun},
  booktitle={Proceedings of the IEEE/CVF conference on computer vision and pattern recognition},
  pages={5317--5327},
  year={2019}
}

@misc{ren2025vgrp,
  title        = {{VGRP-Bench}: Visual Grid Reasoning Puzzle Benchmark for Large Vision-Language Models},
  author       = {Ren, Yufan and Tertikas, Konstantinos and Maiti, Shalini and Han, Junlin and Zhang, Tong and S{\"u}sstrunk, Sabine and Kokkinos, Filippos},
  year         = {2025},
  eprint       = {2503.23064},
  archivePrefix = {arXiv},
  primaryClass = {cs.CV}
}

@inproceedings{structure,
  title     = {With Limited Data for Multimodal Alignment, Let the {STRUCTURE} Guide You},
  author    = {Gr{\"o}ger, Fabian and Wen, Shuo and Le, Huyen and Brbi{\'c}, Maria},
  booktitle = NeurIPS,
  year      = {2025}
}

@inproceedings{dinotxt,
  title     = {{DINOv2} Meets Text: A Unified Framework for Image- and Pixel-Level Vision-Language Alignment},
  author    = {Jose, Cijo and Moutakanni, Th{\'e}o and Kang, Dahyun and Baldassarre, Federico and Darcet, Timoth{\'e}e and Xu, Hu and Li, Daniel and Szafraniec, Marc and Ramamonjisoa, Micha{\"e}l and Oquab, Maxime and Sim{\'e}oni, Oriane and Vo, Huy V. and Labatut, Patrick and Bojanowski, Piotr},
  booktitle = CVPR,
  pages     = {24905--24916},
  year      = {2025}
}

@article{roberta,
  title   = {{RoBERTa}: A Robustly Optimized {BERT} Pretraining Approach},
  author  = {Liu, Yinhan and Ott, Myle and Goyal, Naman and Du, Jingfei and Joshi, Mandar and Chen, Danqi and Levy, Omer and Lewis, Mike and Zettlemoyer, Luke and Stoyanov, Veselin},
  journal = {arXiv preprint arXiv:1907.11692},
  year    = {2019}
}

@misc{openai2023gpt4v,
  title        = {{GPT-4} Technical Report},
  author       = {{OpenAI}},
  year         = {2023},
  eprint       = {2303.08774},
  archivePrefix = {arXiv},
  primaryClass = {cs.CL}
}

@misc{google2023gemini,
  title        = {Gemini: A Family of Highly Capable Multimodal Models},
  author       = {{Gemini Team, Google}},
  year         = {2023},
  eprint       = {2312.11805},
  archivePrefix = {arXiv},
  primaryClass = {cs.CL}
}

@inproceedings{liu2024llava,
  title     = {Visual Instruction Tuning},
  author    = {Liu, Haotian and Li, Chunyuan and Wu, Qingyang and Lee, Yong Jae},
  booktitle = NeurIPS,
  year      = {2025}
}

@inproceedings{chen2024internvl,
  title     = {{InternVL}: Scaling up Vision Foundation Models and Aligning for Generic Visual-Linguistic Tasks},
  author    = {Chen, Zhe and Wu, Jiannan and Wang, Wenhai and Su, Weijie and Chen, Guo and Xing, Sen and Zhong, Muyan and Zhang, Qinglong and Zhu, Xizhou and Lu, Lewei and Li, Bin and Luo, Ping and Lu, Tong and Qiao, Yu and Dai, Jifeng},
  booktitle = {Proceedings of the 2024 IEEE/CVF Conference on Computer Vision and Pattern Recognition},
  pages     = {24185--24198},
  year      = {2024},
  note      = {Oral},
  address={Seattle, Washington},
  publisher={IEEE/CVF},
  doi          = {10.1109/CVPR52733.2024.02283}
}

@inproceedings{cupl,
  title     = {What does a platypus look like? {G}enerating customized prompts for zero-shot image classification},
  author    = {Pratt, Sarah and Covert, Ian and Liu, Rosanne and Farhadi, Ali},
  booktitle = ICCV,
  year      = {2023}
}

@inproceedings{chen2025lazsl,
  title     = {Interpretable Zero-Shot Learning with Locally-Aligned Vision-Language Model},
  author    = {Chen, Shiming and Duan, Bowen and Khan, Salman and Khan, Fahad Shahbaz},
  booktitle = {Proceedings of the 2025 IEEE/CVF International Conference on Computer Vision},
  year      = {2025},
  publisher={IEEE/CVF},
  address={Hawaii},
  pages={478--487},
  doi={10.1109/ICCV51701.2025.00052},
  series={ICCV~'25}
}

@inproceedings{jain2024vcoder,
  title     = {{VCoder}: Versatile Vision Encoders for Multimodal Large Language Models},
  author    = {Jain, Jitesh and Yang, Jianwei and Shi, Humphrey},
  booktitle = CVPR,
  year      = {2024}
}

@article{xu2025lvlmehub,
  title   = {{LVLM-eHub}: A Comprehensive Evaluation Benchmark for Large Vision-Language Models},
  author  = {Xu, Peng and Shao, Wenqi and Zhang, Kaipeng and Gao, Peng and Liu, Shuo and Lei, Meng and Meng, Fanqing and Huang, Siyuan and Qiao, Yu and Luo, Ping},
  journal = PAMI,
  year    = {2025}
}

@inproceedings{ghosal2025algopuzzlevqa,
  title     = {{AlgoPuzzleVQA}: Diagnosing Multimodal Reasoning Challenges of Language Models with Algorithmic Multimodal Puzzles},
  author    = {Ghosal, Deepanway and Toh, Vernon and Chia, Yew Ken and Poria, Soujanya},
  booktitle = {Proc. Conf. North Am. Chapter Assoc. Comput. Linguist.},
  year      = {2025}
}

@inproceedings{lyu2025jigsawpuzzles,
  title     = {Jigsaw-Puzzles: From Seeing to Understanding to Reasoning in Vision-Language Models},
  author    = {Lyu, Zesen and Zhang, Dandan and Ye, Wei and Li, Fangdi and Jiang, Zhihang and Yang, Yao},
  booktitle = {Proc. Conf. Empir. Methods Nat. Lang. Process.},
  pages     = {26003--26014},
  year      = {2025}
}

@inproceedings{jiang2024marvel,
  title     = {{MARVEL}: Multidimensional Abstraction and Reasoning through Visual {EvaLuation}},
  author    = {Jiang, Jonathan and Wan, Yuxuan and Yu, Wenhao and Zhou, Zixin and Wang, Wenya and Yu, Jiating and Liu, Pengyuan},
  booktitle = NeurIPS,
  year      = {2024}
}

@inproceedings{zou2024dynamath,
  title={DynaMath: A Dynamic Visual Benchmark for Evaluating Mathematical Reasoning Robustness of Vision Language Models},
  author={Zou, Chengke and Guo, Xingang and Yang, Rui and Zhang, Junyu and Hu, Bin and Zhang, Huan},
  booktitle={International Conference on Learning Representations},
  year={2025}
}

@inproceedings{tipadapter,
  title     = {{Tip-Adapter}: Training-free Adaption of {CLIP} for Few-shot Classification},
  author    = {Zhang, Renrui and Zhang, Wei and Fang, Rongyao and Gao, Peng and Li, Kunchang and Dai, Jifeng and Qiao, Yu and Li, Hongsheng},
  booktitle = ECCV,
  year      = {2022}
}

@inproceedings{li2023blip2bootstrappinglanguageimagepretraining,
      title={{BLIP-2:} Bootstrapping Language-Image Pre-training with Frozen Image Encoders and Large Language Models}, 
      author={Junnan Li and Dongxu Li and Silvio Savarese and Steven Hoi},
      year={2023},
      booktitle={ICML},
}

@inproceedings{barsellotti2025talkingdinobridgingselfsupervised,
    author={Barsellotti, Luca and Bianchi, Lorenzo and Messina, Nicola and Carrara, Fabio and Cornia, Marcella and Baraldi, Lorenzo and Falchi, Fabrizio and Cucchiara, Rita},
  booktitle={Proceedings of the 2025 IEEE/CVF International Conference on Computer Vision}, 
  title={{Talking to DINO}: {B}ridging {S}elf-{S}upervised {V}ision {B}ackbones with {L}anguage for {O}pen-{V}ocabulary {S}egmentation}, 
  year={2025},
  volume={},
  number={},
  pages={22025--22035},
  doi={10.1109/ICCV51701.2025.02045},
  series={ICCV~'25},
  publisher={IEEE/CVF},
  address={Hawaii}
}

@inproceedings{lit,
  title     = {{LiT}: Zero-Shot Transfer with Locked-Image Text Tuning},
  author    = {Zhai, Xiaohua and Wang, Xiao and Mustafa, Basil and Steiner, Andreas and Keysers, Daniel and Kolesnikov, Alexander and Beyer, Lucas},
  booktitle = CVPR,
  pages     = {18123--18133},
  year      = {2022}
}

@misc{gemma3,
  title         = {{Gemma 3} Technical Report},
  author        = {{Gemma Team}},
  year          = {2025},
  eprint        = {2503.19786},
  archivePrefix = {arXiv},
  primaryClass  = {cs.CL},
  url           = {https://arxiv.org/abs/2503.19786}
}

@article{bai2025qwen25vl,
  title         = {{Qwen2.5-VL} Technical Report},
  author        = {Bai, Shuai and Chen, Keqin and Liu, Xuejing and Wang, Jialin and Ge, Wenbin and Song, Sibo and Dang, Kai and Wang, Peng and Wang, Shijie and Tang, Jun and Zhong, Humen and Zhu, Yuanzhi and Yang, Mingkun and Li, Zhaohai and Wan, Jianqiang and Wang, Pengfei and Ding, Wei and Fu, Zheren and Xu, Yiheng and Ye, Jiabo and Zhang, Xi and Xie, Tianbao and Cheng, Zesen and Zhang, Hang and Yang, Zhibo and Xu, Haiyang and Lin, Junyang},
  year          = {2025},
  journal={arXiv:2502.13923},
  eprint        = {2502.13923},
  archivePrefix = {arXiv},
  primaryClass  = {cs.CV},
  url           = {https://arxiv.org/abs/2502.13923}
}

@inproceedings{coco,
  title     = {{Microsoft COCO}: Common Objects in Context},
  author    = {Lin, Tsung-Yi and Maire, Michael and Belongie, Serge and Hays, James and Perona, Pietro and Ramanan, Deva and Doll{\'a}r, Piotr and Zitnick, C. Lawrence},
  booktitle = ECCV,
  pages     = {740--755},
  year      = {2014}
}

@inproceedings{flickr30k,
  title     = {{Flickr30k} Entities: Collecting Region-to-Phrase Correspondences for Richer Image-to-Sentence Models},
  author    = {Plummer, Bryan A. and Wang, Liwei and Cervantes, Chris M. and Caicedo, Juan C. and Hockenmaier, Julia and Lazebnik, Svetlana},
  booktitle = ICCV,
  pages     = {2641--2649},
  year      = {2015}
}

@inproceedings{food101,
author={Bossard, Lukas
and Guillaumin, Matthieu
and Van Gool, Luc},
title={{Food-101} -- {M}ining {D}iscriminative {C}omponents with {R}andom {F}orests},
booktitle={Proceedings of the 13th European Conference on Computer Vision},
year={2014},
publisher={Springer International Publishing},
address={Zurich},
pages={446--461},
doi={10.1007/978-3-319-10599-4_29},
series={ECCV~'14}
}

@techreport{cifar100,
  title       = {Learning Multiple Layers of Features from Tiny Images},
  author      = {Krizhevsky, Alex},
  institution = {University of Toronto},
  year        = {2009}
}

@inproceedings{caltech101,
  title     = {Learning Generative Visual Models from Few Training Examples: An Incremental {Bayesian} Approach Tested on 101 Object Categories},
  author    = {Fei-Fei, Li and Fergus, Rob and Perona, Pietro},
  booktitle = CVPRW,
  year      = {2004}
}

@article{gtsrb,
  title   = {Man vs. computer: Benchmarking machine learning algorithms for traffic sign recognition},
  author  = {Stallkamp, J. and Schlipsing, M. and Salmen, J. and Igel, C.},
  journal = {Neural Networks},
  volume  = {32},
  pages   = {323--332},
  year    = {2012}
}

@inproceedings{ade20k,
  title     = {Scene Parsing through {ADE20K} Dataset},
  author    = {Zhou, Bolei and Zhao, Hang and Puig, Xavier and Fidler, Sanja and Barriuso, Adela and Torralba, Antonio},
  booktitle = CVPR,
  pages     = {633--641},
  year      = {2017}
}

@misc{beaumont2022laion5b,
  title        = {LAION-5B: A New Era of Open Large-Scale Multi-Modal Datasets},
  author       = {Beaumont, Romain},
  year         = {2022},
  month        = {March},
  howpublished = {\url{https://laion.ai/blog/laion-5b/}},
  note         = {LAION Blog. Accessed: September 7, 2026}
}

@inproceedings{nakata2022knn,
title = {{Revisiting a kNN-based Image Classification System with High-capacity Storage}},
author = {Nakata, Kengo and Ng, Youyang and Miyashita, Daisuke and Maki, Asuka and Lin, Yu-Chieh and Deguchi, Jun},
booktitle = {Proceedings of the European Conference on Computer Vision},
series = {ECCV '22},
year = {2022},
url = {https://www.ecva.net/papers/eccv_2022/papers_ECCV/html/1552_ECCV_2022_paper.php},
}

@misc{internvl3,
  title         = {{InternVL3}: Exploring Advanced Training and Test-Time Recipes for Open-Source Multimodal Models},
  author        = {Zhu, Jinguo and Wang, Weiyun and Chen, Zhe and Liu, Zhaoyang and Ye, Shenglong and Gu, Lixin and Tian, Hao and Duan, Yuchen and Su, Weijie and Shao, Jie and Gao, Zhangwei and Cui, Erfei and Wang, Xuehui and Cao, Yue and Liu, Yangzhou and Wei, Xingguang and Zhang, Hongjie and Wang, Haomin and Xu, Weiye and Li, Hao and Wang, Jiahao and Deng, Nianchen and Li, Songze and He, Yinan and Jiang, Tan and Luo, Jiapeng and Wang, Yi and He, Conghui and Shi, Botian and Zhang, Xingcheng and Lu, Wenqi and Shao, Wenwen and Wang, Yangjun and Yu, Pan and Wang, Bin and Lin, Dahua and Wang, Yu Qiao and Dai, Jifeng and Wang, Wenhai},
  year          = {2025},
  eprint        = {2504.10479},
  archivePrefix = {arXiv},
  primaryClass  = {cs.CV},
  url           = {https://arxiv.org/abs/2504.10479}
}

@inproceedings{Wan2024CRG,
  author = {David Wan and Jaemin Cho and Elias Stengel-Eskin and Mohit Bansal},
  title  = {Contrastive Region Guidance: Improving Grounding in Vision-Language Models without Training},
  year   = {2024},
  booktitle = {ECCV}
}

@inproceedings{cityscapes,
author = {Cordts, Marius and Omran, Mohamed and Ramos, Sebastian and Rehfeld, Timo and Enzweiler, Markus and Benenson, Rodrigo and Franke, Uwe and Roth, Stefan and Schiele, Bernt},
title = {The Cityscapes Dataset for Semantic Urban Scene Understanding},
booktitle = {Proceedings of the IEEE Conference on Computer Vision and Pattern Recognition (CVPR)},
month = {June},
year = {2016}
}

@inproceedings{cocostuff,
  title={COCO-Stuff: Thing and stuff classes in context},
  author={Caesar, Holger and Uijlings, Jasper and Ferrari, Vittorio},
  booktitle={Proceedings of the 2018 IEEE/CVF Conference on Computer Vision and Pattern Recognition},
  address={Utah},
  publisher={IEEE/CVF},
  series={CVPR~'18},
  year={2018},
  doi={10.1109/CVPR.2018.00132}
}

@inproceedings{pascalcontext,
author = {Mottaghi, Roozbeh and Chen, Xianjie and Liu, Xiaobai and Cho, Nam-Gyu and Lee, Seong-Whan and Fidler, Sanja and Urtasun, Raquel and Yuille, Alan},
title = {The Role of Context for Object Detection and Semantic Segmentation in the Wild},
booktitle = {Proceedings of the IEEE Conference on Computer Vision and Pattern Recognition (CVPR)},
month = {June},
year = {2014}
}

@inproceedings{fang2024dfn,
 author = {Fang, Alex and Madappally Jose, Albin and Jain, Amit and Schmidt, Ludwig and Toshev, Alexander and Shankar, Vaishaal},
 booktitle = {International Conference on Learning Representations},
 editor = {B. Kim and Y. Yue and S. Chaudhuri and K. Fragkiadaki and M. Khan and Y. Sun},
 pages = {36221--36237},
 title = {Data Filtering Networks},
 url = {https://proceedings.iclr.cc/paper_files/paper/2024/file/9bf0810a4a1597a36d27ceea58667d92-Paper-Conference.pdf},
 volume = {2024},
 year = {2024}
}

@inproceedings{xu2024metaclip,
 author = {Xu, Hu and Xie, Saining and Tan, Xiaoqing and Huang, Po-Yao and Howes, Russell and Sharma, Vasu and Li, Shang-Wen and Ghosh, Gargi and Zettlemoyer, Luke and Feichtenhofer, Christoph},
 booktitle = {International Conference on Learning Representations},
 editor = {B. Kim and Y. Yue and S. Chaudhuri and K. Fragkiadaki and M. Khan and Y. Sun},
 pages = {47812--47831},
 title = {Demystifying CLIP Data},
 url = {https://proceedings.iclr.cc/paper_files/paper/2024/file/d1450d6c10c6b6cf1b80964357f5fa08-Paper-Conference.pdf},
 volume = {2024},
 year = {2024}
}

@inproceedings{cherti2023scaling,
   title={Reproducible Scaling Laws for Contrastive Language-Image Learning},
   url={http://dx.doi.org/10.1109/CVPR52729.2023.00276},
   DOI={10.1109/cvpr52729.2023.00276},
   booktitle={2023 IEEE/CVF Conference on Computer Vision and Pattern Recognition (CVPR)},
   publisher={IEEE},
   author={Cherti, Mehdi and Beaumont, Romain and Wightman, Ross and Wortsman, Mitchell and Ilharco, Gabriel and Gordon, Cade and Schuhmann, Christoph and Schmidt, Ludwig and Jitsev, Jenia},
   year={2023},
   month={June}, pages={2818–2829} }

@InProceedings{Zhai:2023:SigLIP,
    author    = {Zhai, Xiaohua and Mustafa, Basil and Kolesnikov, Alexander and Beyer, Lucas},
    title     = {Sigmoid Loss for Language Image Pre-Training},
    booktitle = {Proceedings of the IEEE/CVF International Conference on Computer Vision (ICCV)},
    month     = {October},
    year      = {2023},
    pages     = {11975-11986}
}

@misc{yang2023setofmark,
      title={Set-of-Mark Prompting Unleashes Extraordinary Visual Grounding in GPT-4V}, 
      author={Jianwei Yang and Hao Zhang and Feng Li and Xueyan Zou and Chunyuan Li and Jianfeng Gao},
      year={2023},
      eprint={2310.11441},
      archivePrefix={arXiv},
      primaryClass={cs.CV},
      url={https://arxiv.org/abs/2310.11441}, 
}

@article{tschannen2025siglip,
      title={SigLIP 2: Multilingual Vision-Language Encoders with Improved Semantic Understanding, Localization, and Dense Features}, 
      author={Michael Tschannen and Alexey Gritsenko and Xiao Wang and Muhammad Ferjad Naeem and Ibrahim Alabdulmohsin and Nikhil Parthasarathy and Talfan Evans and Lucas Beyer and Ye Xia and Basil Mustafa and Olivier Hénaff and Jeremiah Harmsen and Andreas Steiner and Xiaohua Zhai},
      year={2025},
      journal={arXiv preprint arXiv:2502.14786},
      eprint={2502.14786},
      archivePrefix={arXiv},
      primaryClass={cs.CV},
      url={https://arxiv.org/abs/2502.14786}, 
}

@inproceedings{xie2025fg2,
  title={FG-CLIP 2: A Bilingual Fine-grained Vision-language Alignment Model},
  author={Xie, Chunyu and Wang, Bin and Kong, Fanjing and Li, Jincheng and Liang, Dawei and Ao, Ji and Leng, Dawei and Yin, Yuhui},
  booktitle={International Conference on Machine Learning},
  year={2026}
}

@inproceedings{maskclip,
      title={Extract Free Dense Labels from CLIP},
      author={Zhou, Chong and Loy, Chen Change and Dai, Bo},
      booktitle={Proceedings of the European Conference on Computer Vision (ECCV)},
      year={2022},
      doi={10.1007/978-3-031-19815-1_40}
}

@inproceedings{tcl,
  title={Learning to Generate Text-grounded Mask for Open-world Semantic Segmentation from Only Image-Text Pairs},
  author={Cha, Junbum and Mun, Jonghwan and Roh, Byungseok},
  booktitle={Proceedings of the 2023 IEEE/CVF Conference on Computer Vision and Pattern Recognition},
  year={2023},
  series={CVPR~'23},
  publisher={IEEE/CVF},
  address={Vancouver},
  pages={11165-11174},
  doi={10.1109/CVPR52729.2023.01074}
}

@article{recap,
      title={A Picture is Worth a Thousand Words: Principled Recaptioning Improves Image Generation}, 
      author={Eyal Segalis and Dani Valevski and Danny Lumen and Yossi Matias and Yaniv Leviathan},
      year={2023},
      journal={arXiv:2310.16656},
      eprint={2310.16656},
      archivePrefix={arXiv},
      primaryClass={cs.CV},
      url={https://arxiv.org/abs/2310.16656}, 
}

@inproceedings{clevr,
  title={CLEVR: A Diagnostic Dataset for Compositional Language and Elementary Visual Reasoning},
  author={Johnson, Justin and Hariharan, Bharath and van der Maaten, Laurens
          and Fei-Fei, Li and Zitnick, C Lawrence and Girshick, Ross},
  booktitle={CVPR},
  year={2017}
}

@inproceedings{gqa,
author = {Hudson, Drew A. and Manning, Christopher D.},
title = {GQA: A New Dataset for Real-World Visual Reasoning and Compositional Question Answering},
booktitle = {Proceedings of the IEEE/CVF Conference on Computer Vision and Pattern Recognition (CVPR)},
month = {June},
year = {2019}
}

@misc{kojic2024chessrender360,
  author       = {Koji{\'c}, Marko},
  title        = {{ChessRender360}: High-Fidelity Rendered Chess Dataset with Multi-Modal Annotations},
  year         = {2024},
  publisher    = {Zenodo},
  doi          = {10.5281/zenodo.13356818},
  url          = {https://doi.org/10.5281/zenodo.13356818},
  note         = {Dataset}
}

@misc{laion_relaion_coco,
  author       = {{LAION Team}},
  title        = {{LAION-COCO}},
  howpublished = {\url{https://huggingface.co/datasets/laion/relaion-coco}},
  note         = {Hugging Face dataset},
  year         = {2024},
  urldate = {2026-09-19}
}

@misc{li_laion_coco_aesthetic,
  author       = {Liu, Guangyi},
  title        = {{LAION-COCO-Aesthetic}},
  howpublished = {\url{https://huggingface.co/datasets/guangyil/laion-coco-aesthetic}},
  note         = {Hugging Face dataset},
  urldate      = {2026-09-19},
  year = {2023}
}
\bibliographystyle{tmlr}

\appendix
\setcounter{secnumdepth}{2}
\renewcommand{\thesection}{\Alph{section}}
\renewcommand{\thesubsection}{\thesection.\arabic{subsection}}
\renewcommand{\thesubsubsection}{\thesubsection.\arabic{subsubsection}}

\newpage

\section{Limitations}

Our vision--language alignment uses approximately 590K image--text pairs. Many modern vision--language encoders train on web-scale corpora, so our alignment corpus provides narrower semantic and domain coverage. In particular, it covers specialized categories, such as traffic signs, only sparsely, which limits the zero-shot classification in these domains. 
{On the other hand, one can argue that the limited or sparse data in certain categories may decrease confusion and possibly boost the final perception performance. Although our sampling aims to maintain diversity, this alone does not rule out incidental advantages or disadvantages for individual evaluation categories.
Future work can scale and diversify the alignment corpus, examine the effects of corpus size and composition, and assess whether TDDN retains its dense-perception capabilities.}

We exclude \emph{Puzzle Perception} and all other puzzle-specific data from vision–language alignment. Consequently, our main puzzle results measure zero-shot transfer, while the potential benefits of task-specific adaptation remain unexplored. Within our computational budget, we evaluate downstream utility through CRG with frozen VLMs instead of training a complete VLM with TDDN as its vision encoder. {Future work can directly integrate TDDN's vision encoder into a VLM, instruction-tune the resulting model end-to-end, and evaluate its performance across a broader range of visual reasoning tasks.}

\section{Additional Methodology Details}\label{app:methodology_details}

\subsection{Notation}
\label{app:notation}

We summarize the notation used throughout the paper. Each representation is identified by a base name and, where relevant, a subscript indicating the representation granularity.

\paragraph{Base representations.}
\begin{itemize}
    \item $\mathrm{DN}$: DINOv3~\citep{dinov3} features extracted from the frozen backbone.
    \item $\mathrm{CD}$: CleanDIFT~\citep{cleandift} features extracted from the frozen diffusion UNet.
    \item $\mathrm{SD}$: Vanilla Stable Diffusion features extracted without CleanDIFT's distillation.
    \item $\mathrm{DDN}$: The fused representation obtained by combining $\mathrm{DN}$ and $\mathrm{CD}$. 
\end{itemize}

\paragraph{Granularity subscripts.}
We attach $p$, $g$, and $\bar{p}$ to a base representation to indicate which part of the representation is used:
\begin{itemize}
    \item $p$ (patch): the full spatial grid of patch tokens, e.g., $\mathbf{f}^{\mathrm{DN}_p}_v$.
    \item $g$ (global): the CLS token, e.g., $\mathbf{f}^{\mathrm{DN}_g}_v$.
    \item $\bar{p}$ (patch mean): the average of all patch tokens, e.g., $\mathbf{g}^{\mathrm{DDN}_{\bar{p}}}_v$.
\end{itemize}

\paragraph{Feature notation.}
We denote feature representations before the fusion transformer by $\mathbf{f}$ and the corresponding output representations after the self-attention blocks by $\mathbf{g}$. The subscript $v$ denotes the visual modality (image), while $t$ denotes the text modality. For example, $\mathbf{f}^{\mathrm{DN}_p}_v$ denotes the input visual patch features extracted from $\mathrm{DN}$, and $\mathbf{g}^{\mathrm{DDN}_{\bar{p}}}_v$ denotes the patch-mean representation after the fusion transformer itself.

\paragraph{PCA projection ($X'$).}
For $X \in \{\mathrm{CD}, \mathrm{SD}, \mathrm{DN}, \mathrm{DDN}\}$, $X'$ denotes the representation constructed from PCA-projected constituent feature components. For example, $\mathrm{CD}'$ and $\mathrm{SD}'$ are constructed from the PCA-projected intermediate feature components of their respective models, while $\mathrm{DDN}'$ is obtained by fusing $\mathrm{CD}'$ with $\mathrm{DN}$. During text alignment, we replace these PCA projections with learned MLP projection heads; therefore, we do not use the prime notation for $\mathrm{TDN}$ or $\mathrm{TDDN}$. Appendix~\ref{app:feature_extraction_fusion} provides the construction details.

\paragraph{$T$-prefix.}
Prepending $T$ denotes the text-aligned version of a representation:
\begin{itemize}
    \item $\mathrm{TDN}$: Text-aligned $\mathrm{DN}$ (no feature fusion).
    \item $\mathrm{TDDN}$: Text-aligned $\mathrm{DDN}$, our full model.
\end{itemize}

\paragraph{Embeddings.}
$\mathbf{z}_{\mathrm{img}}$ and $\mathbf{z}_{\mathrm{txt}}$ denote the final image and text embeddings used for contrastive alignment.

\subsection{Construction of Vision Representations}
\label{app:feature_extraction_fusion}
The feature-extraction pipeline below is shared across every task except text alignment. With patch size $16$, an input of $H{\times}H$ pixels yields a shared $N{\times}N$ patch grid, $N{=}H/16$. We construct the $\mathrm{SD}$, $\mathrm{CD}$, and $\mathrm{DDN}$ representations by fusing feature components with different channel dimensions. We project each component to a common dimension to balance their contributions evenly during fusion in the representation.

For vision-only representations, we fit these projections using PCA on the training set, following~\cite{taleoftwo}. During text alignment, we replace the fixed PCA projections with learned MLP projection heads. Throughout the paper, the prime notation ($X'$) always denotes vision-only representations constructed through this PCA feature fusion.

\paragraph{CleanDIFT features ($\mathbf{f}^{\mathrm{CD}'}_{v}$).}
We pass each image through the frozen CleanDIFT UNet and extract intermediate
activations from decoder layers $\ell \in \{2,5,8\}$ (best-performing combination
according to the ablation in Table~\ref{tab:cleandift_layers_spair}), denoted
$\mathbf{f}^{\mathrm{CD}_p}_{\ell}\in\mathbb{R}^{N^2\times d_\ell}$ with
$(d_2,d_5,d_8)=(1280,640,640)$. We spatially interpolate each activation map to a
shared $N\times N$ grid. We project each feature component to a common dimension,
$d_c=512$, using PCA fitted on the training set,
following~\cite{taleoftwo}, for dimensional consistency:

\begin{equation}
\label{eq:cd_layers}
\hat{\mathbf{f}}^{\mathrm{CD}'}_{\ell}
=
\widehat{P_\ell\!\left(\mathbf{f}^{\mathrm{CD}_p}_{\ell}\right)}
\in
\mathbb{R}^{N^2\times d_c},
\qquad
\ell\in\{2,5,8\},
\end{equation}

where $P_\ell(\cdot):\mathbb{R}^{d_\ell}\rightarrow\mathbb{R}^{d_c}$ denotes the
per-layer PCA projection and $\hat{\cdot}$ denotes $\ell_2$ normalization, respectively. The
projected feature components are then concatenated along the channel dimension
to construct the $\mathrm{CD}'$ representation:

\begin{equation}
\label{eq:cd_concat}
\mathbf{f}^{\mathrm{CD}'}_{v}
=
\hat{\mathbf{f}}^{\mathrm{CD}'}_{2}
\oplus
\hat{\mathbf{f}}^{\mathrm{CD}'}_{5}
\oplus
\hat{\mathbf{f}}^{\mathrm{CD}'}_{8}
\in
\mathbb{R}^{N^2\times3d_c},
\end{equation}

where $\oplus$ denotes channel-wise concatenation and $3d_c=1536$. $\mathrm{SD}'$ features are extracted in the same manner. As Stable Diffusion requires a noisy latent to produce informative representations, we extract features at the diffusion timestep $t=261$, following prior standard feature-extraction practice.

\begin{table}[!t]
\centering
\caption{Keypoint matching on SPair-71k (keypoint-weighted PCK@0.1 on
bounding-box crops). The combined fusion of layers $2\oplus5\oplus8$ outperforms every
individual layer.}
\label{tab:cleandift_layers_spair}

\setlength{\tabcolsep}{2pt}

\begin{tabular*}{\linewidth}{@{\extracolsep{\fill}}l*{13}{c}@{}}
\toprule
\textbf{Metric}
& \textbf{L0}
& \textbf{L1}
& \textbf{L2}
& \textbf{L3}
& \textbf{L4}
& \textbf{L5}
& \textbf{L6}
& \textbf{L7}
& \textbf{L8}
& \textbf{L9}
& \textbf{L10}
& \textbf{L11}
& $\boldsymbol{2\oplus5\oplus8}$ \\
\midrule
$\mathrm{PCK@0.1}^{\mathrm{bbox}}$
& 8.87
& 11.91
& 8.72
& 38.69
& 45.83
& 50.98
& 43.53
& 36.07
& 25.85
& 18.71
& 9.10
& 6.84
& \textbf{61.20} \\
\bottomrule
\end{tabular*}
\end{table}
\paragraph{DINOv3 features ($\mathbf{f}^{\mathrm{DN}}_v$).}
We extract patch tokens from the final transformer layer of the frozen DINOv3
backbone (no register tokens), yielding the patch features
$\mathbf{f}^{\mathrm{DN}_p}_v \in \mathbb{R}^{N^2 \times d_n}$, the mean-pooled patch
representation $\mathbf{f}^{\mathrm{DN}_{\bar{p}}}_v \in \mathbb{R}^{d_n}$, and the
global CLS representation $\mathbf{f}^{\mathrm{DN}_g}_v \in \mathbb{R}^{d_n}$, where
$d_n = 1280$.

\paragraph{\method{} features ($\mathbf{f}^{\mathrm{DDN}'}_v$).}
We fuse the $\mathrm{CD}'$ and $\mathrm{DN}$ patch representations using the
normalize-and-concatenate strategy of \citet{taleoftwo}:

\begin{equation}
\label{eq:fusion}
\mathbf{f}^{\mathrm{DDN}'_p}_v =
  \alpha\,\hat{\mathbf{f}}^{\mathrm{CD}'}_v
  \;\oplus\;
  (1-\alpha)\,\hat{\mathbf{f}}^{\mathrm{DN}_p}_v
  \;\in\;
  \mathbb{R}^{N^2 \times (3d_c+d_n)}.
\end{equation}

For tasks requiring a single global image representation, we concatenate the
mean-pooled fused patch representation,
$\mathbf{f}^{\mathrm{DDN}'_{\bar{p}}}_v \in \mathbb{R}^{3d_c+d_n}$,
with the DINOv3 CLS representation:

\begin{equation}
\label{eq:fusion_global}
\mathbf{f}^{\mathrm{DDN}'}_v =
\hat{\mathbf{f}}^{\mathrm{DN}_g}_v
\;\oplus\;
\hat{\mathbf{f}}^{\mathrm{DDN}'_{\bar{p}}}_v
\;\in\;
\mathbb{R}^{4096},
\end{equation}

where $\alpha\in[0,1]$ balances the contribution of the two feature
sources. We set $\alpha=0.5$ throughout.

\paragraph{Per-task feature usage.}
For keypoint matching, we use patch-level features
$\mathbf{f}^{X_p}_v$ as dense descriptors, where
$X \in \{\mathrm{CD}', \mathrm{SD}', \mathrm{DN}, \mathrm{DDN}', \mathrm{CLIP}, \mathrm{TDN}, \mathrm{TDDN}\}$.
We match source keypoints to points in the target image using nearest-neighbor search in feature space without additional training.
For semantic segmentation, we train a lightweight linear probe on $\mathbf{f}^{X_p}_v$ to predict each patch's semantic label, using the same protocol across representations unless otherwise specified. For image classification, we apply $k$-nearest neighbor classification to global features $\mathbf{f}^{X}_v$. Among the evaluated tasks, only semantic segmentation requires task-specific training.

\paragraph{Choice of activation layers.}
We adopt the layer combination (2, 5, 8) for CleanDIFT/SD feature extraction, following the recipe established by~\cite{taleoftwo}. To verify its competitiveness with alternative single-layer extractions, we evaluate semantic keypoint matching on SPair-71k across all twelve UNet decoder activation maps; Table~\ref{tab:cleandift_layers_spair} clearly confirms that fused 2+5+8 outperforms every individual layer.

\paragraph{Choice of  $\alpha$}
Empirically, we observe that $\alpha = 0.5$ provides the best trade-off between the two feature representations, effectively capturing their complementary properties.

\paragraph{Choice of CleanDIFT over Stable Diffusion in DDN.}
We adopt CleanDIFT in place of standard Stable Diffusion features for two reasons. First, CleanDIFT removes the iterative denoising process and therefore avoids timestep selection while reducing inference overhead. Second, in our experiments, CleanDIFT consistently provides stronger visual representations across all downstream perception tasks. As shown in Table~\ref{tab:quantitative_results}, CleanDIFT outperforms Stable Diffusion on semantic segmentation (79.20 vs. 74.77 mIoU), keypoint matching (61.20 vs. 56.46 PCK@0.1), and image classification (38.74 vs. 12.54 top-1 accuracy). We therefore use CleanDIFT as the default diffusion-based component throughout the paper.

\subsection{Text Alignment Architecture for TDN}
\label{app:tdn_details}

To align DINOv3 with text without feature fusion, we adopt the same
text-alignment pipeline as \tmethod{}, replacing the fused vision
representation with the DINOv3 representation. Specifically, the vision
encoder processes only $\mathrm{DN}$ features, while the text encoder,
training objective, and optimization procedure remain identical to those
described in Section~\ref{sec:text_align}.

\paragraph{Vision encoder.}
We refine the DINOv3 representation
$\mathbf{f}^{\mathrm{DN}}_v$ using two stacked trainable self-attention
blocks, $g_{\mathrm{SA}}^{(2)}(\cdot)$:

\begin{equation}
\left[
\mathbf{g}^{\mathrm{DN}_g}_v,\,
\mathbf{g}^{\mathrm{DN}_p}_v
\right]
=
g_{\mathrm{SA}}^{(2)}
\!\left(
\mathbf{f}^{\mathrm{DN}}_v
\right),
\end{equation}

where
$\mathbf{g}^{\mathrm{DN}_g}_v\in\mathbb{R}^{d_n}$ and
$\mathbf{g}^{\mathrm{DN}_p}_v\in\mathbb{R}^{N^2\times d_n}$.
We average-pool the patch representations to obtain :

\begin{equation}
\mathbf{g}^{\mathrm{DN}_{\bar p}}_v
=
\mathrm{AvgPool}
\!\left(
\mathbf{g}^{\mathrm{DN}_p}_v
\right)
\in
\mathbb{R}^{d_n},
\end{equation}

and concatenate the global and mean-pooled patch representations to form the
final image embedding :

\begin{equation}
\mathbf{z}_{\mathrm{img}}
=
\mathbf{g}^{\mathrm{DN}_g}_v
\oplus
\mathbf{g}^{\mathrm{DN}_{\bar p}}_v
\in
\mathbb{R}^{2d_n}.
\end{equation}

\begin{table}[!t]
\centering
\caption{Frozen and trainable components shared by TDN and TDDN.
Vision-side rows marked ``TDDN only'' apply exclusively to the
CleanDIFT-fused variant.}
\label{tab:frozen_trainable}

\setlength{\tabcolsep}{4pt}
\renewcommand{\arraystretch}{1.15}

\begin{tabularx}{\linewidth}{
  l
  l
  >{\raggedright\arraybackslash}X
}
\toprule
\textbf{Component}
& \textbf{Status}
& \textbf{Notes} \\
\midrule

\rowcolor{gray!15}
\multicolumn{3}{l}{\textit{Vision}} \\

DINOv3 backbone
& Frozen
& Shared by TDN and TDDN. \\

CleanDIFT backbone
& Frozen
& TDDN only; fused with DINOv3 features. \\

Per-layer MLPs
& Trainable
& Projects each CleanDIFT layer (2, 5, 8) to a common dimension. \\

Post-fusion MLP
& Trainable
& Projects fused patch tokens to same dimension as CLS token. \\

Vision attention blocks ($K$ layers)
& Trainable
& Applied to the visual representations with RoPE to align with the
text features. \\

\midrule

\rowcolor{gray!15}
\multicolumn{3}{l}{\textit{Text}} \\

RoBERTa encoder
& Frozen
& Shared by TDN and TDDN \\

Text attention blocks ($K$ layers)
& Trainable
& Applied to the final RoBERTa representations to align with the
visual features. \\

Text projection $W_t$
& Trainable
& Projects text tokens to the joint embedding dimension. \\

\bottomrule
\end{tabularx}
\end{table}

\paragraph{Text encoder.}
We use the same text encoder as described in
Section~\ref{sec:text_align}. Specifically, we refine the frozen
RoBERTa-L token representations using two self-attention blocks,
aggregate them with masked mean pooling, and apply a trainable linear
projection to obtain the text embedding
$\mathbf{z}_{\mathrm{txt}}\in\mathbb{R}^{2d_n}$.
\subsection{Additional Vision--Language Alignment Details}
\label{app:vision-text_alignment}

\paragraph{Shared alignment pipeline.}
TDN and TDDN share the same vision--language alignment pipeline and differ
only in the input visual representation. TDN uses the DINOv3 representation
$\mathbf{f}^{\mathrm{DN}}_v$, whereas TDDN uses the fused representation
$\mathbf{f}^{\mathrm{DDN}}_v$. The text encoder, training objective and hyperparameters otherwise remain identical for both models.

\paragraph{Architecture details.}
On both the vision and text branches, we refine the frozen backbone
representations using two stacked self-attention blocks with rotary position
embeddings (RoPE). For the vision branch, we form the final image embedding by
concatenating the global representation with the mean-pooled patch
representation. For the text branch, we aggregate the refined token
representations using masked mean pooling followed by a trainable linear
projection. These design choices closely follow prior work on text alignment for visual representations~\citep{dinotxt} while also explicitly preserving the dense spatial information required for downstream perception tasks.
We optimize the image and text embeddings using a symmetric InfoNCE objective
together with the STRUCTURE regularizer~\citep{structure}.



The InfoNCE objective learns a shared vision--language embedding space,
whereas the STRUCTURE regularizer preserves the intrinsic neighborhood
structure of the pretrained visual and text representations during alignment. This improves optimization in the limited image--text pair regime by
retaining the geometric structure learned by the pretrained encoder instead of
allowing it to drift during contrastive training. For both TDN and TDDN, we
compute the STRUCTURE target from the frozen DINOv3 representation, allowing
TDDN's CleanDIFT features to contribute through the learned fusion
pathway while preserving DINOv3 as the geometric anchor.

\paragraph{Frozen and trainable components.}
Table~\ref{tab:frozen_trainable} lists the alignment modules. Both models freeze DINOv3 and RoBERTa; TDDN also freezes CleanDIFT. We train attention and projection modules, including TDDN's per-layer and post-fusion MLPs; learned layer projections replace vision-only PCA for diffusion features. Both models use approximately 80M trainable parameters.


\begin{table}[!t]
\centering
\caption{\textbf{Training hyperparameters.} TDN and TDDN share all
settings except the CleanDIFT-specific entries, which apply only to TDDN.}
\label{tab:training_recipe}

\setlength{\tabcolsep}{3pt}
\renewcommand{\arraystretch}{1.05}

\begin{minipage}[t]{0.485\linewidth}
\vspace{0pt}
\centering

\begin{tabularx}{0.98\linewidth}{
  >{\raggedright\arraybackslash}X
  >{\raggedleft\arraybackslash}p{0.27\linewidth}
}
\toprule
\textbf{Hyperparameter} & \textbf{Value} \\
\midrule

\rowcolor{gray!15}
\multicolumn{2}{l}{\emph{Architecture}} \\

Image embedding ($z_{\mathrm{img}}$) & 2560 \\
Text embedding ($d_t$) & 1024 \\
Input resolution & $336\times336$ \\
Patch tokens ($N$) & 441 \\
Attention block depth ($K$) & 2 \\
Vision heads / head dim & $20/64$ \\
Text heads / head dim & $16/64$ \\
FFN & SwiGLU \\
Drop-path & 0.1 \\
LayerScale initialization & $10^{-5}$ \\
Vision RoPE base ($\theta$) & 100 \\
CleanDIFT extraction layers & $\{2,5,8\}$ \\
CleanDIFT projection dimension & 512 \\
CleanDIFT common grid & $21\times21$ \\

\midrule

\rowcolor{gray!15}
\multicolumn{2}{l}{\emph{Objective}} \\

InfoNCE temperature ($\tau$) & 0.05 \\
STRUCTURE temperature ($\tau_s$) & 0.05 \\
STRUCTURE weight ($\lambda$) & 10 \\
STRUCTURE warm-up iterations & 500 \\
Label smoothing & 0 \\

\bottomrule
\end{tabularx}
\end{minipage}
\hfill
\begin{minipage}[t]{0.485\linewidth}
\vspace{0pt}
\centering

\begin{tabularx}{0.98\linewidth}{
  >{\raggedright\arraybackslash}X
  >{\raggedleft\arraybackslash}p{0.46\linewidth}
}
\toprule
\textbf{Hyperparameter} & \textbf{Value} \\
\midrule

\rowcolor{gray!15}
\multicolumn{2}{l}{\emph{Optimization}} \\

Optimizer & AdamW \\
$(\beta_1,\beta_2,\epsilon)$ & $(0.9,0.95,10^{-8})$ \\
Weight decay & $10^{-4}$ \\
Peak learning rate & $10^{-3}$ \\
Minimum learning rate & $10^{-5}$ \\
Warm-up iterations & 500 \\
Total iterations & 5{,}000 \\
Learning-rate schedule & Cosine \\
Gradient clipping ($\ell_2$) & 1.0 \\

\midrule

\rowcolor{gray!15}
\multicolumn{2}{l}{\emph{Batching \& Hardware}} \\

Per-GPU batch size & 64 \\
GradCache accumulation & 16 \\
GPUs & $4\times$ A100 \\
Effective contrastive batch size & 4096 \\
Precision & bf16 (fp32 reduction) \\
Sharding & FSDP + activation checkpointing \\

\bottomrule
\end{tabularx}
\end{minipage}
\end{table}

\section{Additional Experimental Setup}

In this section, we describe the vision--language alignment architecture, optimization, and compute requirements, followed by evaluation protocols for open-vocabulary segmentation, zero-shot and few-shot classification, and Contrastive Region Guidance. These protocols specify datasets, feature readouts, prompts, adaptation procedures, region masking, and metrics.

\subsection{Implementation of Vision-Language Alignment}\label{sec:impl_details}
We use a DINOv3 ViT-H/16+ image backbone with $d_n=1280$ at an input resolution of $336\times336$, yielding $N=441$ visual tokens, and RoBERTa-Large as the text backbone with $d_t=1024$ (layer-24 features). We use $K=2$ self-attention blocks for both modalities. The vision branch uses $20$ attention heads, whereas the text branch uses $16$ attention heads (head dimension $64$ in both cases), together with SwiGLU feed-forward networks, a drop-path rate of $0.1$, and a LayerScale initialization of $10^{-5}$. We additionally apply rotary positional embeddings (RoPE) with base $\theta=100$ to the vision branch.

For TDDN, we extract CleanDIFT activations from decoder layers ${2,5,8}$ of a frozen CleanDIFT UNet (Stable Diffusion 2.1 architecture) at $t=0$ using an empty text prompt. We project each layer independently to a common dimension, $d_c=512$, and resample the feature maps to a shared $21\times21$ grid before applying the fusion procedure defined in Eq.~\ref{eq:fusion}.

We set the contrastive temperature to $\tau=0.05$ and the STRUCTURE temperature to $\tau_s=0.05$. We use $\lambda=10$ and apply a $500$-iteration warm-up to $\mathcal{L}_{\mathrm{struct}}$. We optimize all trainable parameters using AdamW ($\beta_1=0.9$, $\beta_2=0.95$, $\epsilon=10^{-8}$, and weight decay $10^{-4}$) for $5{,}000$ iterations. We use a peak learning rate of $10^{-3}$ with linear warm-up over the first $500$ iterations, followed by cosine decay to $10^{-5}$, and apply gradient clipping with a maximum norm of $1.0$.

We train all models on $4\times$ NVIDIA A100 GPUs (80\,GB) using FSDP, bfloat16 parameters, and fp32 gradient reduction. We use GradCache to accumulate $16$ micro-batches of size $64$ per GPU, resulting in an effective contrastive batch size of $4{,}096$. Across both TDN and TDDN, since we keep all pretrained backbones frozen and train only the vision/text alignment modules, it results in approximately 80M trainable parameters. Text-alignment training requires approximately $15$ hours. We summarize all hyperparameters in Table~\ref{tab:training_recipe}.

\subsection{Open-Vocabulary Segmentation}
\label{app:ovseg}

We evaluate open-vocabulary segmentation on ADE20K~\citep{ade20k}, Cityscapes~\citep{cityscapes}, COCO-Stuff~\citep{cocostuff}, and PASCAL Context (C59)~\citep{pascalcontext}, reporting mean Intersection-over-Union (mIoU). We construct a text classifier from the respective class names and assign each patch to its highest-similarity class following the standard sliding-window protocol of TCL~\citep{tcl}.

Most vision--language models, including CLIP~\citep{clip}, SigLIP\,2~\citep{tschannen2025siglip}, and \tmethod{}, do not explicitly supervise final patch representations during pretraining. Their final patch embeddings are therefore unsuitable for dense prediction. We use the standard MaskCLIP~\citep{maskclip} readout for these models, extracting final-layer value embeddings without attention pooling.

This yields patch features in the text-aligned representation space. We apply the same readout across all models in this group so that performance differences reflect the underlying representations rather than the feature extraction procedure. FG-CLIP\,2~\citep{xie2025fg2} is the sole exception. It provides a dedicated dense-feature head trained explicitly for localization tasks, which we use as intended. Consequently, its results directly include an additional task-specific dense prediction component unavailable to the other models considered here.

\subsection{Zero-shot and Few-shot Experiments}

\paragraph{Zero-shot Classification.}

Following CLIP~\citep{clip}, we encode each class name using the standard 80 OpenAI prompt templates (e.g., \textit{``a photo of a {class}''}). We pass each prompt through the text encoder and average the resulting $\mathbf{z_\mathrm{txt}}$ embeddings within a class to obtain a classifier $W \in \mathbb{R}^{C\times2560}$. The $c$-th row corresponds to the prototype embedding for class $c$. We predict the class label by selecting the prototype with the highest cosine similarity to $\mathbf{z_\mathrm{img}}$. The remaining variants explore two orthogonal directions for improving zero-shot performance: richer text supervision and limited visual supervision.

\paragraph{Text supervision with CuPL prompts.}
Generic prompt templates compress each class into a short phrase and often omit visually distinguishing details. CuPL~\citep{cupl} addresses this limitation by generating multiple visual descriptions per class using a large language model. We follow this approach and generate 50 descriptions per class using GPT-4.1, equally divided between the following prompt templates:

\begin{itemize}\itemsep2pt\parsep0pt
\item \emph{"Describe what \{class\} looks like in a photograph. Focus on visual appearance---colors, textures, shapes, and distinguishing features. One paragraph, 2--3 sentences only.''}
    \item \emph{"What visual features distinguish \{class\} from other similar objects or scenes? Focus on what makes it visually unique. One paragraph, 2--3 sentences only."}
\end{itemize}

For zero-shot classification, we prepend each description with the literal prefix \texttt{``a photo of {class}.''} before encoding it with the text encoder. We then compute the class prototype $\mathbf{u}_c\in\mathbb{R}^{2d_n}$ as the $\ell_2$-normalized mean of the resulting embeddings and predict the class label using $\arg\max\limits_{c}\,
\mathbf{z}_{\mathrm{img}}^{\top}\mathbf{u}_{c}$.

Both prompt variants yield one prototype per class. CuPL enriches text supervision while keeping the image representation and similarity-based decision rule fixed; few-shot adaptation adds labeled images.

\paragraph{Few-shot Classification using Tip-Adapter.}
Tip-Adapter~\citep{tipadapter} improves visual recognition using a small set of labeled examples per class. It constructs a cache of image embeddings and their corresponding labels and combines cache-based similarity scores with the standard text-based predictions.

For few-shot evaluation, we use the training-free Tip-Adapter framework of \citet{tipadapter}. 
For each shot setting $K\in{1,2,4,8,16}$, we sample $K$ training examples per class using a fixed random seed and construct a cache of $\ell_2$-normalized image embeddings and labels. During inference, we combine the cache-based scores with the text logits. We select the cache scaling parameter independently for each dataset and shot setting and fix the cache similarity parameter to $\beta=5.5$. 

To ensure consistency and comparability across shot settings, we construct a shared pool using the $K=16$ examples per class sampled with random seed $42$. Lower-shot caches are then obtained by selecting the first $k$ examples from this shared pool for each class. Consequently, the few-shot subsets are nested: every example used in a lower-shot setting is also present in all higher-shot settings. This construction reduces variation due to independent sampling and makes differences across shot settings more directly attributable to the amount of available labeled data.

\subsection{Contrastive Region Guidance (CRG) Experiment}
\label{app:crg}

\begin{table}[!t]
\centering
\caption{\textbf{Perception questions used in the CRG experiment.} All questions are multiple choice; the \textbf{ablated region} denotes the piece(s) blacked out in the CRG negative. From \PuzzlePerception{}, we use 8 \textsc{Chess} questions spanning five categories, each evaluated on 100 balanced boards, and 4 \textsc{N-Queens} questions that are coordinate-free and extreme-referenced, evaluated on 100 boards.
}
\label{tab:crg-questions}

\setlength{\tabcolsep}{3pt}
\renewcommand{\arraystretch}{1.0}

\begin{tabularx}{0.98\linewidth}{
  >{\raggedright\arraybackslash}p{0.14\linewidth}
  >{\raggedright\arraybackslash}X
  >{\raggedright\arraybackslash}p{0.16\linewidth}
  >{\raggedright\arraybackslash}p{0.20\linewidth}
}
\toprule
\textbf{Category}
& \textbf{Question}
& \textbf{Options}
& \textbf{Ablated region} \\
\midrule

\rowcolor{gray!15}
\multicolumn{4}{l}{\PuzzlePerception{} (\textsc{Chess})} \\

localization
& Which quarter of the board is the white queen in? (A=top-left,
B=top-right, C=bottom-left, D=bottom-right)
& A/B/C/D
& white queen \\

localization
& Which quarter of the board is the white bishop in? (A/B/C/D as above)
& A/B/C/D
& white bishop \\

localization
& Is the leftmost white pawn in the top half or the bottom half?
& top/bottom
& leftmost white pawn \\

relation
& Is the white king above or below the black knight?
& above/below
& white king, black knight \\

geometric
& Are the white rook and the black king in the same row or the same column?
& Yes/No
& white rook, black king \\

geometric
& Are the white king and the black king on adjacent (touching) squares?
& Yes/No
& white king, black king \\

tactical
& Is the white queen on the same rank, file, or diagonal as the black king?
& Yes/No
& white queen \\

existence
& Is there a white queen on the board?
& Yes/No
& white queen \\

\midrule

\rowcolor{gray!15}
\multicolumn{4}{l}{\PuzzlePerception{} (\textsc{N-Queens})} \\

localization
& Is the leftmost queen in the top half or the bottom half?
& top/bottom
& leftmost queen \\

localization
& Is the topmost queen in the left half or the right half?
& left/right
& topmost queen \\

localization
& Is the leftmost queen in the top, middle, or bottom third?
& top/mid/bot
& leftmost queen \\

relation
& Is the leftmost queen above or below the rightmost queen?
& above/below
& leftmost \& rightmost queen \\

\bottomrule
\end{tabularx}
\end{table}

\paragraph{Datasets.}
We evaluate CRG on the \textsc{Chess} and \textsc{N-Queens} subsets of \PuzzlePerception{}, comprising short multiple-choice perception questions about the board (Table~\ref{tab:crg-questions}). We exclude Maze because its start marker is positionally fixed and its walls and paths occupy most of the board, making region guidance uninformative. 

For \textsc{Chess}, we evaluate 8 questions spanning five categories---localization, relation, geometric, tactical, and existence---on a balanced set of 100 boards per question (800 boards in total). The \textsc{N-Queens} evaluation comprises 4 questions on 100 boards, using the leftmost, rightmost, or topmost queen as a reference rather than explicit square coordinates.

The probing questions in Table~\ref{tab:crg-questions} use quarter-, half-, and third-board choices for localization, directional choices for relative position, and binary choices for presence, adjacency, and alignment. Each question is paired with a blackout region targeting the evidence needed to answer it: localization and existence questions target a single piece, whereas relation and geometric questions target both referenced pieces. For the tactical Chess question, only the white queen is removed, leaving the black king visible. This pairing allows CRG to test how disrupting specific visual evidence changes a model's answer. Uniform-guessing baselines are $50\%$, $33.3\%$, and $25\%$ for two-, three-, and four-option questions, respectively. Comparing image-only and guided conditions within each question fixes the answer space and isolates the effect of region guidance.

\newcommand{\pvqacell}[4][0pt]{%
  \begin{minipage}[t][\dimexpr0.5\textheight-24pt\relax][t]{\dimexpr0.5\linewidth-2\tabcolsep-1.5\arrayrulewidth\relax}
    \vspace{0pt}

    \vspace*{6pt}
    \noindent\includegraphics[width=\linewidth,keepaspectratio]{#2}\par
    \vspace{6pt}
    \nointerlineskip

    \raggedright
    \ifdim#1>0pt\relax
      \renewcommand{\bluecaption}[1]{%
        \colorbox{skyblue}{%
          \parbox[t][#1][t]{\dimexpr\linewidth-2\fboxsep\relax}{\strut ##1\strut}%
        }%
      }%
      \renewcommand{\orangecaption}[1]{%
        \colorbox{lightorange}{%
          \parbox[t][\baselineskip][t]{\dimexpr\linewidth-2\fboxsep\relax}{\strut ##1\strut}%
        }%
      }%
    \fi
    #3\par

    \vspace{0.3em}
    \vfill

    \raggedright
    #4\par

    \vspace*{6pt}
  \end{minipage}%
}

\begin{table}[!tp]
\centering
\caption{Chess examples from \textsc{PuzzlePerception} (part 1 of 2), showing the board image
alongside its question (Q) and ground-truth answer (A).}
\label{tab:pvqa_examples}

\setlength{\tabcolsep}{6pt}
\renewcommand{\arraystretch}{0}

\begin{tabular}{|c|c|}
\hline

\pvqacell[3\baselineskip]
{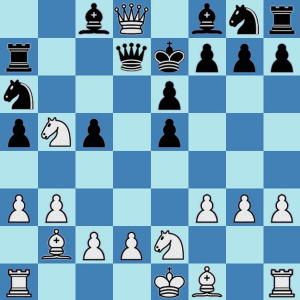}
{\bluecaption{\textbf{Q:} Which quarter of the board is the
white queen in? (A=top-left, B=top-right, C=bottom-left, D=bottom-right)}}
{\orangecaption{\textbf{A:} A}}
&
\pvqacell[3\baselineskip]
{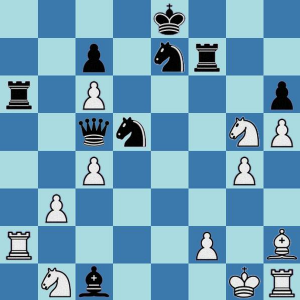}
{\bluecaption{\textbf{Q:} Which quarter of the board is the
white bishop in? (A=top-left, B=top-right, C=bottom-left, D=bottom-right)}}
{\orangecaption{\textbf{A:} D}}
\\
\hline

\pvqacell[3\baselineskip]
{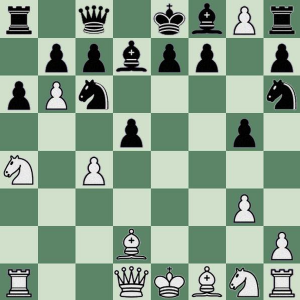}
{\bluecaption{\textbf{Q:} Is the leftmost white pawn in the top half or the
bottom half of the board?}}
{\orangecaption{\textbf{A:} Top}}
&
\pvqacell[3\baselineskip]
{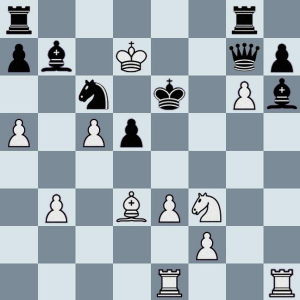}
{\bluecaption{\textbf{Q:} Is the white king above or below the black knight?}}
{\orangecaption{\textbf{A:} Above}}
\\
\hline

\end{tabular}
\end{table}

\begin{table}[!tp]
\centering
\caption{Chess examples from \textsc{PuzzlePerception} (part 2 of 2), showing the board image
alongside its question (Q) and ground-truth answer (A).}
\label{tab:pvqa_examples_chess_2}

\setlength{\tabcolsep}{6pt}
\renewcommand{\arraystretch}{0}

\begin{tabular}{|c|c|}
\hline

\pvqacell[3\baselineskip]
{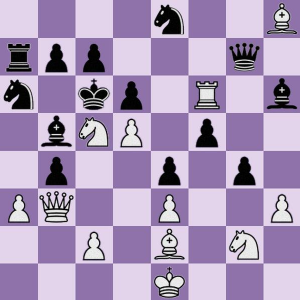}
{\bluecaption{\textbf{Q:} Are the white rook and the black king in the same
row or the same column?}}
{\orangecaption{\textbf{A:} Yes}}
&
\pvqacell[3\baselineskip]
{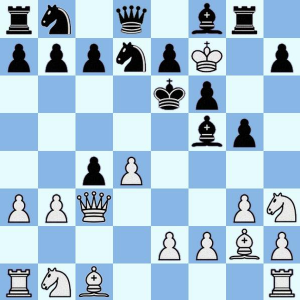}
{\bluecaption{\textbf{Q:} Are the white king and the black king on adjacent
(touching) squares?}}
{\orangecaption{\textbf{A:} Yes}}
\\
\hline

\pvqacell[3\baselineskip]
{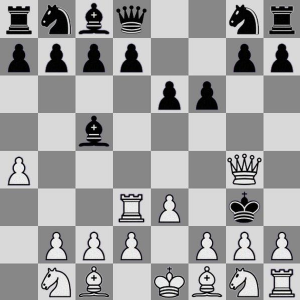}
{\bluecaption{\textbf{Q:} Is the white queen on the same rank, file, or
diagonal as the black king?}}
{\orangecaption{\textbf{A:} Yes}}
&
\pvqacell[3\baselineskip]
{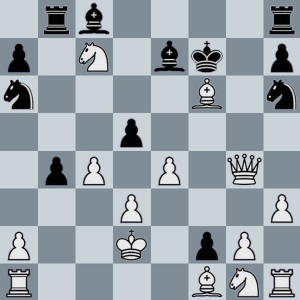}
{\bluecaption{\textbf{Q:} Is there a white queen on the board?}}
{\orangecaption{\textbf{A:} Yes}}
\\
\hline

\end{tabular}
\end{table}

\begin{table}[!tp]
\centering
\caption{N-Queens examples from \textsc{PuzzlePerception}, showing the board image
alongside its question (Q) and ground-truth answer (A).}
\label{tab:pvqa_examples_nqueens}

\setlength{\tabcolsep}{6pt}
\renewcommand{\arraystretch}{0}

\begin{tabular}{|c|c|}
\hline

\pvqacell[3\baselineskip]
{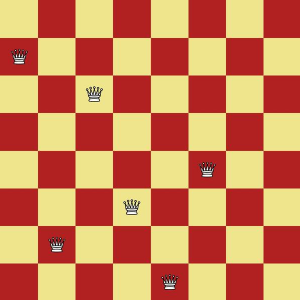}
{\bluecaption{\textbf{Q:} Is the leftmost queen in the top half or the
bottom half of the board?}}
{\orangecaption{\textbf{A:} Top}}
&
\pvqacell[3\baselineskip]
{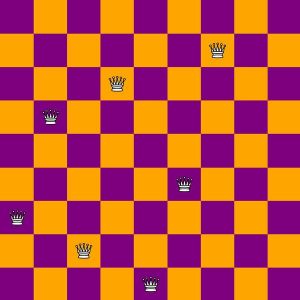}
{\bluecaption{\textbf{Q:} Is the topmost queen in the left half or the
right half of the board?}}
{\orangecaption{\textbf{A:} Right}}
\\
\hline

\pvqacell[3\baselineskip]
{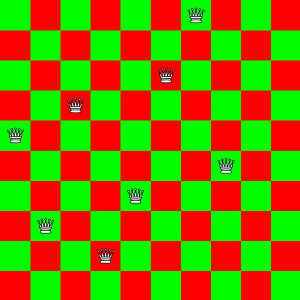}
{\bluecaption{\textbf{Q:} Is the leftmost queen in the top, middle, or
bottom third of the board?}}
{\orangecaption{\textbf{A:} Mid}}
&
\pvqacell[3\baselineskip]
{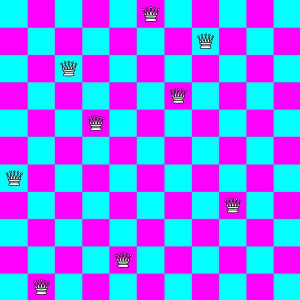}
{\bluecaption{\textbf{Q:} Is the leftmost queen above or below the
rightmost queen?}}
{\orangecaption{\textbf{A:} Above}}
\\
\hline

\end{tabular}
\end{table}

Tables~\ref{tab:pvqa_examples} and~\ref{tab:pvqa_examples_chess_2} show eight Chess image--question--answer examples in two $2\times2$ grids; Table~\ref{tab:pvqa_examples_nqueens} shows four N-Queens examples in a $2\times2$ grid. Each cell pairs the rendered board with its question and ground-truth answer, so readers can inspect the requested visual evidence directly. The Chess examples range from locating a queen, bishop, or pawn within a board region to comparing the positions of two pieces, testing adjacency or line alignment, and identifying whether a white queen is present. The Chess questions also require the model to distinguish white from black pieces. The N-Queens examples use extremal pieces as reference points and ask about halves, thirds, or relative vertical order. Different board colors and piece arrangements illustrate variation in appearance and spatial configuration while each question specifies a precise target. The examples also distinguish answer formats: quarter-board queries use lettered choices, relative-position queries use directional terms, and presence or alignment queries use binary answers. Quarter and half queries use the board boundaries as a reference frame, while above--below and adjacency queries compare one piece with another. For these spatial queries, recognizing piece types alone is insufficient; the model must also recover the relevant locations and relationships. Together, these cases assess recognition and localization of visible board content without requiring a move sequence or a complete puzzle solution.

\paragraph{Experimental setup.}
Contrastive Region Guidance (CRG)~\citep{Wan2024CRG} steers a frozen VLM by contrasting its predictions on the original image with those obtained after blacking out the queried region, following the contrastive scoring procedure proposed by \citet{Wan2024CRG}. Because CRG requires white-box access to model logits, we evaluate only open-weight VLMs, which we run locally in bf16 (using model parallelism for the largest models). We disable model-specific reasoning extensions for all models to ensure a fair comparison and isolate visual perception. We evaluate eight VLMs spanning three model families~-~Qwen2.5-VL/Qwen3.6, Gemma-3/4, and InternVL3. For each puzzle, we compare three conditions: \textbf{Img} (the original image), \textbf{CRG-oracle} ($\Delta_{\mathrm{o}}$; ground-truth region mask), and \textbf{CRG-${\mathrm{TDDN}}$} ($\Delta_{\mathrm{TDDN}}$; the region predicted by \tmethod{} used as a zero-shot text-promptable segmenter). Since \tmethod{} receives no puzzle segmentation supervision, $\Delta_{\mathrm{TDDN}}$ measures whether vision--language alignment alone produces sufficiently accurate regions to steer the VLM.

The oracle condition tests whether each VLM benefits from contrastive guidance with known object locations. Comparing it with the predicted-region condition helps distinguish limited use of spatial evidence from errors in text-conditioned localization. We treat oracle guidance as a reference, since accurate masks alone do not guarantee improved perception.

\paragraph{Region blackout with \tmethod{}.}
Given a queried object, we prompt \tmethod{} with a short textual description and obtain its patch-level predictions, yielding a coarse text-conditioned segmentation of the board. We then black out the region assigned to the queried object while leaving the remainder of the image unchanged. For the oracle condition, we apply the same procedure using the ground-truth object location instead of the prediction produced by \tmethod{}.

\paragraph{Metrics.}
Every question is multiple choice and graded by accuracy, macro-averaged over a task's questions. We report raw accuracy (Img) and the CRG changes $\Delta_{\text{o}}$ (oracle) and $\Delta_{\mathrm{TDDN}}$ (\textsc{tddn}) relative to Img.

\FloatBarrier

\section{Additional Experimental Results}
In this section, we analyze global representation quality, report additional open-vocabulary segmentation results, and assess zero-shot and few-shot classification. We then compare oracle and \tmethod{}-predicted regions for guiding VLM perception and visualize patch-level features on additional puzzles.

\subsection{Global-Level Representation Analysis}
\label{app:global}
\begin{figure}[!t]
  \centering
  \begin{minipage}[b]{0.39\linewidth}
    \centering
    \includegraphics[width=\linewidth]{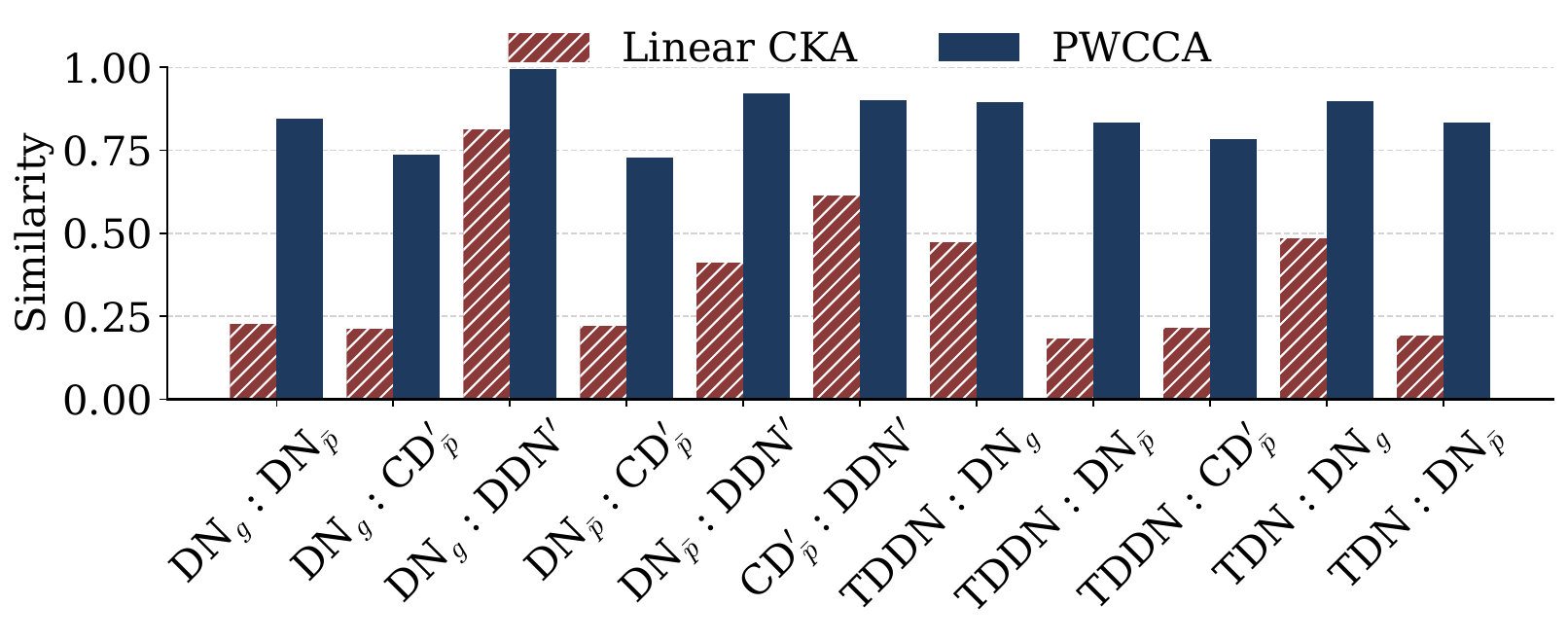}\\[2pt]
    (a) Global similarity between encoders
  \end{minipage}
  \hfill
  \begin{minipage}[b]{0.28\linewidth}
    \centering
    \includegraphics[width=\linewidth]{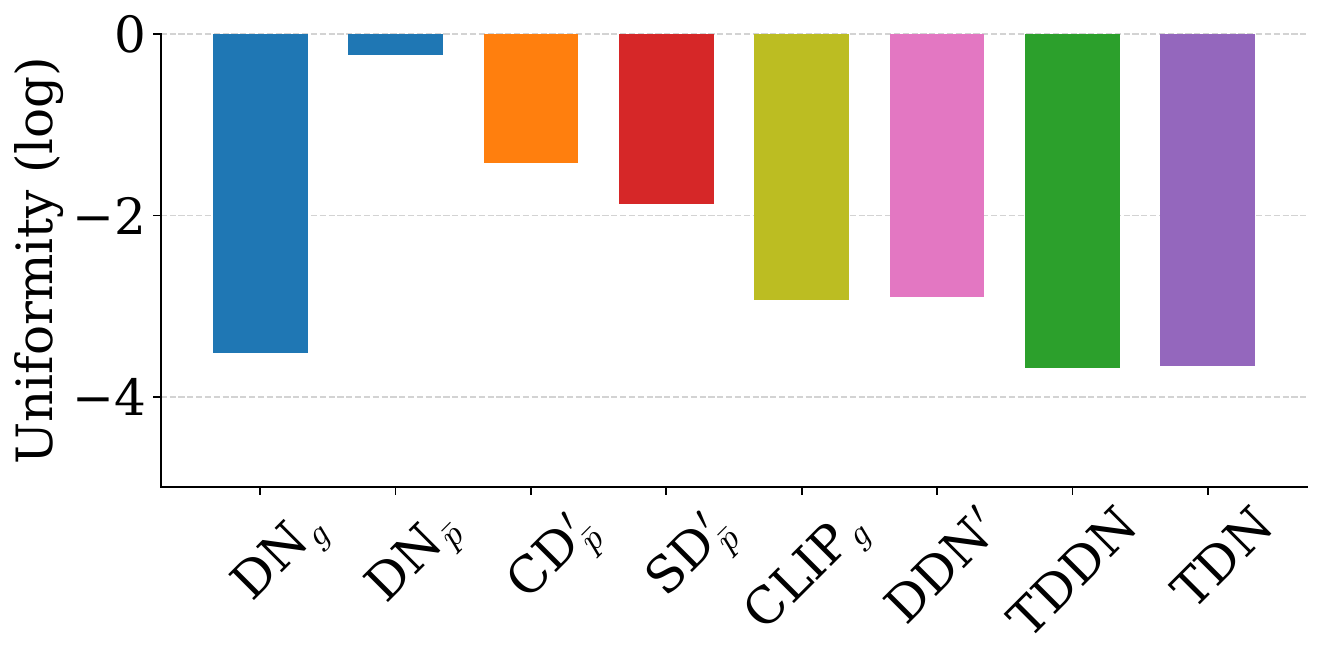}\\[2pt]
    (b) Uniformity
  \end{minipage}
  \hfill
  \begin{minipage}[b]{0.28\linewidth}
    \centering
    \includegraphics[width=\linewidth]{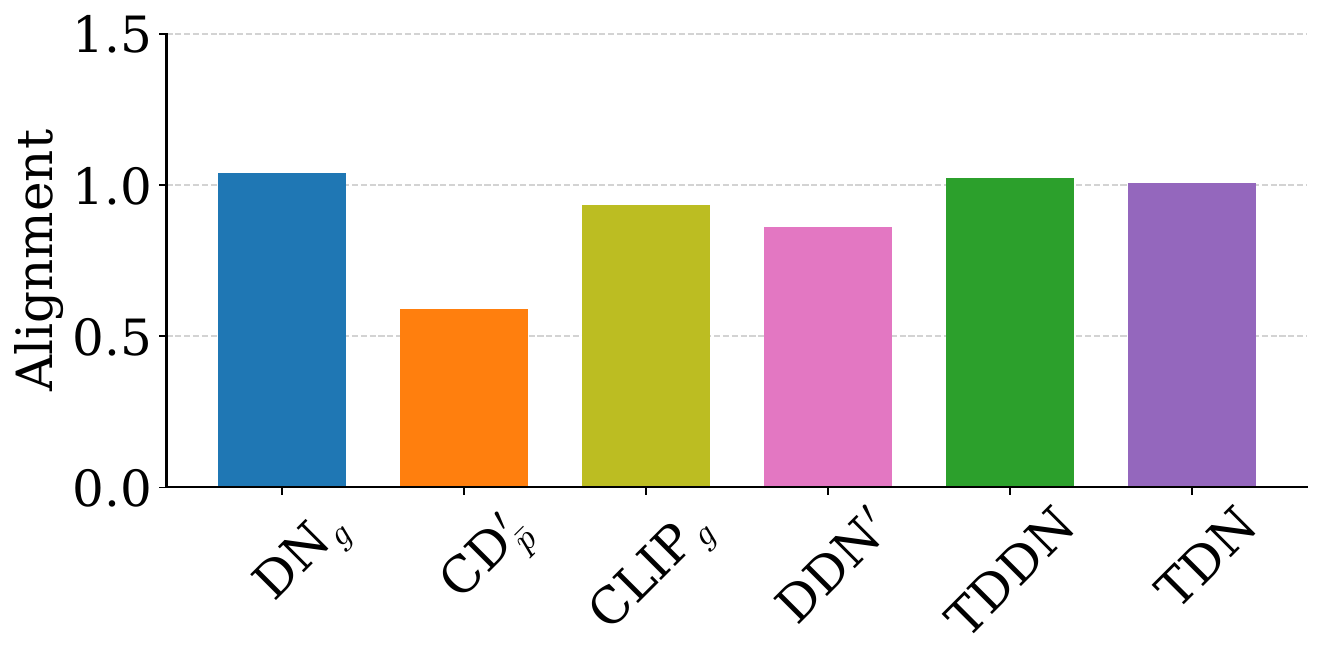}\\[2pt]
    (c) Alignment
  \end{minipage}

  \caption{
    \textbf{Representational similarity and quality analysis of
    $\mathrm{DDN}$, $\mathrm{DN}$, and $\mathrm{CD}$ features
    (Linear CKA and PWCCA on $5{,}000$ images) across pooling strategies.}
    Subscripts $g$ and $p$ denote global (CLS) and patch-mean pooling.
    Lower ($\downarrow$) is better for (b) and (c).
  }
  \label{fig:global_similarity_analysis}
\end{figure}

At the global-level, the fused representation remains closest to DINOv3 rather than CleanDIFT (Fig.~\ref{fig:global_similarity_analysis}). Specifically, $\mathrm{DDN}'$ attains CKA~$0.81$ and PWCCA~$0.99$ with $\mathrm{DN}$, compared to $0.61$ and $0.90$ with $\mathrm{CD}'$. This behavior complements the patch-level analysis in the main text, where $\mathrm{DDN}'$ is closest to $\mathrm{CD}'$ (CKA~$0.96$, PWCCA~$1.00$). Together, these results indicate that the fused representation divides roles across granularities: $\mathrm{CD}'$ supplies fine-grained patch structure, whereas $\mathrm{DN}$ anchors global semantics.

For discriminability, we again consider Alignment and Uniformity jointly (lower is better on both). $\mathrm{DDN}'$ closely tracks $\mathrm{DN}$ on both axes, indicating that fusion preserves DINOv3's image-level geometry without sacrificing its discriminative properties. After text alignment, $\mathrm{TDDN}$ and $\mathrm{TDN}$ show the increased uniformity expected from the InfoNCE objective~\citep{wang2020uniformity}, consistent with the patch-level behavior reported in the main text.

This preservation is reflected in downstream image-level tasks. On ImageNet-1k $k$-nearest-neighbor classification (Table~\ref{tab:quantitative_results}), $\mathrm{DDN}'$ achieves $83.80$ top-1 accuracy, substantially outperforming CLIP ($73.10$), while $\mathrm{TDDN}$ retains $76.08$ after text alignment. The performance of both variants indicates that learned visual features remain highly discriminative across categories. 

These results demonstrate that vision-language alignment preserves the semantic geometry of the fused representation while enabling competitive retrieval performance with substantially fewer image--text pairs than CLIP.

\FloatBarrier

\subsection{Open-Vocabulary Segmentation Results}
\label{app:ovseg_results}

Table~\ref{tab:main_results} reports zero-shot open-vocabulary segmentation results on ADE20K, COCO-Stuff, and PASCAL-Context, using the evaluation protocol described in Appendix~\ref{app:ovseg}. \tmethod{} improves over CLIP and TDN on all three benchmarks and achieves the highest mIoU on COCO-Stuff.

\begin{table}[!htbp]
    \centering
    \caption{\textbf{Comparison of TDN and \tmethod{} with baselines on zero-shot open-vocabulary segmentation (Seg)}. Higher mIoU ($\uparrow$) is better. \textbf{Bold} = best in column.}
    \label{tab:main_results}
    \setlength{\tabcolsep}{4pt}
    \begin{tabular}{l ccc}
        \toprule
        \multirow{3}{*}{\textbf{Model}}
        & \textbf{ADE20K} & \textbf{COCO-Stuff} & \textbf{PASCAL-Ctx} \\
        \cmidrule(lr){2-2}\cmidrule(lr){3-3}\cmidrule(lr){4-4}
        & \textbf{Seg} & \textbf{Seg} & \textbf{Seg} \\
        & (mIoU$\uparrow$) & (mIoU$\uparrow$) & (mIoU$\uparrow$) \\
        \midrule
        CLIP ViT-L/14 \citep{clip} & 5.20 & 7.35 & 10.44 \\

        OpenCLIP ViT-L/14 \citep{cherti2023scaling} & 2.76 & 5.39 & 8.09 \\

        MetaCLIP ViT-L/14 \citep{xu2024metaclip} & 7.57 & 10.59 & 13.92 \\

        DFN ViT-L/14 \citep{fang2024dfn} & 4.30 & 4.55 & 8.63 \\

        SigLIP\,2 ViT-L/16 \citep{tschannen2025siglip} & 16.97 & 17.38 & 23.44 \\

        FG-CLIP\,2 \citep{xie2025fg2} & \textbf{23.61} & 24.17 & \textbf{35.47} \\


        \midrule

        \rowcolor{gray!25}
        TDN (ours) & 16.51 & 20.67 & 27.55 \\

        \rowcolor{gray!25}
        \textbf{TDDN (ours)} & 18.11 & \textbf{24.44} & 32.48 \\
        \bottomrule
    \end{tabular}
\end{table}

\subsection{Zero-shot and Few-shot experiments}
\label{app:zeroshot_fewshot}

\begin{table}[!t]
\centering
\begingroup
\centering
\caption{Zero-shot top-1 classification accuracy. TIP-Adapter uses
$K{=}16$ shots with template-based text. $\dagger$~Our methods.
\textbf{Bold} = best in column. Higher ($\uparrow$) is better.}
\label{tab:classification_tasks}

\setlength{\tabcolsep}{4pt}
\renewcommand{\arraystretch}{1.05}

\begin{tabular*}{\linewidth}{
  @{\extracolsep{\fill}}lcccc@{}
}
\toprule
\textbf{Model}
& \textbf{CIFAR-100}
& \textbf{Caltech-101}
& \textbf{Food-101}
& \textbf{GTSRB} \\
\midrule

CLIP ViT-L/14 \citep{clip}
& 77.79 & 86.49 & 92.04 & 39.06 \\

OpenCLIP ViT-L/14 \citep{cherti2023scaling}
& 83.81 & 91.23 & 88.91 & 39.90 \\

MetaCLIP ViT-L/14 \citep{xu2024metaclip}
& 84.24 & 90.17 & 91.69 & \textbf{48.61} \\

DFN ViT-L/14 \citep{fang2024dfn}
& \textbf{88.40} & 93.03 & 92.58 & 47.60 \\

SigLIP\,2 ViT-L/16 \citep{tschannen2025siglip}
& 82.17 & \textbf{93.93} & \textbf{94.45} & 48.38 \\

FG-CLIP\,2 \citep{xie2025fg2}
& 81.85 & 91.13 & 93.85 & 33.65 \\

\midrule

TDN$\dagger$
& 75.09 & 77.61 & 51.81 & 8.30 \\

TDDN$\dagger$
& 78.40 & 79.66 & 51.30 & 9.40 \\

\midrule

TDN + CuPL$\dagger$
& 75.70 & 78.30 & 63.80 & 14.80 \\

TDDN + CuPL$\dagger$
& 75.30 & 77.70 & 66.80 & 11.10 \\

TDN + TIP-Adapter$\dagger$
& 85.70 & 89.60 & 84.70 & 41.60 \\

TDDN + TIP-Adapter$\dagger$
& 86.80 & 90.50 & 86.20 & 36.30 \\

\bottomrule
\end{tabular*}
\par
\endgroup
\end{table}

\paragraph{Zero-shot classification.}
Table~\ref{tab:classification_tasks} summarizes results on four classification benchmarks. On CIFAR-100, our model exceeds CLIP ($\mathrm{TDDN}$: $78.40$, $\mathrm{TDN}$: $75.09$, CLIP: $77.79$). On Food-101, Caltech-101, and GTSRB, CLIP and recent encoders (SigLIP\,2~\citep{tschannen2025siglip}, DFN~\citep{fang2024dfn}, MetaCLIP~\citep{xu2024metaclip}, OpenCLIP~\citep{cherti2023scaling}, FG-CLIP\,2~\citep{xie2025fg2}) retain an advantage, consistent with their larger and more diverse pretraining corpora relative to our $590$K image--text pairs. The largest gap is on GTSRB ($9.4$ vs.\ CLIP's $39.0$), comprising German traffic signs---a concept largely absent from our training captions.
This is expected because zero-shot classification requires broad semantic coverage across diverse categories, whereas image--text retrieval primarily tests whether images and captions share an embedding space. Our smaller alignment corpus therefore enables competitive cross-modal alignment but lacks the diversity needed to match models trained on hundreds of millions of pairs across all classification benchmarks.

We focus the remaining analysis on CLIP, which most closely matches our training paradigm. To test the data-coverage hypothesis, Table~\ref{tab:seen_classes} reports top-1 accuracy on classes observed during text alignment. A class is \emph{Seen} if all content-word lemmas in its name occur at least 10 times in the alignment corpus. We lemmatize class names and captions and construct an inverted index to identify sufficiently represented concepts.

\begin{table}[!t]
\begingroup
\centering
\caption{\textbf{Seen-class zero-shot accuracy} (\%). We define a class as
\emph{Seen} if all content-word lemmas in its name occur at least 10 times in
the text-alignment corpus. We report top-1 accuracy using CuPL prompts.
$\Delta$ denotes the performance difference between TDDN and CLIP
(in percentage points).}
\label{tab:seen_classes}

\setlength{\tabcolsep}{4pt}
\renewcommand{\arraystretch}{1.1}

\begin{tabular*}{0.82\linewidth}{
  @{\extracolsep{\fill}}lccccc@{}
}
\toprule
\textbf{Dataset}
& \textbf{Seen}
& \textbf{TDDN}
& \textbf{TDN}
& \textbf{CLIP}
& \textbf{$\Delta$} \\
\midrule

Food-101
& 22 / 101
& \textbf{94.0}
& 92.7
& 93.8
& \textbf{+0.2} \\

CIFAR-100
& 63 / 100
& 88.2
& \textbf{88.3}
& 78.9
& \textbf{+9.3} \\

Caltech-101
& 65 / 101
& \textbf{91.3}
& 88.4
& 88.1
& \textbf{+3.2} \\

\bottomrule
\end{tabular*}
\par
\endgroup
\end{table}

\begin{table}[!t]
\centering
\caption{\textbf{Few-shot classification with Tip-Adapter}
(top-1 accuracy, \%). Best per (dataset, $K$) in bold.}
\label{tab:fewshot_tip}

\setlength{\tabcolsep}{4pt}
\renewcommand{\arraystretch}{1.05}

\begin{tabular}{lcccccccccc}
\toprule
& \multicolumn{5}{c}{\textbf{TDDN}}
& \multicolumn{5}{c}{\textbf{TDN}} \\
\cmidrule(lr){2-6}
\cmidrule(lr){7-11}
\textbf{Dataset}
& $K{=}1$ & $K{=}2$ & $K{=}4$ & $K{=}8$ & $K{=}16$
& $K{=}1$ & $K{=}2$ & $K{=}4$ & $K{=}8$ & $K{=}16$ \\
\midrule
Food-101
& $\mathbf{74.6}$ & $\mathbf{78.9}$ & $\mathbf{82.9}$
& $\mathbf{84.9}$ & $\mathbf{86.2}$
& $72.5$ & $76.9$ & $80.7$ & $83.2$ & $84.7$ \\
\midrule
CIFAR-100
& $\mathbf{84.7}$ & $\mathbf{85.2}$ & $\mathbf{85.8}$
& $\mathbf{86.5}$ & $\mathbf{86.8}$
& $78.9$ & $80.5$ & $82.9$ & $84.8$ & $85.7$ \\
\midrule
Caltech-101
& $\mathbf{86.9}$ & $\mathbf{87.6}$ & $\mathbf{88.4}$
& $\mathbf{89.6}$ & $\mathbf{90.5}$
& $86.1$ & $86.7$ & $87.3$ & $88.8$ & $89.6$ \\
\midrule
GTSRB
& $19.1$ & $27.2$ & $29.2$ & $32.0$ & $36.3$
& $\mathbf{24.2}$ & $\mathbf{33.9}$ & $\mathbf{37.4}$
& $\mathbf{38.7}$ & $\mathbf{41.6}$ \\
\bottomrule
\end{tabular}
\end{table}

On these subsets, $\mathrm{TDDN}$ exceeds CLIP by $0.2$, $9.3$, and $3.2$ percentage points on Food-101, CIFAR-100, and Caltech-101, respectively, using CuPL prompts. The subsets contain $22/101$, $63/100$, and $65/101$ classes, so they probe different fractions of each benchmark. These results support the role of semantic coverage, while the different class compositions limit direct comparison with full-benchmark accuracy. The lexical criterion measures caption coverage; it does not establish equal visual exposure to every retained class.

\paragraph{Few-shot adaptation.}
To examine how much of the remaining gap stems from limited semantic coverage, we equip $\mathrm{TDDN}$ with TIP-Adapter~\citep{tipadapter} using $K{=}16$ shots and template-based prompts. We report results for all shot settings in Table~\ref{tab:fewshot_tip}. The adapter recovers most of the deficit and surpasses CLIP on two of the four tasks: $\mathrm{TDDN}{+}\text{TIP-Adapter}$ reaches $86.8$ on CIFAR-100, $86.2$ on Food-101, $90.5$ on Caltech-101, and $36.3$ on GTSRB, compared to CLIP's $39.0$.

These results support the data-coverage hypothesis discussed above. A small amount of task-specific supervision is sufficient to close most of the performance gap, suggesting that the primary limitation of $\mathrm{TDDN}$ on zero-shot classification is not the quality of its visual representations, but the limited diversity of the image--text pairs used during alignment.

\subsection{Region-guided VLM Perception with \tmethod{}}
Table~\ref{tab:crg} reports the CRG comparison on the Chess and N-Queens splits of Puzzle Perception. Here, $\Delta_{\mathrm{o}}$ and $\Delta_{\mathrm{TDDN}}$ denote the accuracy gains obtained using oracle and \tmethod{}-predicted regions relative to the image-only baseline.

CRG improves performance across a diverse set of VLMs, although the magnitude of the gains varies considerably across models and tasks. On Chess, the largest oracle gains are observed for Gemma-4-12B-it ($+6.4$), Qwen2.5-VL-7B-Instruct ($+5.6$), and Qwen3.6-27B ($+4.6$), while on N-Queens, Qwen3.6-27B and Qwen3.6-35B-A3B benefit the most ($+7.0$ each). Several models exhibit improvements on both splits, including Qwen2.5-VL-7B-Instruct and Qwen3.6-27B, indicating that region guidance provides useful spatial cues across architectures and parameter scales.

However, the benefits are not uniform across tasks. Gemma-4-12B-it achieves substantial gains on Chess ($+6.4$/$+2.6$, baseline $60.0$) but remains flat on N-Queens ($-0.8$/$-1.2$, baseline $75.0$). Conversely, Qwen3.6-35B-A3B experiences a slight degradation on Chess ($-2.1$/$-3.8$, baseline $79.9$) yet improves substantially on N-Queens ($+7.0$/$+8.8$, baseline $58.0$). These results suggest that the utility of region guidance depends jointly on the model and the underlying task.

Finally, oracle and \tmethod{} guidance often track one another, although the gap between them varies across models and tasks. These results suggest that the utility of region guidance depends on both region quality and the VLM's ability to exploit region-level information.

\begin{table}[!t]
\centering
\caption{\textbf{Contrastive Region Guidance (CRG) with TDDN-predicted
regions.} VLM question accuracy (\%) on Puzzle Perception, macro-averaged
over questions. \textbf{Img} is the image-only baseline.
$\Delta_{\mathrm{o}}$ and $\Delta_{\mathrm{TDDN}}$ denote the accuracy gains
from CRG using oracle and TDDN-predicted regions, respectively. Standard
errors computed over bootstrap samples of the boards are shown in smaller
font. The final columns report the mean gains across Chess and N-Queens.}
\label{tab:crg}

\setlength{\tabcolsep}{2pt}
\renewcommand{\arraystretch}{1.05}

\begin{tabular*}{\linewidth}{
  @{\extracolsep{\fill}}lcccccccc@{}
}
\toprule
& \multicolumn{3}{c}{Chess}
& \multicolumn{3}{c}{N-Queens}
& \multicolumn{2}{c}{\textbf{Avg}} \\
\cmidrule(lr){2-4}
\cmidrule(lr){5-7}
\cmidrule(lr){8-9}

\textbf{Model}
& Img
& $\Delta_{\mathrm{o}}$
& $\Delta_{\mathrm{TDDN}}$
& Img
& $\Delta_{\mathrm{o}}$
& $\Delta_{\mathrm{TDDN}}$
& $\Delta_{\mathrm{o}}$
& $\Delta_{\mathrm{TDDN}}$ \\
\midrule

Qwen2.5-VL-7B-Instruct
& 48.4
& $+5.6$\stderr{1.1}
& $+3.8$\stderr{1.2}
& 51.0
& $+4.0$\stderr{1.3}
& $+1.8$\stderr{1.4}
& $+4.8$
& $+2.8$ \\

InternVL3-8B
& 48.0
& $+3.2$\stderr{1.1}
& $+2.1$\stderr{1.1}
& 49.8
& $+2.0$\stderr{1.6}
& $+1.2$\stderr{1.4}
& $+2.6$
& $+1.6$ \\

Gemma-4-12B-it
& 60.0
& $+6.4$\stderr{1.2}
& $+2.6$\stderr{1.4}
& 75.0
& $-0.8$\stderr{1.8}
& $-1.2$\stderr{1.8}
& $+2.8$
& $+0.7$ \\

Gemma-4-26B-A4B-it
& 72.8
& $+2.5$\stderr{1.0}
& $+1.1$\stderr{1.0}
& 78.0
& $+2.5$\stderr{1.3}
& $+3.0$\stderr{1.3}
& $+2.5$
& $+2.0$ \\

Gemma-3-27B-it
& 57.8
& $+1.7$\stderr{0.9}
& $-0.6$\stderr{0.8}
& 48.2
& $+0.8$\stderr{1.5}
& $+2.2$\stderr{1.4}
& $+1.2$
& $+0.8$ \\

Qwen3.6-27B
& 80.1
& $+4.6$\stderr{1.1}
& $+2.2$\stderr{1.1}
& 68.8
& $+7.0$\stderr{1.9}
& $+6.5$\stderr{1.9}
& $+5.8$
& $+4.3$ \\

\midrule

Gemma-4-31B-it
& 81.1
& $+1.5$\stderr{0.9}
& $-0.4$\stderr{1.0}
& 78.0
& $+2.2$\stderr{1.1}
& $+1.5$\stderr{0.9}
& $+1.9$
& $+0.6$ \\

Qwen3.6-35B-A3B
& 79.9
& $-2.1$\stderr{1.3}
& $-3.8$\stderr{1.3}
& 58.0
& $+7.0$\stderr{2.3}
& $+8.8$\stderr{2.2}
& $+2.5$
& $+2.5$ \\

\bottomrule
\end{tabular*}
\end{table}

\FloatBarrier

\subsection{PCA Visualization of Puzzle Image Features}
\label{app:pca_vis}
\begin{figure}[!p]
  \centering

  \newcommand{\figseparator}{%
    \par\vspace{8.75pt}
    \noindent\makebox[\linewidth][c]{%
      \makebox[\linewidth]{%
        \leaders\hbox{\rule{4pt}{0.4pt}\hspace{3pt}}\hfill\kern0pt
      }%
    }%
    \par\vspace{8.75pt}
  }

  \begin{subfigure}[c]{\textwidth}
    \centering

    \begin{minipage}[c]{0.22\linewidth}
      \centering
      \includegraphics[width=\linewidth]
        {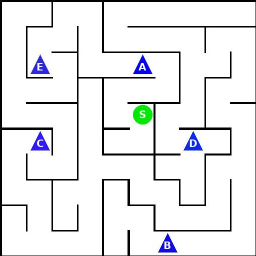}\\[2pt]
      {(a) Maze (Input Image)}
    \end{minipage}
    \hfill
    \begin{minipage}[c]{0.74\linewidth}
      \centering
      \setlength{\tabcolsep}{2pt}
      \renewcommand{\arraystretch}{0.8}

      \begin{tabular}{cccc}
        \includegraphics[width=0.22\linewidth]
          {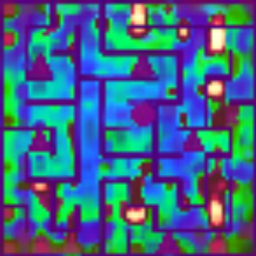} &
        \includegraphics[width=0.22\linewidth]
          {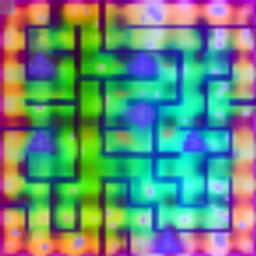} &
        \includegraphics[width=0.22\linewidth]
          {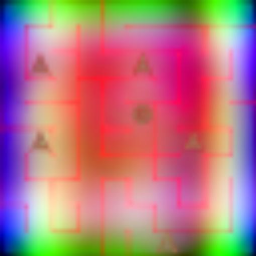} &
        \includegraphics[width=0.22\linewidth]
          {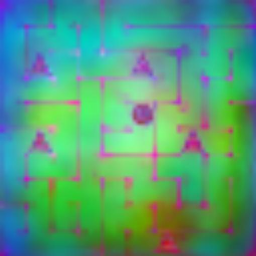} \\

        {(a.1) CLIP} &
        {(a.2) DINOv2} &
        {(a.3) DINOv3} &
        {(a.4) TDN} \\[2pt]

        \includegraphics[width=0.22\linewidth]
          {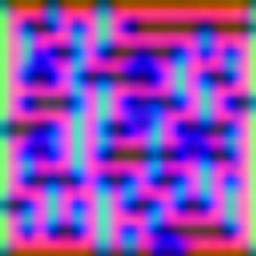} &
        \includegraphics[width=0.22\linewidth]
          {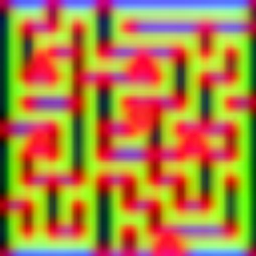} &
        \includegraphics[width=0.22\linewidth]
          {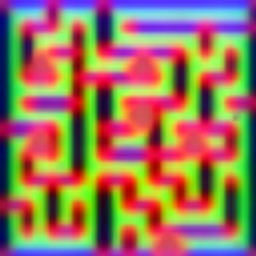} &
        \includegraphics[width=0.22\linewidth]
          {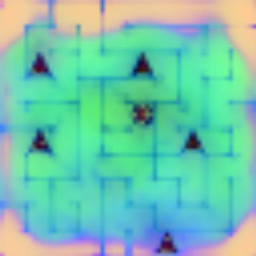} \\

        {(a.5) Stable Diffusion} &
        {(a.6) CleanDIFT} &
        {(a.7) \method{}} &
        {(a.8) \tmethod{}}
      \end{tabular}
    \end{minipage}

    \phantomcaption
    \label{fig:pca_examples_maze}
  \end{subfigure}

  \figseparator

  \begin{subfigure}[c]{\textwidth}
    \centering

    \begin{minipage}[c]{0.22\linewidth}
      \centering
      \includegraphics[width=\linewidth]
        {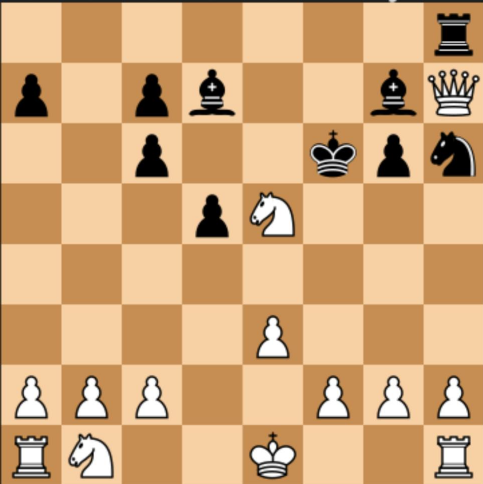}\\[2pt]
      {(b) Chess (Input Image)}
    \end{minipage}
    \hfill
    \begin{minipage}[c]{0.74\linewidth}
      \centering
      \setlength{\tabcolsep}{2pt}
      \renewcommand{\arraystretch}{0.8}

      \begin{tabular}{cccc}
        \includegraphics[width=0.22\linewidth]
          {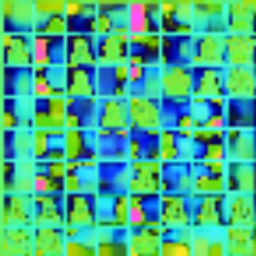} &
        \includegraphics[width=0.22\linewidth]
          {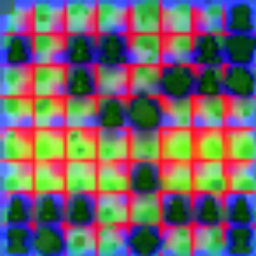} &
        \includegraphics[width=0.22\linewidth]
          {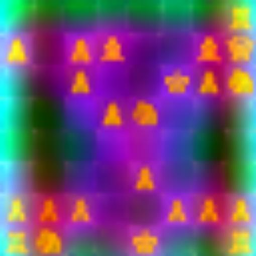} &
        \includegraphics[width=0.22\linewidth]
          {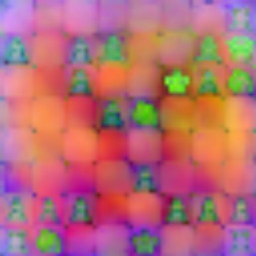} \\

        {(b.1) CLIP} &
        {(b.2) DINOv2} &
        {(b.3) DINOv3} &
        {(b.4) TDN} \\[2pt]

        \includegraphics[width=0.22\linewidth]
          {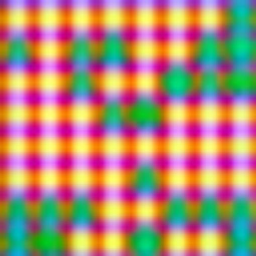} &
        \includegraphics[width=0.22\linewidth]
          {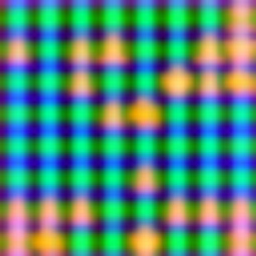} &
        \includegraphics[width=0.22\linewidth]
          {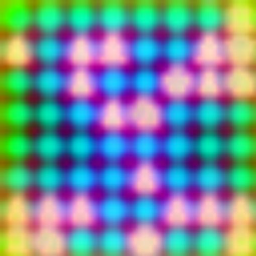} &
        \includegraphics[width=0.22\linewidth]
          {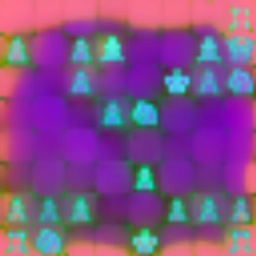} \\

        {(b.5) Stable Diffusion} &
        {(b.6) CleanDIFT} &
        {(b.7) \method{}} &
        {(b.8) TDDN}
      \end{tabular}
    \end{minipage}

    \phantomcaption
    \label{fig:pca_examples_chess}
  \end{subfigure}

  \figseparator

  \begin{subfigure}[c]{\textwidth}
    \centering

    \begin{minipage}[c]{0.22\linewidth}
      \centering
      \includegraphics[width=\linewidth]
        {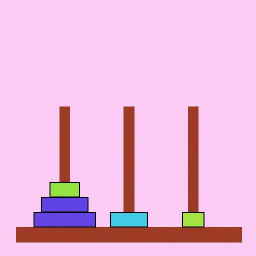}\\[2pt]
      {(c) Tower-of-Hanoi (Input Image)}
    \end{minipage}
    \hfill
    \begin{minipage}[c]{0.74\linewidth}
      \centering
      \setlength{\tabcolsep}{2pt}
      \renewcommand{\arraystretch}{0.8}

      \begin{tabular}{cccc}
        \includegraphics[width=0.22\linewidth]
          {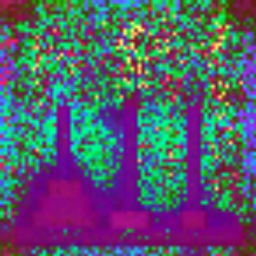} &
        \includegraphics[width=0.22\linewidth]
          {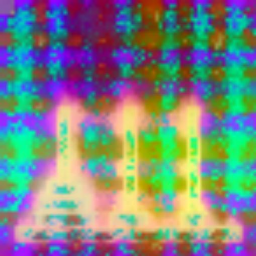} &
        \includegraphics[width=0.22\linewidth]
          {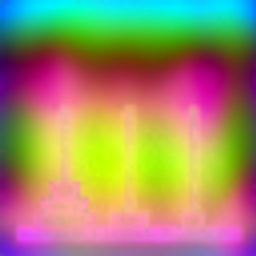} &
        \includegraphics[width=0.22\linewidth]
          {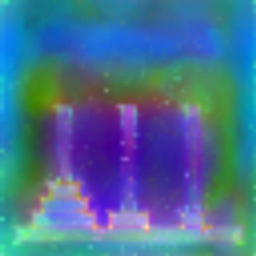} \\

        {(c.1) CLIP} &
        {(c.2) DINOv2} &
        {(c.3) DINOv3} &
        {(c.4) TDN} \\[2pt]

        \includegraphics[width=0.22\linewidth]
          {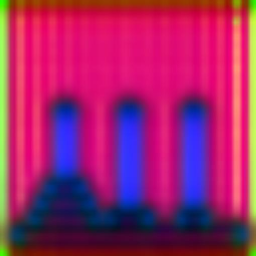} &
        \includegraphics[width=0.22\linewidth]
          {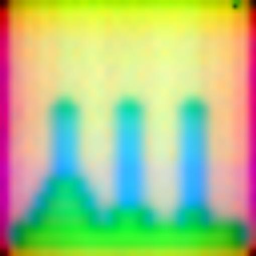} &
        \includegraphics[width=0.22\linewidth]
          {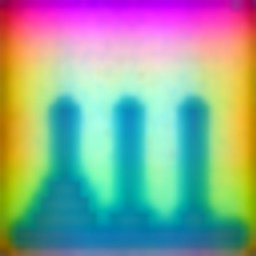} &
        \includegraphics[width=0.22\linewidth]
          {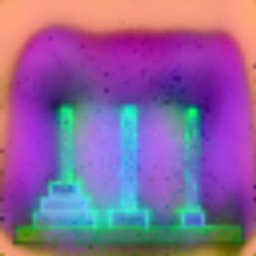} \\

        {(c.5) Stable Diffusion} &
        {(c.6) CleanDIFT} &
        {(c.7) \method{}} &
        {(c.8) TDDN}
      \end{tabular}
    \end{minipage}

    \phantomcaption
    \label{fig:pca_examples_toh}
  \end{subfigure}

  \caption{
    PCA visualization of patch-level features on
    (a) maze, (b) chess, and (c) Tower-of-Hanoi input images at
    $512 \times 512$ resolution.
    Each model's patch features have shape $(N^2, z_{\mathrm{img}})$,
    where $N=32$ for a patch size of $16$.
    The features are projected onto the top three principal components,
    yielding an $(N,N,3)$ map that is mapped to RGB and bilinearly
    interpolated to the input resolution.
    Sub-indices denote feature visualizations from the corresponding models.
  }
  \label{fig:pca_examples}
\end{figure}

Figure~\ref{fig:pca_examples} compares patch-level features across three puzzles: wall continuity in Maze, piece--board separation in Chess, and disc boundaries in Tower-of-Hanoi.

\paragraph{Maze.}
Figure~\ref{fig:pca_examples_maze} contains thin walls, connected corridors, and triangular and circular markers. CLIP and DINOv2 reveal wall segments but introduce local color variation, reducing region uniformity. DN produces smoother regions while weakening fine wall geometry. Stable Diffusion highlights the layout, and CD provides cleaner wall--corridor separation at turns and junctions. DDN retains this structure, consistent with the complementary geometric and semantic signals described by \citet{taleoftwo}. After text alignment, TDN and \tmethod{} make the markers more distinct, although wall contrast remains weaker than in CD and DDN. These maps distinguish sensitivity to small objects from sensitivity to maze connectivity.

\paragraph{Chess.}
Figure~\ref{fig:pca_examples_chess} combines a regular checkerboard with small pieces. Stable Diffusion and CD emphasize the alternating board pattern, whereas DN highlights occupied locations against smoother backgrounds. DINOv2 retains board structure and local changes around pieces; CLIP shows more fragmented variation. DDN preserves the grid and local features at occupied squares. TDN and \tmethod{} produce more recognizable piece-shaped regions, with \tmethod{} retaining distinct silhouettes across the upper and lower groups. This comparison highlights the need to separate objects from repetitive texture: recovering the grid alone does not identify the pieces occupying each square.

\paragraph{Tower-of-Hanoi.}
Figure~\ref{fig:pca_examples_toh} contains three pegs, a multi-disc stack on the left, and individual discs on the other pegs. CLIP and DINOv2 show substantial variation outside the objects, while DN forms broad color bands that blur object boundaries. Stable Diffusion and CD recover the peg arrangement and stack outline; DDN retains this foreground structure with smoother regions. After text alignment, \tmethod{} reveals horizontal boundaries within the stack and preserves separate discs near the other pegs. Its stack profile shows clearer internal separation than TDN's. These boundaries indicate disc placement and relative stack height, without establishing correct counting or move selection.

PCA projections are computed separately for each encoder, so matching colors need not indicate matching features or objects. We compare region coherence, boundary continuity, and object separation within each map, rather than interpreting colors as semantic classes. These examples illustrate how fusion and text alignment affect spatial structure; quantitative task metrics establish whether those features support accurate predictions.

\FloatBarrier

\begin{table}[!t]
\centering
\caption{
Public training and evaluation datasets grouped by task category.
For vision--language alignment, we train exclusively on MS-COCO-2014
($\sim$82K images) and ReLAION-COCOA, a $\sim$508K-image subset of
LAION-COCO-Aesthetic that we recaption. For evaluation across the different
tasks, we use the corresponding test splits of the remaining datasets.
}

\label{tab:eval_datasets}

\setlength{\tabcolsep}{3pt}
\renewcommand{\arraystretch}{1.1}

\begin{tabularx}{\linewidth}{
  >{\raggedright\arraybackslash}p{0.21\linewidth}
  >{\raggedright\arraybackslash}p{0.19\linewidth}
  c r r
  >{\raggedright\arraybackslash}X
}
\toprule
\textbf{Task}
& \textbf{Dataset}
& \textbf{Classes}
& \textbf{Train}
& \textbf{Test}
& \textbf{License} \\
\midrule

\multirow{2}{*}{\shortstack[l]{Vision--Language\\Alignment}}
& MS-COCO-2014
& --
& 82{,}783
& 40{,}504
& CC-BY 4.0 (annot.); Flickr ToU \\

& ReLAION-COCOA
& --
& 508{,}025
& --
& CC-BY 4.0 \\

\midrule

Retrieval (T2I / I2T)
& Flickr30K
& --
& 29{,}783
& 5{,}000
& Flickr ToU (research) \\

\midrule

\multirow{5}{*}{Segmentation}
& ADE20K
& 150
& 20{,}210
& 2{,}000
& BSD-3-Clause \\

& COCO-Stuff
& 171
& 118{,}000
& 5{,}000
& CC BY 4.0 \\

& Cityscapes
& 19
& 2{,}975
& 500
& - \\

& Puzzle Perception
& 30
& 6{,}000
& 1{,}500
& - \\

& PASCAL-Context
& 59
& 4{,}998
& 5{,}104
& PASCAL-VOC \\

\midrule

Keypoint Matching
& SPair-71K
& 18
& 53{,}340
& 12{,}234
& PASCAL-VOC + Flickr ToU \\

\midrule

\multirow{5}{*}{Classification}
& Food-101
& 101
& 75{,}750
& 25{,}250
& Foodspotting ToU (research) \\

& CIFAR-100
& 100
& 50{,}000
& 10{,}000
& MIT \\

& Caltech-101
& 101
& 3{,}060
& 6{,}084
& CC-BY 4.0 \\

& GTSRB
& 43
& 26{,}640
& 12{,}630
& CC-BY-SA 4.0 \\

& ImageNet-1K
& 1000
& 1{,}281{,}167
& 50{,}000
& N/A \\

\bottomrule
\end{tabularx}
\end{table}

\section{Additional Dataset Details}
\label{app:additional_dataset_details}

In this section, we summarize the training and evaluation datasets and compare \PuzzlePerception{} with existing segmentation and visual question answering benchmarks. We then describe the visual variations and semantic labels for Maze, Chess, Tower-of-Hanoi, and N-Queens.

\subsection{Training and Evaluation Data}

Table~\ref{tab:eval_datasets} summarizes the training and evaluation datasets we use throughout this work, including their sizes, class counts, and licenses. For the Puzzle Perception benchmark (Table~\ref{tab:puzzle_perception_dataset}), we obtain segmentation masks directly from the rendering pipeline, which provides noise-free, pixel-level annotations for the puzzle variations we describe above. This enables controlled evaluation of fine-grained spatial representations without annotation ambiguity. We additionally include PuzzleVQA-style question--answer pairs for Chess and N-Queens and use these puzzles for our Contrastive Region Guidance (CRG) evaluation (Section~\ref{sec:crg}), which assesses whether region-level visual information can support perception-grounded multiple-choice reasoning. 
We use Maze and Tower-of-Hanoi for dense prediction evaluation, complementing the reasoning-based tasks. For Chess, paired annotations support segmentation and question answering; for N-Queens, object locations support region-guided question answering. This distinction helps interpret cases in which a model recognizes individual objects but struggles to use their arrangement to answer a question.

For vision--language alignment, we use MS-COCO-2014 \citep{coco} together with our ReLAION-COCOA subset, yielding approximately 590K image--text pairs in total. We construct ReLAION-COCOA with approximately 508K images from LAION-COCO-Aesthetic \citep{li_laion_coco_aesthetic}. We retain samples with aesthetic scores above $0.5$, remove Not Safe for Work (NSFW) content, and recaption the retained images using Gemma-3 to improve image--text correspondence. This compact alignment corpus allows us to evaluate whether we can align strong visual representations with language without relying on web-scale training data. We use these data exclusively to align the frozen visual backbones, without additional puzzle-specific supervision.

\subsection{Puzzle Perception vs Other Datasets}

Compared with existing segmentation and visual question answering (VQA) datasets, Puzzle Perception bridges dense visual perception and puzzle-level understanding within a unified benchmark (Table~\ref{tab:dataset_comparison}). Natural-image datasets such as COCO-Stuff and ADE20K provide pixel-level segmentation annotations but do not evaluate question answering or puzzle understanding, whereas VQA benchmarks such as CLEVR and GQA assess visual reasoning without dense segmentation supervision. Puzzle-oriented datasets, including PuzzleVQA and AlgoPuzzleVQA, further target puzzle understanding and reasoning, but lack pixel-level image annotations. 

In contrast, Puzzle Perception combines dense semantic segmentation with VQA-based Puzzle Understanding (PZU), enabling models to be evaluated on both fine-grained visual grounding and puzzle-level understanding over the same structured scenes. This combination makes it possible to study whether representations that accurately localize puzzle entities also support higher-level understanding of their configurations and relationships.

\begin{table}[!t]
\centering
\caption{
Comparison of segmentation (Seg), visual question answering
(VQA), reasoning, and puzzle understanding (PZU) datasets.
Puzzle Perception pairs pixel-level segmentation with puzzle VQA.}
\label{tab:dataset_comparison}

\setlength{\tabcolsep}{4pt}
\renewcommand{\arraystretch}{1.1}

\begin{tabular*}{0.82\linewidth}{
  @{\extracolsep{\fill}}lcccc@{}
}
\toprule
\textbf{Dataset}
& \textbf{Seg}
& \textbf{VQA}
& \textbf{Reasoning}
& \textbf{PZU} \\
\midrule

COCO-Stuff
& \cmark & \xmark & \xmark & \xmark \\

ADE20K
& \cmark & \xmark & \xmark & \xmark \\

CLEVR
& \xmark & \cmark & \cmark & \xmark \\

GQA
& \xmark & \cmark & \cmark & \xmark \\

PuzzleVQA
& \xmark & \cmark & \cmark & \cmark \\

AlgoPuzzleVQA
& \xmark & \cmark & \cmark & \cmark \\

\midrule

\textbf{Puzzle Perception (Ours)}
& \cmark & \cmark & \xmark & \cmark \\

\bottomrule
\end{tabular*}
\end{table}

\subsection{Variations in Puzzle Perception}
For each puzzle type, we systematically vary its visual appearance and configuration while preserving its underlying structure and semantics, yielding diverse instances for evaluating puzzle perception. The variations alter the rendered instance rather than the definition of a semantic label. Consequently, each task retains a consistent set of prediction targets across its visual styles and configurations.

\paragraph{Maze.}
We vary wall appearance across plain and natural textures (\emph{grass}, \emph{rock}, \emph{brick}, and \emph{steel}) while preserving the underlying topology. We consider 8 semantic classes: \emph{wall}, \emph{path}, \emph{start}, and five color-coded \emph{destinations}. This separates the layout of traversable paths from the texture used to render obstacles.

\paragraph{Chess.}
We vary board configurations and rendering styles, combining procedurally generated $2$D boards with $3$D ChessRender360~\citep{kojic2024chessrender360} renderings. We consider 15 semantic classes: \emph{background}, \emph{light} and \emph{dark} squares, and six piece types in both colors. The label set distinguishes board squares from pieces and preserves both piece type and color.

\paragraph{Tower-of-Hanoi.}
We vary puzzle configurations with up to five disks per peg. We consider 7 semantic classes: \emph{background}, \emph{peg}, and five size-ordered disks. Disk labels follow size order, so the segmentation target identifies individual disk sizes rather than only occupied regions.

\paragraph{N-Queens.}
We vary board sizes from $8{\times}8$ to $11{\times}11$ and color schemes. We consider two semantic classes: \emph{queen} and \emph{background}. The segmentation label set is binary; the question-answering task separately evaluates the relative placement of queens.

\section{CRG Examples}
\label{app:crg-examples}

In this section, we present the Chess prompt template and qualitative CRG examples for a target VLM~-~InternVL3-8B. We compare image-only, oracle-guided, and \tmethod{}-guided responses to show successful corrections and distinguish localization failures from errors that persist with oracle regions.

\subsection{Prompt template} For all qualitative examples, the target VLM receives a single user message consisting of the board image followed by the text prompt shown below. The template provides a brief description of the board representation, inserts the example-specific question, and constrains the model to answer with exactly one of the listed candidate options.
\noindent
\begin{promptbox}[CRG Prompt Template (Chess)]
\ttfamily
\textcolor[HTML]{1F3B99}{$\langle\,\mathrm{board\ image}\,\rangle$}\\[5pt]
You are looking at an 8x8 chess board. White pieces are light/hollow;
black pieces are dark/solid. Answer directly and concisely.\\[5pt]
\textcolor[HTML]{1E7B34}{\{question\}}\\[2pt]
Answer with only \textcolor[HTML]{9B1C1C}{\{options\}}.\par
\end{promptbox}


\subsection{Qualitative Examples}
\begin{figure}[!tp]
\centering
\renewcommand{\imcell}[2]{%
  \begin{minipage}{0.23\linewidth}
    \centering
    \includegraphics[width=\linewidth]{\crgdir/#1}\\[1pt]
    {\normalsize\textcolor{black!55}{#2}\par}
  \end{minipage}%
}

\begin{subfigure}[t]{\linewidth}
\centering
\crgexample{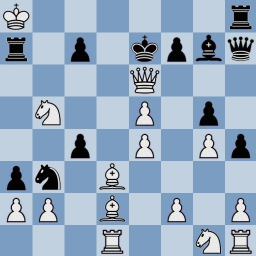}{Chess — Relation (success)}
{Is the white king above or below the black knight?}
{below~\xmark}{above~\cmark}{above~\cmark}{above}
\caption{}
\end{subfigure}

\vspace{4pt}

\begin{subfigure}[t]{\linewidth}
\centering
\crgexample{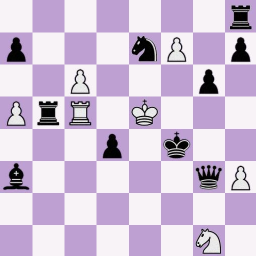}{Chess — Geometric (success)}
{Are the white king and the black king on adjacent (touching) squares?}
{No~\xmark}{Yes~\cmark}{Yes~\cmark}{Yes}
\caption{}
\end{subfigure}

\vspace{4pt}

\begin{subfigure}[t]{\linewidth}
\centering
\crgexample{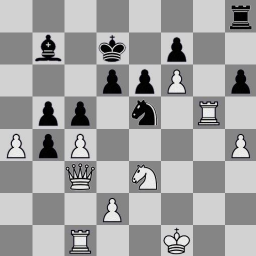}{Chess — Existence (success)}
{Is there a white queen on the board?}
{No~\xmark}{Yes~\cmark}{Yes~\cmark}{Yes}
\caption{}
\end{subfigure}
\caption{\textbf{CRG qualitative examples: successful cases.}
The TDDN-predicted region matches the oracle region, and CRG corrects InternVL3-8B's answer in the relation, geometric, and existence examples.}
\label{fig:crg-qualitative-examples}
\end{figure}

\begin{figure}[!t]
\centering
\renewcommand{\imcell}[2]{%
  \begin{minipage}{0.23\linewidth}
    \centering
    \includegraphics[width=\linewidth]{\crgdir/#1}\\[1pt]
    {\normalsize\textcolor{black!55}{#2}\par}
  \end{minipage}%
}

\begin{subfigure}[t]{\linewidth}
\centering
\crgexample{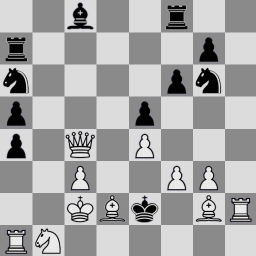}
{Chess — Tactical (oracle-region failure)}
{Is the white queen on the same rank, file, or diagonal as the black king?}
{No~\xmark}{No~\xmark}{No~\xmark}{Yes}
\caption{}
\end{subfigure}

\vspace{4pt}

\begin{subfigure}[t]{\linewidth}
\centering
\crgexample{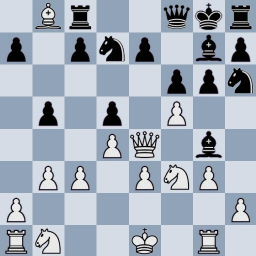}
  {Chess — Localization (TDDN localization failure)}
  {Which quarter of the board is the white bishop in?}
  {C~\xmark}{A~\cmark}{C~\xmark}{A}
\caption{}
\end{subfigure}
\caption{\textbf{CRG qualitative examples: failure cases.}
The tactical example illustrates a case where even the oracle region is insufficient to correct InternVL3-8B's answer. The localization example illustrates TDDN mislocalizing the relevant region, failing where oracle guidance succeeds.}
\label{fig:crg-qualitative-failures}
\end{figure}

Figs. \ref{fig:crg-qualitative-examples} and \ref{fig:crg-qualitative-failures} show successful examples and failure cases, respectively, from the CRG experiment. Each box shows the \emph{source} board, the ground-truth (\emph{oracle}) region
blacked out, and the \emph{TDDN}-predicted region blacked out, with InternVL3-8B's
answer per condition. On the three successes TDDN blacks the same region as the
oracle and CRG flips the answer from wrong to right; we also show two failure modes
--- (i)~even the oracle region does not help, and (ii)~TDDN mislocalizes, failing
where the oracle succeeds.

The relation example changes ``below'' to ``above'' when the model compares the white king with the black knight. The geometric and existence examples likewise change ``No'' to ``Yes'' for king adjacency and white-queen presence. In the tactical example, replacing the predicted mask with the oracle mask still leaves the response incorrect, showing that accurate localization alone does not ensure a correct answer under the same prompt. In the localization example, oracle guidance identifies quarter A, whereas the raw and \tmethod{}-guided responses choose C. The relation and adjacency questions test relationships between pieces, whereas the existence and localization questions concern the presence and position of individual pieces.

The three image variants provide a controlled comparison for each question: the board configuration, question, and answer choices remain fixed while the blacked-out region changes. The image-only response establishes the baseline, oracle guidance tests whether the annotated region can support a correction, and \tmethod{} guidance tests whether a predicted region provides comparable information. When oracle and predicted guidance both change an incorrect answer to the ground-truth option, the predicted region is sufficient for that correction in the displayed example. Matching answers alone, however, do not establish that the VLM follows the same internal reasoning in both conditions.

The two failure categories distinguish errors in supplying the region from errors that persist despite oracle localization. In the tactical example, the oracle-guided response remains incorrect, so improving the predicted mask alone would not resolve the observed failure under the same prompt. In the localization example, the oracle succeeds while \tmethod{} guidance fails, highlighting the effect of a misplaced region on the answer. Together, these cases show why region quality and the VLM's ability to use region-level information must be assessed separately. The examples illustrate these behaviors but do not estimate their prevalence; Table~\ref{tab:crg} reports overall accuracy changes under CRG.


\clearpage

\section{Recaptioning LAION-COCO-Aesthetic Samples}
\label{sec:recaptioning}

\subsection{Prompt Template}
We use the following prompt template to generate new captions for ReLAION-COCOA, comprising a subset of $\sim$508K samples from the LAION-COCO-Aesthetic~\citep{li_laion_coco_aesthetic} dataset. The prompt instructs the model to describe the input image in \emph{20~-~30 words}, while avoiding any \emph{preamble or introduction}.

\begin{promptbox}[Recaptioning Prompt Template]
\ttfamily
\textcolor[HTML]{1F3B99}{$\langle\,\mathrm{image}\,\rangle$}\\[5pt]
Describe this image in 20--30 words. Do not include any preamble or
introduction, just the description.
\end{promptbox}

\pagebreak[4]

\newcommand{\examplecell}[3]{%
  \begin{minipage}[t]{0.315\linewidth}
    \vspace{0pt}

    \begin{minipage}[t][1.29in][t]{\linewidth}
      \centering
      \vspace{0pt}
      \includegraphics[
        width=\linewidth,
        height=1.23in,
        keepaspectratio
      ]{#1}
    \end{minipage}

    \vspace{0.15em}

    \begin{minipage}[t][0.28in][t]{\linewidth}
      \scriptsize
      \raggedright
      #2
    \end{minipage}

    \vspace{0.05em}

    \scriptsize
    \raggedright
    #3\par

    \vspace{0.4em}
  \end{minipage}%
}

\begin{table}[!t]
\centering
\caption{Examples from LAION-COCO-Aesthetic, comparing the
\textbf{original captions} with \textbf{new captions}
(by Gemma-3-27B-it).}
\label{tab:recaptioned_examples}

\setlength{\tabcolsep}{3pt}
\renewcommand{\arraystretch}{1.0}
\renewcommand{\bluecaption}[1]{%
  \setlength{\fboxsep}{1pt}%
  \colorbox{skyblue}{%
    \parbox[t][2\baselineskip][t]{\dimexpr\linewidth-2\fboxsep\relax}{\strut #1\strut}%
  }%
}
\renewcommand{\orangecaption}[1]{%
  \setlength{\fboxsep}{1pt}%
  \colorbox{lightorange}{%
    \parbox[t][4\baselineskip][t]{\dimexpr\linewidth-2\fboxsep\relax}{\strut #1\strut}%
  }%
}

\begin{tabular}{|c|c|c|}
\hline

\examplecell
{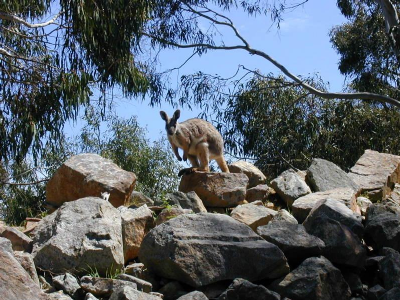}
{\bluecaption{\textbf{Old:} A cat standing on top of rocks in the woods.}}
{\orangecaption{\textbf{New:} A wallaby stands alert atop a rocky outcrop, framed by
eucalyptus trees and a clear blue sky. It appears to be looking directly
at the camera.}}
&
\examplecell
{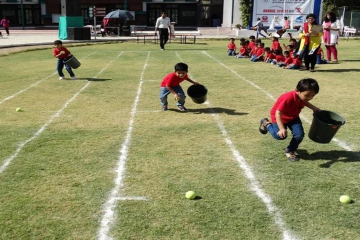}
{\bluecaption{\textbf{Old:} Children playing baseball on the grass in front of spectators.}}
{\orangecaption{\textbf{New:} Young children in red shirts race across a grassy field,
attempting to collect tennis balls in buckets during a school sports day.}}
&
\examplecell
{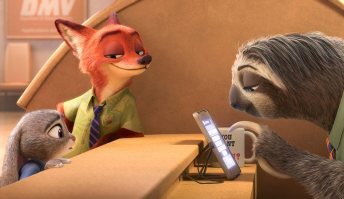}
{\bluecaption{\textbf{Old:} Two sloths sitting at a table with a cell phone.}}
{\orangecaption{\textbf{New:} A fox and rabbit officer face a remarkably slow-moving
sloth working at a DMV counter, creating a humorous, frustrating scene.}}
\\
\hline

\examplecell
{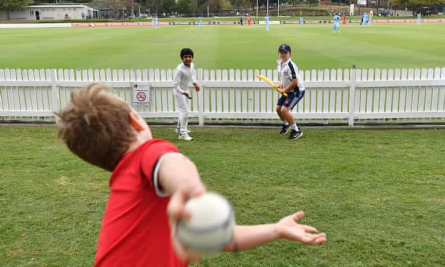}
{\bluecaption{\textbf{Old:} Two boys playing baseball in the grass.}}
{\orangecaption{\textbf{New:} A young bowler in red releases a cricket ball towards a
batter with a raised bat, during a youth cricket match on a green field.}}
&
\examplecell
{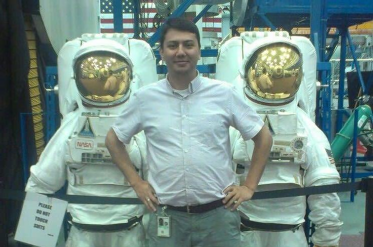}
{\bluecaption{\textbf{Old:} A man standing next to two astronauts in space.}}
{\orangecaption{\textbf{New:} A man stands between two NASA astronaut suits at a museum,
with an American flag visible in the background. A sign requests visitors
not to touch the suits.}}
&
\examplecell
{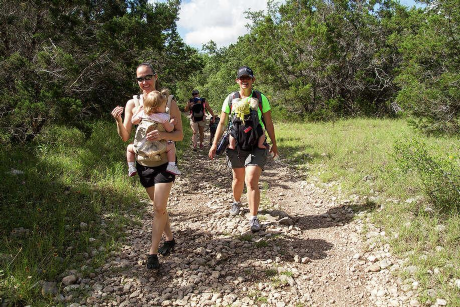}
{\bluecaption{\textbf{Old:} Two women and a baby elephant walking down a trail.}}
{\orangecaption{\textbf{New:} Two women hike a rocky trail, each carrying a baby in a
carrier, with two others following behind them.}}
\\
\hline

\examplecell
{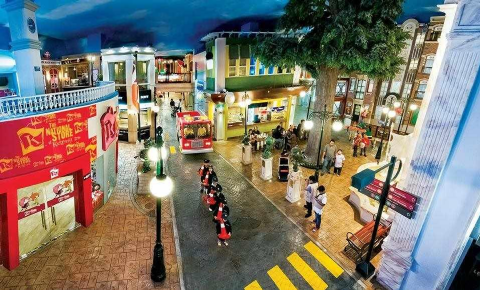}
{\bluecaption{\textbf{Old:} People walking down the street in an Asian shopping mall.}}
{\orangecaption{\textbf{New:} A brightly lit, miniature cityscape features detailed
buildings, a red bus, and marching toy soldiers along a patterned street.}}
&
\examplecell
{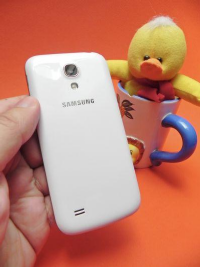}
{\bluecaption{\textbf{Old:} A phone is sitting next to a cup.}}
{\orangecaption{\textbf{New:} A hand holds a white Samsung phone next to a yellow chick
plushie peeking from a patterned mug on an orange background.}}
&
\examplecell
{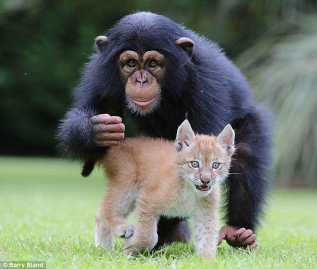}
{\bluecaption{\textbf{Old:} An adult chimpan and its baby in the grass.}}
{\orangecaption{\textbf{New:} A chimpanzee gently holds a small, spotted lynx kitten
while they both walk across a lush green lawn.}}
\\
\hline

\end{tabular}
\end{table}

\subsection{Recaptioned Examples}
Table~\ref{tab:recaptioned_examples} contrasts inaccurate or generic LAION-COCO-Aesthetic captions with Gemma-3-27B-it recaptions. Corrections include a ``cat standing on top of rocks'' becoming a ``wallaby on a rocky outcrop,'' ``playing baseball'' becoming a youth cricket match, and ``an Asian shopping mall'' becoming a miniature cityscape. Added context identifies ``a man standing next to two astronauts in space'' as NASA astronaut suits displayed in a museum, and ``a phone sitting next to a cup'' as a Samsung phone alongside a chick plushie in a patterned mug. These examples illustrate more accurate, specific, and visually grounded image--text pairs.

\clearpage

\begin{table}[!t]
\centering
\caption{Model checkpoints used in this work. Model names link to resources.}
\label{tab:online_resources}

\begin{tabular}{@{}l@{\qquad\qquad}l@{}}
\toprule
\textbf{Model} & \textbf{License} \\
\midrule
\rowcolor{gray!15}
\multicolumn{2}{l}{\textit{Visual backbone models}} \\

\href{https://huggingface.co/facebook/dinov2-giant}
{\textcolor{black}{DINOv2 (ViT-G/14)}} \citep{dinov2}
& Apache-2.0 \\

\href{https://huggingface.co/facebook/dinov3-vith16plus-pretrain-lvd1689m}
{\textcolor{black}{DINOv3 (ViT-H/16+)}} \citep{dinov3}
& DINOv3 License \\

\href{https://github.com/CompVis/cleandift}
{\textcolor{black}{CleanDIFT}} \citep{cleandift}
& MIT \\

\href{https://huggingface.co/Charles-Elena/stable-diffusion-2-1}
{\textcolor{black}{Stable-Diffusion 2.1}} \citep{sd21}
& CreativeML Open RAIL++-M \\

\href{https://huggingface.co/openai/clip-vit-large-patch14}
{\textcolor{black}{CLIP-ViT-L/14}} \citep{clip}
& MIT \\

\href{https://huggingface.co/timm/vit_large_patch14_clip_224.metaclip_2pt5b}
{\textcolor{black}{MetaCLIP (ViT-L/14)}} \citep{xu2024metaclip}
& CC-BY-NC-4.0 \\

\href{https://huggingface.co/apple/DFN2B-CLIP-ViT-L-14}
{\textcolor{black}{DFN2B-CLIP (ViT-L/14)}} \citep{fang2024dfn}
& Apple AMLR \\

\href{https://huggingface.co/laion/CLIP-ViT-L-14-laion2B-s32B-b82K}
{\textcolor{black}{OpenCLIP (ViT-L/14)}} \citep{cherti2023scaling}
& MIT \\

\href{https://huggingface.co/timm/ViT-L-16-SigLIP2-384}
{\textcolor{black}{SigLIP2 (ViT-L/16)}} \citep{tschannen2025siglip}
& Apache-2.0 \\

\href{https://huggingface.co/qihoo360/fg-clip2-large}
{\textcolor{black}{FG-CLIP2 (ViT-L/16)}} \citep{xie2025fg2}
& Apache-2.0 \\

\midrule
\rowcolor{gray!15}
\multicolumn{2}{l}{\textit{Language \& multimodal models}} \\

\href{https://huggingface.co/sentence-transformers/all-roberta-large-v1}
{\textcolor{black}{RoBERTa-large}} \citep{roberta}
& Apache-2.0 \\

\href{https://huggingface.co/OpenGVLab/InternVL3-8B}
{\textcolor{black}{InternVL3-8B}} \citep{internvl3}
& Apache-2.0 \\

\href{https://huggingface.co/Qwen/Qwen2.5-VL-7B-Instruct}
{\textcolor{black}{Qwen2.5-VL-7B-Instruct}} \citep{bai2025qwen25vl}
& Apache-2.0 \\

\href{https://huggingface.co/Qwen/Qwen3.6-27B}
{\textcolor{black}{Qwen3.6-27B}}
& Apache-2.0 \\

\href{https://huggingface.co/Qwen/Qwen3.6-35B-A3B}
{\textcolor{black}{Qwen3.6-35B-A3B}}
& Apache-2.0 \\

\href{https://huggingface.co/google/gemma-3-27b-it}
{\textcolor{black}{Gemma-3-27B-it}}
& Gemma \\

\href{https://huggingface.co/google/gemma-4-26B-A4B-it}
{\textcolor{black}{Gemma-4-26B-A4B-it}}
& Apache-2.0 \\

\href{https://huggingface.co/google/gemma-4-12B-it}
{\textcolor{black}{Gemma-4-12B-it}}
& Apache-2.0 \\

\href{https://huggingface.co/google/gemma-4-31B-it}
{\textcolor{black}{Gemma-4-31B-it}}
& Apache-2.0 \\

\href{https://platform.openai.com/docs/models/gpt-4.1}
{\textcolor{black}{GPT-4.1 (2025-04-14)}}
& Proprietary (OpenAI) \\

\bottomrule
\end{tabular}
\end{table}
\section{Online Resources}
\label{app:online_resources}

Table~\ref{tab:online_resources} lists the model checkpoints, repositories, and licenses used in this work to facilitate reproducibility and future comparisons. Table~\ref{tab:model_urls} provides their URLs. Our experiments combine publicly available vision encoders, language models, and multimodal foundation models spanning open and proprietary ecosystems.

We group the entries into visual backbones and language or multimodal models to clarify their roles in the experimental pipeline. The visual group covers the constituent encoders of \method{} and the comparison backbones, while the language and multimodal group covers text alignment, image recaptioning, and the evaluation of region-guided responses. Each entry pairs a model identifier with its checkpoint or project page and the corresponding license designation. These identifiers distinguish model families, sizes, and variants that share similar names, helping readers select the same resources when reproducing an experiment. The links complement the architecture, data, and optimization settings in the preceding appendices: checkpoint selection identifies the starting model, whereas those settings specify how we construct and evaluate its representations.

\begin{table}[!t]
\centering
\caption{URLs for the model checkpoints and resources listed in Table~\ref{tab:online_resources}.}
\label{tab:model_urls}
\urlstyle{same}
\setlength{\tabcolsep}{6pt}
\renewcommand{\arraystretch}{1.1}
\begin{tabularx}{\linewidth}{@{}>{\raggedright\arraybackslash}p{0.26\linewidth}@{\hspace{8pt}}>{\raggedright\arraybackslash}X@{}}
\toprule
\textbf{Model} & \textbf{URL} \\
\midrule
\rowcolor{gray!15}[0pt][0pt]
\multicolumn{2}{@{}l@{}}{\textit{Visual backbone models}} \\
DINOv2 (ViT-G/14) & \url{https://huggingface.co/facebook/dinov2-giant} \\
DINOv3 (ViT-H/16+) & \url{https://huggingface.co/facebook/dinov3-vith16plus-pretrain-lvd1689m} \\
CleanDIFT & \url{https://github.com/CompVis/cleandift} \\
Stable-Diffusion 2.1 & \url{https://huggingface.co/Charles-Elena/stable-diffusion-2-1} \\
CLIP-ViT-L/14 & \url{https://huggingface.co/openai/clip-vit-large-patch14} \\
MetaCLIP (ViT-L/14) & \url{https://huggingface.co/timm/vit_large_patch14_clip_224.metaclip_2pt5b} \\
DFN2B-CLIP (ViT-L/14) & \url{https://huggingface.co/apple/DFN2B-CLIP-ViT-L-14} \\
OpenCLIP (ViT-L/14) & \url{https://huggingface.co/laion/CLIP-ViT-L-14-laion2B-s32B-b82K} \\
SigLIP2 (ViT-L/16) & \url{https://huggingface.co/timm/ViT-L-16-SigLIP2-384} \\
FG-CLIP2 (ViT-L/16) & \url{https://huggingface.co/qihoo360/fg-clip2-large} \\
\midrule
\rowcolor{gray!15}[0pt][0pt]
\multicolumn{2}{@{}l@{}}{\textit{Language \& multimodal models}} \\
RoBERTa-large & \url{https://huggingface.co/sentence-transformers/all-roberta-large-v1} \\
InternVL3-8B & \url{https://huggingface.co/OpenGVLab/InternVL3-8B} \\
Qwen2.5-VL-7B-Instruct & \url{https://huggingface.co/Qwen/Qwen2.5-VL-7B-Instruct} \\
Qwen3.6-27B & \url{https://huggingface.co/Qwen/Qwen3.6-27B} \\
Qwen3.6-35B-A3B & \url{https://huggingface.co/Qwen/Qwen3.6-35B-A3B} \\
Gemma-3-27B-it & \url{https://huggingface.co/google/gemma-3-27b-it} \\
Gemma-4-26B-A4B-it & \url{https://huggingface.co/google/gemma-4-26B-A4B-it} \\
Gemma-4-12B-it & \url{https://huggingface.co/google/gemma-4-12B-it} \\
Gemma-4-31B-it & \url{https://huggingface.co/google/gemma-4-31B-it} \\
GPT-4.1 (2025-04-14) & \url{https://platform.openai.com/docs/models/gpt-4.1} \\
\bottomrule
\end{tabularx}
\end{table}

\end{document}